\documentclass[10pt]{article}
\usepackage[a4paper, total={5in, 8in}]{geometry}
\usepackage{times}
\usepackage{epsfig}

\usepackage[utf8]{inputenc}
\usepackage{graphicx,verbatimbox}
\usepackage{tcolorbox}

\usepackage{rotating}
\usepackage{palatino}

\usepackage{tikz}
\usepackage{comment}
\usepackage{amsmath,amssymb} %

\usepackage{booktabs} %
\usepackage{xcolor}
\usepackage{color,colortbl}
\usepackage{pifont}
\usepackage[colorlinks=true,linkcolor=blue, citecolor=cyan]{hyperref}%

\usepackage{multirow}
\usepackage[font=footnotesize]{caption}
\usepackage{placeins}
   
\usepackage{xspace}

\usepackage[accsupp]{axessibility}  %

\usepackage{booktabs}
\usepackage{multirow}

\definecolor{Goldenrod}{RGB}{245,245,220}

\newcommand{\grtext}[1]{\textcolor{lightgray}{#1}}%

\newcommand{\cmark}{\ding{51}\xspace}%
\newcommand{\cmarkg}{\textcolor{lightgray}{\ding{51}}\xspace}%
\newcommand{\xmark}{\ding{55}\xspace}%
\newcommand{\xmarkg}{\textcolor{lightgray}{\ding{55}}\xspace}%

\def \pzo {\phantom{0}} 
\def \dzo {\phantom{00}} 
\def \tzo {\phantom{000}}

\title{DeiT III: Revenge of the ViT} 
\author{
\begin{minipage}{\linewidth}
\begin{center}
\normalsize Hugo Touvron$^{\star,\dagger}$ \hspace{0.23cm} Matthieu Cord$^{\dagger}$ \hspace{0.23cm} Herv\'e J\'egou$^{\star}$ \\[0.5cm] 
\scalebox{1.}{$^\star$Meta AI\hspace{0.6cm} $^\dagger$Sorbonne University}\\
\end{center}
\end{minipage}
}

\date{~}

\begin{document}

\maketitle
\begin{abstract} 
A Vision Transformer (ViT) is a simple neural architecture amenable to serve several computer vision tasks. It has limited built-in architectural priors, in contrast to more recent architectures that  incorporate priors either about the input data or of specific tasks. 
Recent works show that ViTs benefit from self-supervised pre-training, in particular BerT-like pre-training like BeiT. 

In this paper, we revisit the supervised training of ViTs. 
Our procedure builds upon and simplifies a recipe introduced for training ResNet-50. It includes a new simple data-augmentation procedure with only 3 augmentations, closer to the practice in self-supervised learning.  
Our evaluations on Image classification (ImageNet-1k with and without pre-training on ImageNet-21k), transfer learning and semantic segmentation show that our procedure outperforms by a large margin previous fully supervised training recipes for ViT. It also reveals that the performance of our ViT trained with supervision is comparable to that of more recent architectures. Our results could serve as better baselines for recent self-supervised approaches demonstrated on ViT. 

\begin{figure}[b!]
    \centering
    \begin{tabular}{cc}
        ImageNet-1k & \quad \quad \quad ImageNet-21k \\
         \includegraphics[width=0.42\linewidth]{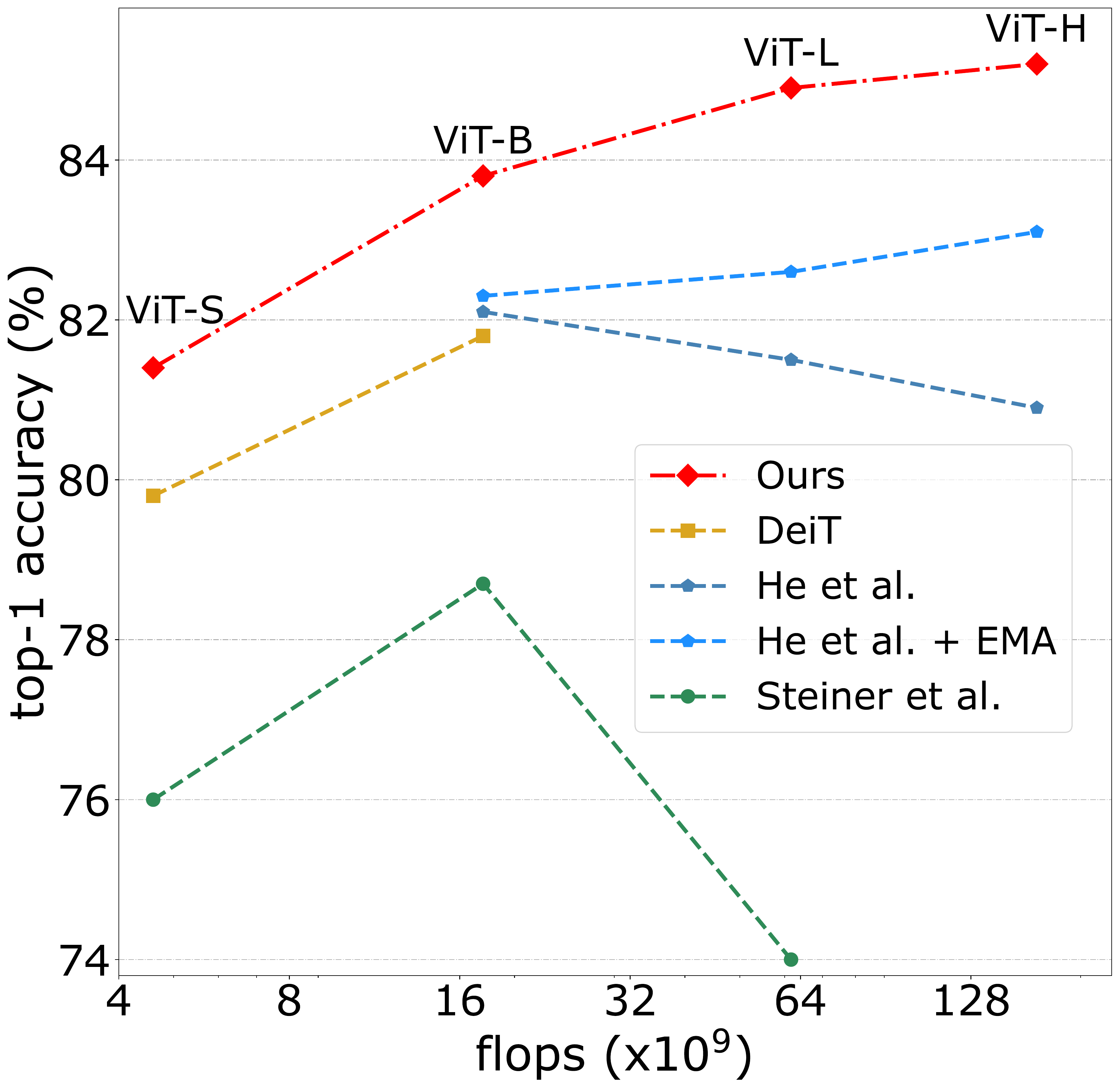}   & 
         \includegraphics[width=0.42\linewidth]{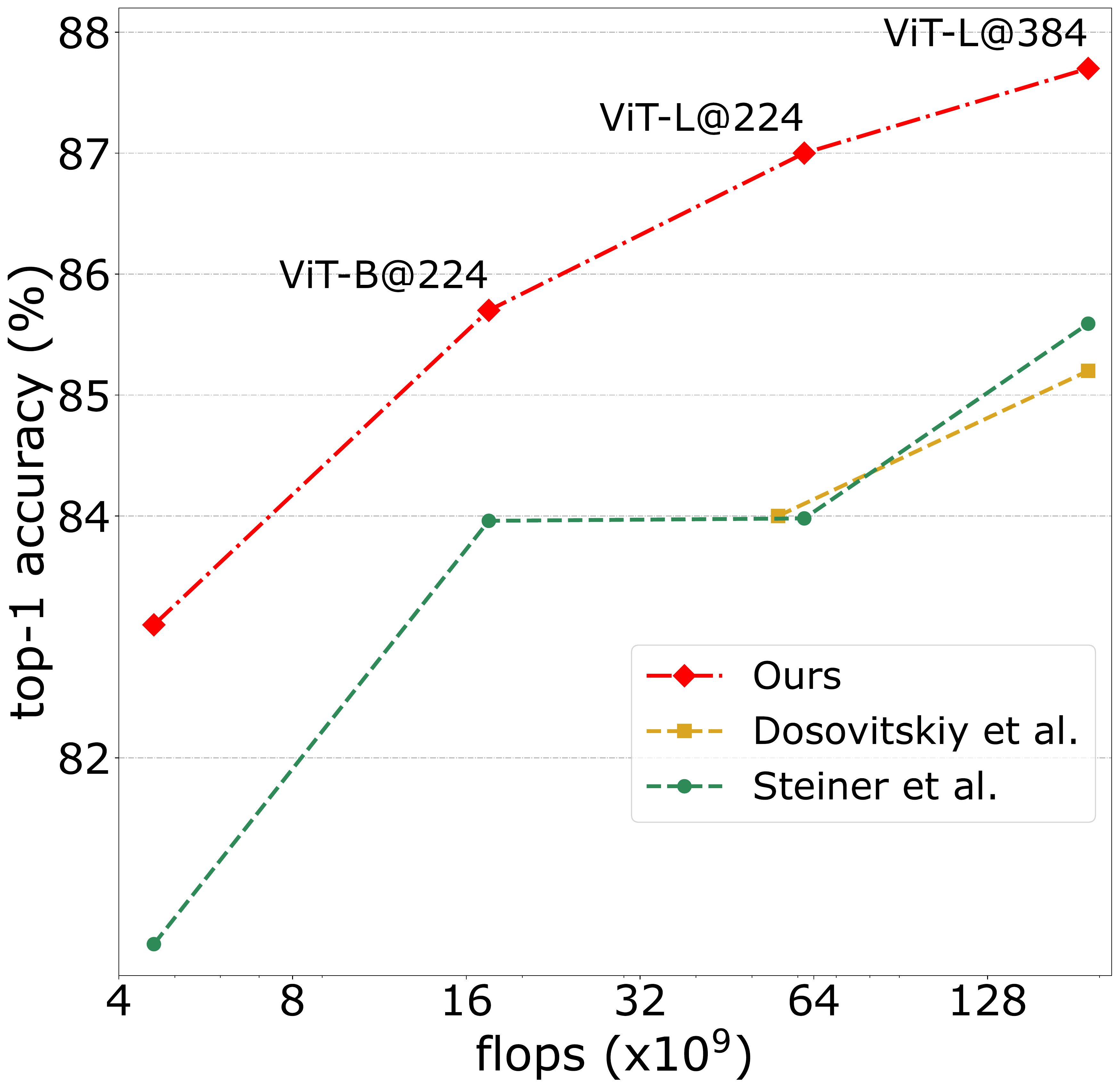} \\
    \end{tabular}
    \caption{Comparison of training recipes for (\emph{left}) vanilla vision transformers trained on ImageNet-1k and evaluated at resolution 224$\times$224, and (\emph{right}) pre-trained on ImageNet-21k at 224$\times$224 and finetuned on ImageNet-1k at resolution 224$\times$224 or 384$\times$384.%
    \label{fig:method_fig}}
\end{figure}

\end{abstract}

\section{Introduction}

After their vast success in NLP, transformers models~\cite{vaswani2017attention}  and their derivatives are increasingly popular in computer vision. 
They are increasingly used in image classification~\cite{dosovitskiy2020image}, detection \& segmentation~\cite{carion2020end}, video analysis, etc.
In particular, the vision transformers (ViT) of Dosovistky et al.~\cite{dosovitskiy2020image} are a reasonable alternative to convolutional architectures. 
This supports the adoption of transformers as a general architecture able to learn convolutions as well as longer range 
operations through the attention process~\cite{cordonnier2019relationship,dAscoli2021ConViTIV}. 
In contrast, convolutional networks~\cite{He2016ResNet,Krizhevsky2012AlexNet,lecun1998gradient,Simonyan2015VGG} implicitly offer built-in translation invariance. As a result their training does not have to learn this prior. It is therefore not surprising that hybrid architectures that include convolution converge faster than vanilla transformers~\cite{graham2021levit}. 

Because they incorporate as priors only the co-localisation of pixels in patches, transformers have to learn about the structure of images while optimizing the model such that it processes the input with the objective of solving a given task. This can be either reproducing labels in the supervised case, or other proxy tasks in the case of self-supervised approaches. 
Nevertheless, despite their huge success, there has been only few works in computer vision studying how to efficiently train vision transformers, and in particular on a midsize dataset like ImageNet-1k. 
Since the work of Dosovistky et al.~\cite{dosovitskiy2020image}, the training procedures are mostly variants from the proposal of Touvron et al.~\cite{Touvron2020TrainingDI} and Steiner et al.~\cite{Steiner2021HowTT}. %
In contrast, multiple works have proposed alternative architectures by introducing pooling, more efficient attention, or hybrid architectures re-incorporating convolutions and a pyramid structure. These new designs, while being particularly effective for some tasks, are less general. One difficult question to address is whether the improved performance is due to a specific architectural design, or because it facilitates the optimization as suggested it is the case for convolutions with ViTs~\cite{xiao2021early}. 

Recently, self-supervised approaches inspired by the popular BerT pre-training have raised hopes for a BerT moment in computer vision. There are some analogies between the fields of NLP and computer vision, starting with the transformer architecture itself. However these fields are not identical in every way: The modalities processed are of different nature (continuous versus discrete). Computer vision offer large annotated databases like ImageNet~\cite{Russakovsky2015ImageNet12}, and fully supervised pre-training on ImageNet is effective for handling different downstream tasks such as transfer learning~\cite{oquab2014learning} or semantic segmentation.

Without further work on fully supervised approaches on ImageNet it is difficult to conclude if the intriguing performance of self-supervised approaches like BeiT~\cite{bao2021beit} is due to the training, e.g. data augmentation, regularization, optimization, or to an underlying mechanism that is capable of learning more general implicit representations.
In this paper, we do not pretend to answer this difficult question, but we want to feed this debate by renewing the training procedure for vanilla ViT architectures. We hope to contribute to a better understanding on how to fully exploit the potential of transformers and of the importance of BerT-like pre-training.
Our work builds upon  the recent state of the art on fully supervised and self-supervised approaches, with new insights regarding data-augmentation. 
We propose new training recipes for vision transformers on ImageNet-1k and ImageNet-21k.
The main ingredients are as follows:
\begin{itemize}
\item We build upon the work of Wightman et al.~\cite{wightman2021resnet} introduced for ResNet50. In particular we adopt a binary cross entropy loss for Imagenet1k only training. We adapt this method by including ingredients that significantly improve the training of large ViT~\cite{touvron2021going}, namely stochastic depth~\cite{Huang2016DeepNW} and LayerScale~\cite{touvron2021going}. 

   \item \textbf{3-Augment}: is a simple data augmentation inspired by that employed for self-supervised learning. Surprisingly, with ViT we observe that it works better than the usual automatic/learned data-augmentation employed to train vision transformers like RandAugment~\cite{Cubuk2019RandAugmentPA}.
    
    \item \textbf{Simple Random Cropping} is more effective than Random Resize Cropping when pre-training on a larger set like ImageNet-21k. 
    
    \item \textbf{A lower resolution} at training time. This choice reduces the train-test discrepancy~\cite{Touvron2019FixRes} but has not been much exploited with ViT. We observe that it also has a regularizing effect  for the largest models by preventing overfitting. 
    For instance, for a target resolution of $224\times 224$, a ViT-H pre-trained at resolution $126\times 126$ (81 tokens) achieves a better performance on ImageNet-1k than when pre-training  at resolution $224\times 224$ (256 tokens). This is also less demanding at pre-training time, as there are 70\% fewer tokens. From this perspective it offers similar scaling properties as mask-autoencoders~\cite{He2021MaskedAA}. 
\end{itemize}

Our ``new'' training strategies do not saturate with the largest models, making another step beyond the Data-Efficient Image Transformer (DeiT) by Touvron et al.~\cite{Touvron2020TrainingDI}. 
As a result, we obtain a competitive performance in image classification and segmentation, even when compared to recent popular architectures such as SwinTransformers~\cite{liu2021swin} or modern convnet architectures like ConvNext~\cite{Liu2022convnext}. Below we point out a few interesting outcomes.

\begin{itemize}
    \item  We leverage models with more capacity even on midsize datasets. For instance we reach 85.2\% in top-1 accuracy when training a ViT-H on ImageNet1k only, which is an improvement of $+5.1\%$ over the best ViT-H with supervised training procedure reported in the literature at resolution 224$\times$224.
    
    \item Our training procedure for ImageNet-1k allow us to train a \textbf{billion-parameter ViT-H}  (52 layers) without any hyper-parameter adaptation, just using the same stochastic depth drop-rate as for the ViT-H. It attains 84.9\% at 224$\times$224, i.e., +0.2\% higher than the corresponding ViT-H trained in the same setting.

    \item Without sacrificing performance, we  \textbf{divide by more than 2} the number of GPUs required and the training time for ViT-H, making it effectively possible to train such models without a reduced amount of resources. This is thanks to our pre-training at lower resolution, which reduces the peak memory.
    
     \item For ViT-B and Vit-L models, our supervised training approach is on par with BerT-like self-supervised approaches~\cite{bao2021beit,He2021MaskedAA} with their default setting and when using the same level of annotations and less epochs, both for the tasks of image classification and of semantic segmentation. 
    
    \item With this improved training procedure, a vanilla ViT closes the gap with recent state-of-the art architectures, often offering better compute/performance trade-offs. Our models are also comparatively better on the additional test set ImageNet-V2~\cite{Recht2019ImageNetv2}, which indicates that our trained models generalize better to another validation set than most prior works. 

    \item An ablation on the effect of the crop ratio employed in transfer learning classification tasks. We observe that it has a noticeable impact on the performance but that the best value depends a lot on the target dataset/task. 
\end{itemize}

\section{Related work}

\paragraph{\textbf{Vision Transformers}} were introduced by Dosovitskiy et al.~\cite{dosovitskiy2020image}. This architecture, which derives from the transformer by Vaswani et al.~\cite{vaswani2017attention},  
is now used as an alternative to convnets in many tasks: image classification~\cite{dosovitskiy2020image,Touvron2020TrainingDI}, detection~\cite{carion2020end,liu2021swin}, semantic segmentation~\cite{bao2021beit,liu2021swin} video analysis~\cite{Fan2021MultiscaleVT,Neimark2021videovit}, to name only a few. 
This greater flexibility typically comes with the downside that 
they need larger datasets, or the training must be adapted when the data is scarcer~\cite{el2021large,Touvron2020TrainingDI}.
Many variants have been introduced to reduce the cost of attention by introducing for example more efficient attention~\cite{el2021xcit,Fan2021MultiscaleVT,liu2021swin} or pooling layers~\cite{Heo2021RethinkingSD,liu2021swin,Wang2021PyramidVT}. 
Some papers re-introduce spatial biases specific to convolutions within hybrid architectures~\cite{graham2021levit,Wu2021CvTIC,xiao2021early}.  These models are less general than vanilla transformers but generally perform well in certain computer vision tasks, because their architectural priors reduce the need to learn from scratch the task biases. This is especially important for smaller models, where specialized models do not have to devote some capacity to reproduce known priors such as translation invariance. The models are formally less flexible but they do not require sophisticated training procedures.
\paragraph{\textbf{Training procedures:}}

The first procedure proposed in the ViT paper~\cite{dosovitskiy2020image} was mostly effective for larger models trained on large datasets. In particular the ViT were not competitive with convnets when trained from scratch on ImageNet. Touvron et al.~\cite{Touvron2020TrainingDI} showed that by adapting the training procedure, it is possible to achieve a performance comparable to that of convnets with Imagenet training only. After this Data Efficient Image Transformer procedure (DeiT), only few adaptations have been proposed to improve the training vision transformers. Steiner et al.~\cite{Steiner2021HowTT} published a complete study on how to train vision transformers on different datasets by doing a complete ablation of the different training components. Their results on ImageNet~\cite{Russakovsky2015ImageNet12} are slightly inferior to those of DeiT but they report improvements on ImageNet-21k compared to Dosovitskiy et al.~\cite{dosovitskiy2020image}. The self-supervised approach referred to as masked auto-encoder (MAE)~\cite{He2021MaskedAA} proposes an improved supervised baseline for the larger ViT models. 

\paragraph{\textbf{BerT pre-training:}}

In the absence of a strong fully supervised training procedure, BerT~\cite{devlin2018bert}-like approaches that train ViT with a self-supervised proxy objective, followed by full finetuning on the target dataset, seem to be the best paradigm to fully exploit the potential of vision transformers. Indeed, BeiT~\cite{bao2021beit} or  MAE~\cite{He2021MaskedAA} significantly outperform the fully-supervised approach, especially for the largest models. Nevertheless, to date these approaches have mostly shown their interest in the context of mid-size datasets. For example MAE~\cite{He2021MaskedAA}  report its most impressive results when pre-training on ImageNet-1k with a full finetuning on ImageNet-1k. When pre-training on ImageNet-21k and finetuning on ImageNet-1k, BeiT~\cite{bao2021beit} requires a full 90-epochs finetuning on ImageNet-21k followed by another full finetuning on ImageNet-1k to reach its best performance, suggesting that a large labeled dataset is needed so that BeiT realizes its best potential. A recent work suggests that such auto-encoders are mostly interesting in a data starving context~\cite{el2021training}, but this questions their advantage in the case where more labelled data is actually available. 

\paragraph{\textbf{Data-augmentation:}}
For supervised training, the community commonly employs data-augmentations offered by automatic design procedures such as RandAugment~\cite{Cubuk2019RandAugmentPA} or Auto-Augment~\cite{Ekin2018AutoAugment}. These data-augmentations seem to be essential for training vision transformers~\cite{Touvron2020TrainingDI}. Nevertheless, papers like TrivialAugment~\cite{Mller2021TrivialAugmentTY} and Uniform Augment~\cite{LingChen2020UniformAugmentAS} have shown that it is possible to reach interesting performance levels when simplifying the approaches. However these approaches were initially optimized for convnets. 
In our work we propose to go further in this direction and drastically limit and simplify data-augmentation: we introduce a data-augmentation policy that employs only 3 different transformations randomly drawn with uniform probability. That's it.

\section{\makebox{Revisit training \& pre-training for Vision Transformers}}

In this section, we present our training procedure for vision transformers and compare it with existing approaches. We detail the different ingredients in Table~\ref{tab:comp_hyperparameters}.
Building upon Wightman et al.~\cite{wightman2021resnet} and Touvron et al.~\cite{Touvron2020TrainingDI}, we introduce several changes that have a significant impact on the final model accuracy.

\begin{table}[t]
\caption{
Summary of our training procedures with ImageNet-1k and ImageNet-21k. We also provide DeiT~\cite{Touvron2020TrainingDI}, Wightman et al~\cite{wightman2021resnet} and Steiner et al.~\cite{Steiner2021HowTT} baselines for reference. Adapt. means the hparams is adapted to the size of the model. 
For finetuning to higher resolution with model pre-trained on ImageNet-1k only we use the finetuning procedure from DeiT see section~\ref{sec:finetuning_hres} for more details.%
\label{tab:comp_hyperparameters}}
\centering
\scalebox{0.79}
{%
\begin{tabular}{@{\ }l|cccc|ccc@{\ }}
\toprule
 & \multicolumn{4}{c|}{Previous approaches} &  \multicolumn{3}{c}{Ours} \\
\cmidrule{2-5}
\cmidrule{6-8}
Procedure $\rightarrow$ & 
ViT & 
Steiner &
DeiT & 
Wightman & 
ImNet-1k & \multicolumn{2}{c}{ImNet-21k} \\
Reference & 
\cite{dosovitskiy2020image}& 
et al. \cite{Steiner2021HowTT} &
\cite{Touvron2020TrainingDI} & 
et al. \cite{wightman2021resnet} & & Pretrain. & Finetune. \\[3pt]
\midrule
Batch size & 
4096 & 
4096 &
1024 & 
2048 & 
2048 & 2048 & 2048\\
Optimizer &
AdamW & 
AdamW &
AdamW &
LAMB &
LAMB &
LAMB & LAMB\\
LR      & 
$3.10^{-3}$ & 
$3.10^{-3}$ &
$1.10^{-3}$  & 
$5.10^{-3}$  & 
$3.10^{-3}$ &
$3.10^{-3}$ & 
$3.10^{-4}$\\
LR decay& 
cosine  &
cosine & 
cosine & 
cosine & 
cosine &
cosine & cosine\\
Weight decay     &
0.1  & 
0.3  & 
0.05 & 
0.02 &
0.02 &
0.02 & 0.02\\
Warmup epochs & 
3.4 &
3.4 &
5 & 
5  &
5 &
5 & 5 \\
\midrule
Label smoothing $\varepsilon$ & 
0.1 & 
0.1 &
0.1  &
\xmarkg & 
\xmarkg  &
0.1  & 0.1 \\%
Dropout      & 
\cmark  & 
\cmark & 
\xmarkg & 
\xmarkg  & 
\xmarkg &
\xmarkg & \xmarkg\\
Stoch. Depth & 
\xmarkg & 
\cmark & 
\cmark & 
\cmark &
\cmark&
\cmark & \cmark \\
Repeated Aug & 
\xmarkg & 
\xmarkg & 
\cmark &
\cmark &
\cmark &
\xmarkg & \xmarkg\\
Gradient Clip. & 
1.0  & 
1.0 & 
\xmarkg & 
1.0 & 
1.0 &
1.0 & 1.0\\
\midrule
H. flip  & 
\cmark & 
\cmark &
\cmark & 
\cmark & 
\cmark  &
\cmark  & \cmark
\\
RRC & 
\cmark & 
\cmark & 
\cmark & 
\cmark & 
\cmark & 
\xmarkg & \xmarkg\\
Rand Augment  &
\xmarkg & 
Adapt. &
9/0.5 &
7/0.5 &
\xmarkg &
\xmarkg & \xmarkg  \\
3 Augment (ours)  &
\xmarkg  & 
\xmarkg  &
\xmarkg  &
\xmarkg &
\cmark &
\cmark & \cmark \\

LayerScale  & 
\xmarkg  & 
\xmarkg  & 
\xmarkg  &
\xmarkg  &
  \cmark&
  \cmark &  \cmark\\

Mixup alpha  & 
\xmarkg & 
Adapt. & 
0.8 &
0.2 & 
0.8  &
\xmarkg & \xmarkg \\
Cutmix alpha &
\xmarkg & 
\xmarkg & 
1.0 &
1.0 & 
1.0 &
1.0 & 1.0 \\
Erasing prob. &
\xmarkg    &
\xmarkg    &
0.25 &
\xmarkg  &
\xmarkg & 
\xmarkg & \xmarkg \\
ColorJitter  & 
\xmarkg   & 
\xmarkg   & 
\xmarkg  &
\xmarkg  &
  0.3 & 
 0.3 & 0.3\\

\midrule
Test  crop ratio & 
0.875 & 
0.875 &
0.875 & 
0.95 &
1.0 & 
1.0 & 1.0 \\
\midrule
Loss &
CE & 
CE & 
CE &
BCE & 
BCE &
CE & CE \\

 \bottomrule
\end{tabular}}
\end{table}

\subsection{Regularization \& loss}

\paragraph{\textbf{Stochastic depth}}is a regularization method that is especially useful for training deep networks. We use a uniform drop rate across all layers and adapt it according to the model size~\cite{touvron2021going}. Table~\ref{tab:std_depth_rate} (\ref{appd:hparams}) gives the stochastic depth drop-rate per model.

\paragraph{\textbf{LayerScale}.}
We use LayerScale~\cite{touvron2021going}. This method was introduced to facilitate the convergence of deep transformers. With our training procedure, we do not have convergence problems, however we observe that LayerScale allows our models to attain a higher accuracy for the largest  models. 
In the original paper~\cite{touvron2021going}, the initialization of LayerScale is adapted according to the depth. 
In order to simplify the method we use the same initialization ($10^{-4}$) for all our models.

\paragraph{\textbf{Binary Cross entropy.}}

Wigthman et al.~\cite{wightman2021resnet} adopt a binary cross-entropy (BCE) loss instead of the more common cross-entropy (CE) to train ResNet-50.
They conclude that the gains are limited compared to the CE loss but that this choice is more convenient when employed with Mixup~\cite{Zhang2017Mixup} and CutMix~\cite{Yun2019CutMix}. 
For larger ViTs and with our training procedure on ImageNet-1k, the BCE loss provides us a significant improvement in performance, see an ablation in Table~\ref{tab:ablation}. 
We did not achieve compelling results during our exploration phase on Imagenet21k, and therefore keep CE when pre-training with this dataset as well as for the subsequent fine-tuning.  

\paragraph{\textbf{The optimizer}}
  is LAMB \cite{you20lamb}, a derivative of AdamW~\cite{Loshchilov2017AdamW}. It includes gradient clipping by default in \texttt{Apex}'s~\cite{Apex} implementation.

\subsection{Data-augmentation}

Since the advent of AlexNet, there has been significant modifications to the data-augmentation procedures employed to train neural networks. Interestingly, the same data augmentation, like RandAugment~\cite{Cubuk2019RandAugmentPA}, is widely employed for ViT while their policy was initially learned for convnets. Given that the architectural priors and biases are quite different in these architectures, the augmentation policy may not be adapted, and possibly overfitted considering the large amount of choices involved in their selection. We therefore revisit this prior choice.

\paragraph{\textbf{3-Augment:}}

\begin{figure}
    \begin{tabular}{@{}c@{\ \ \ }c@{\ \ \ }c@{\ \ \ }c@{\ \ \ }c@{\ \ \ }c@{}}
        \rotatebox{90}{\ \ Original} & 
                \includegraphics[width = 0.17\linewidth]{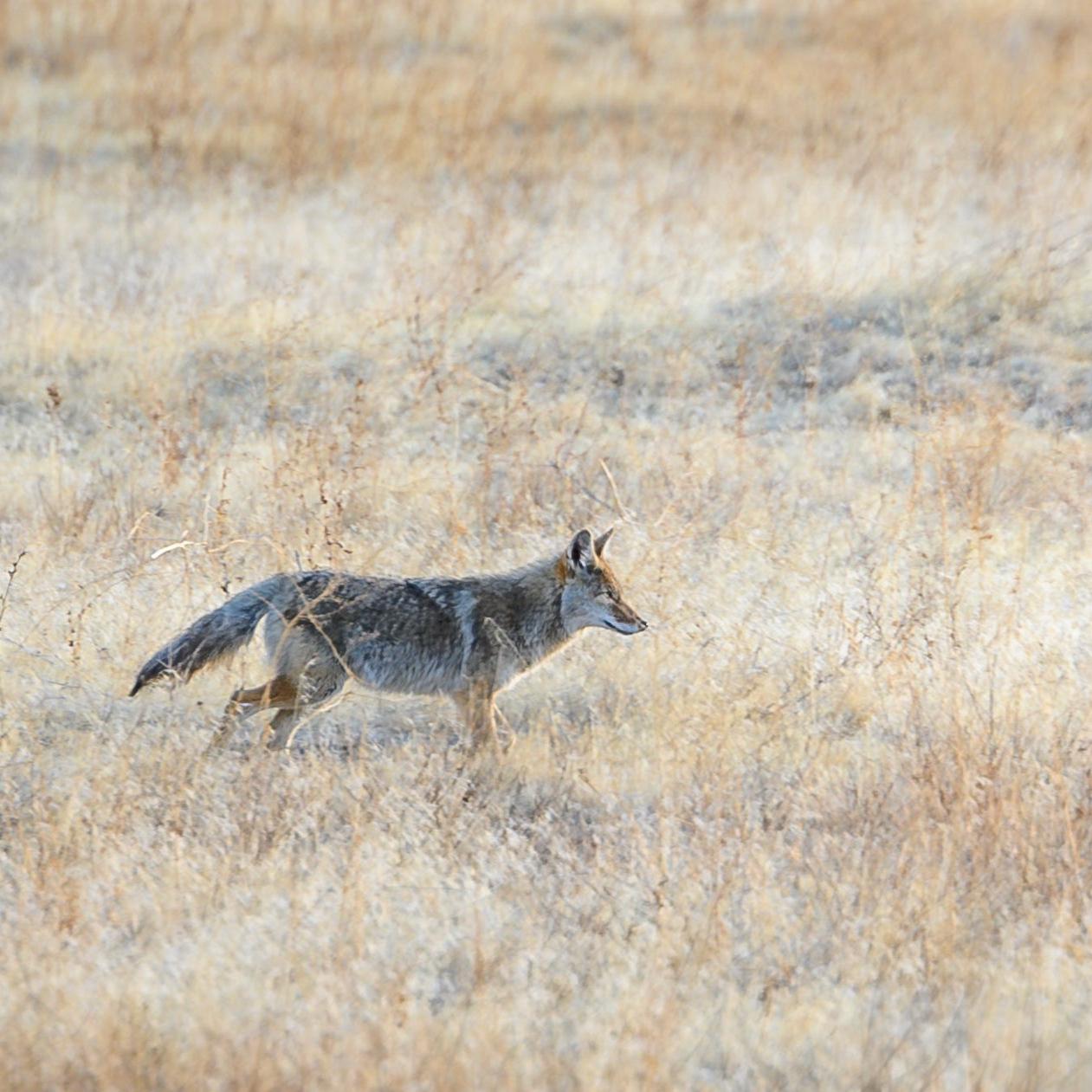} & 
                \includegraphics[width = 0.17\linewidth]{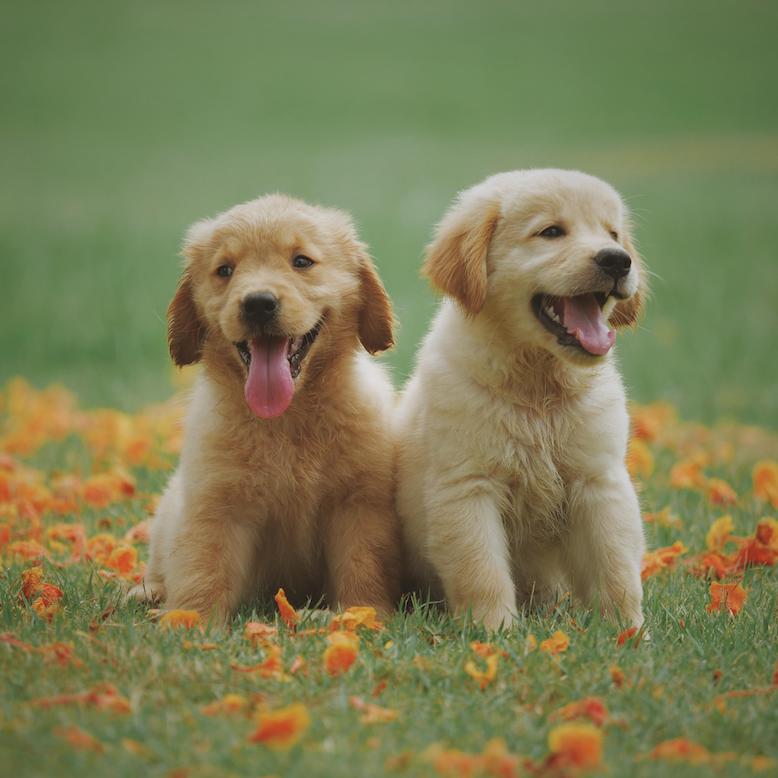} &
                \includegraphics[width = 0.17\linewidth]{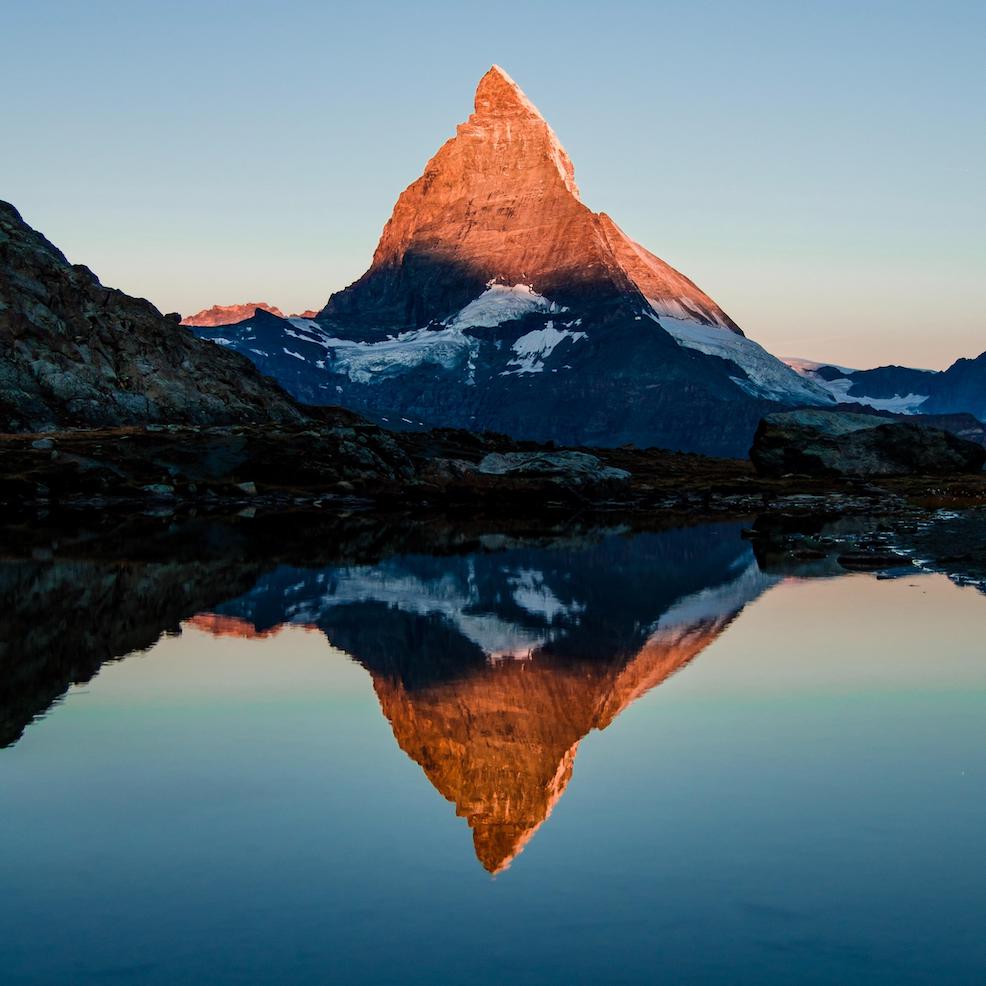} &
                \includegraphics[width = 0.17\linewidth]{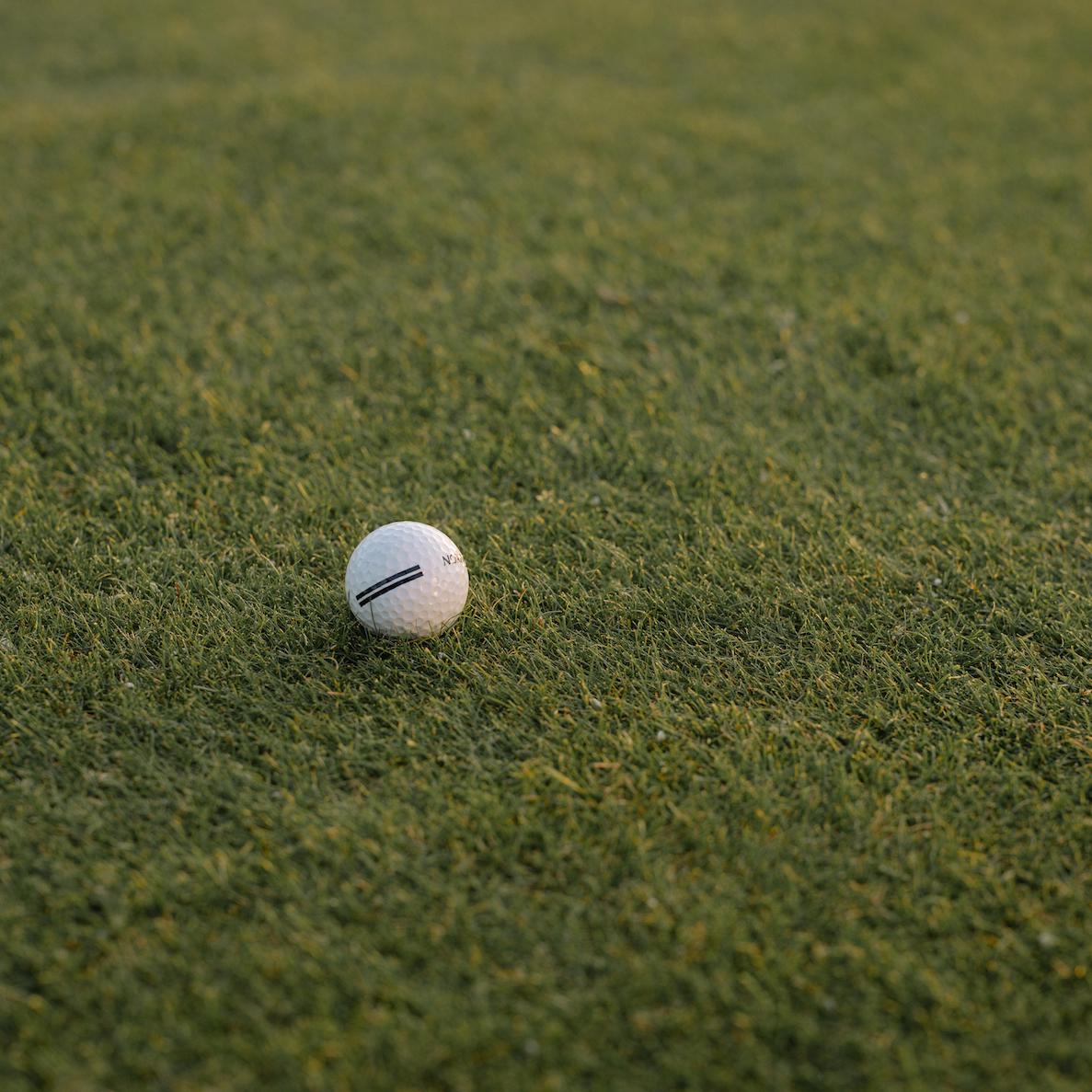} &
                \includegraphics[width = 0.17\linewidth]{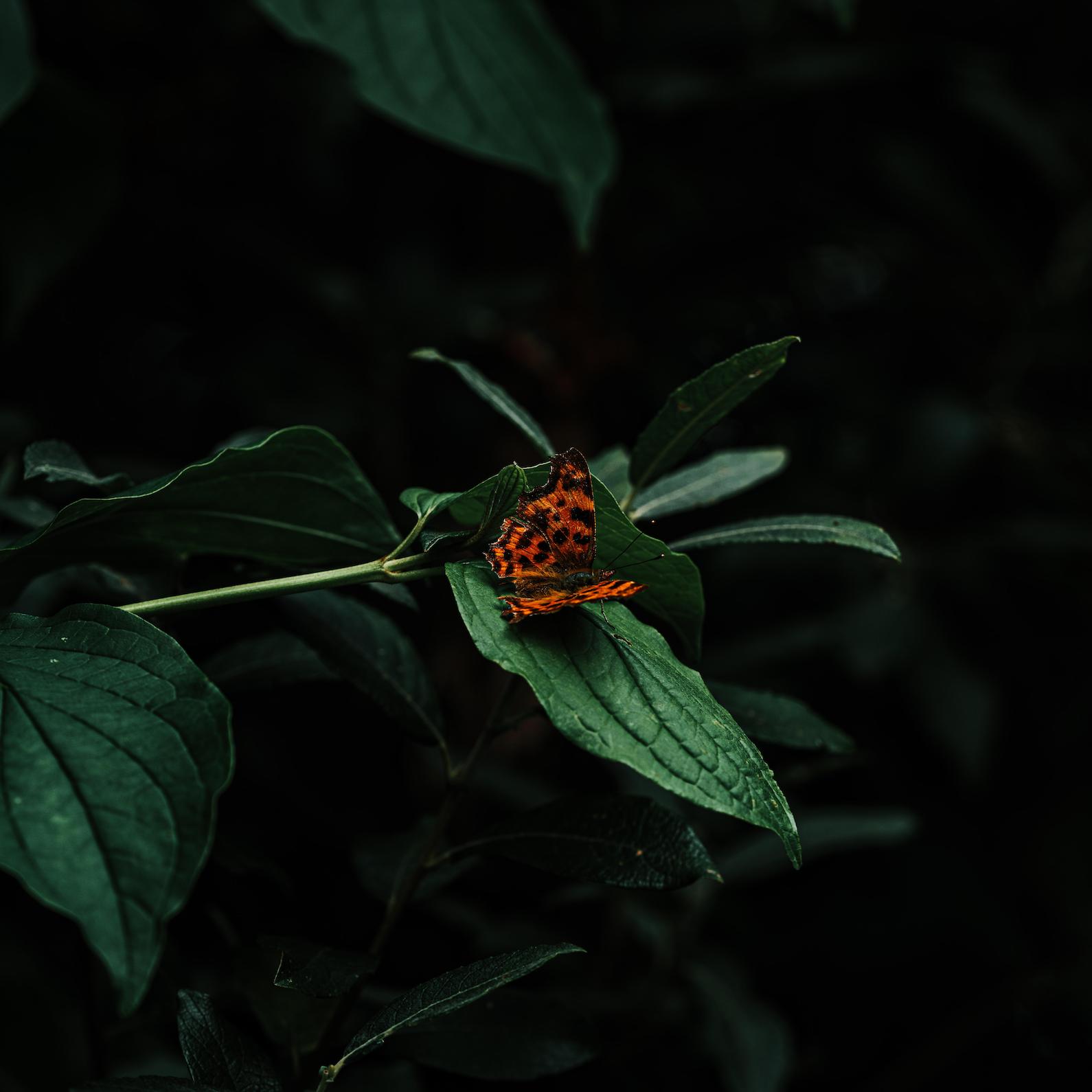}
                \\
         \rotatebox{90}{Gauss. Blurr} & 
                \includegraphics[width = 0.17\linewidth]{data_aug/org_pexels-brett-sayles-6124723.jpg}  & 
                \includegraphics[width = 0.17\linewidth]{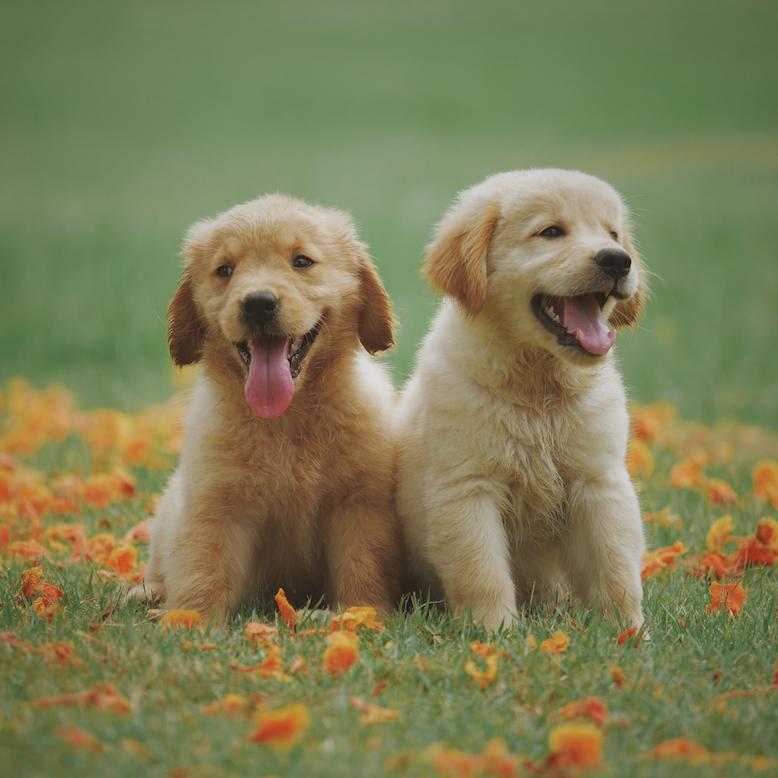} &
                \includegraphics[width = 0.17\linewidth]{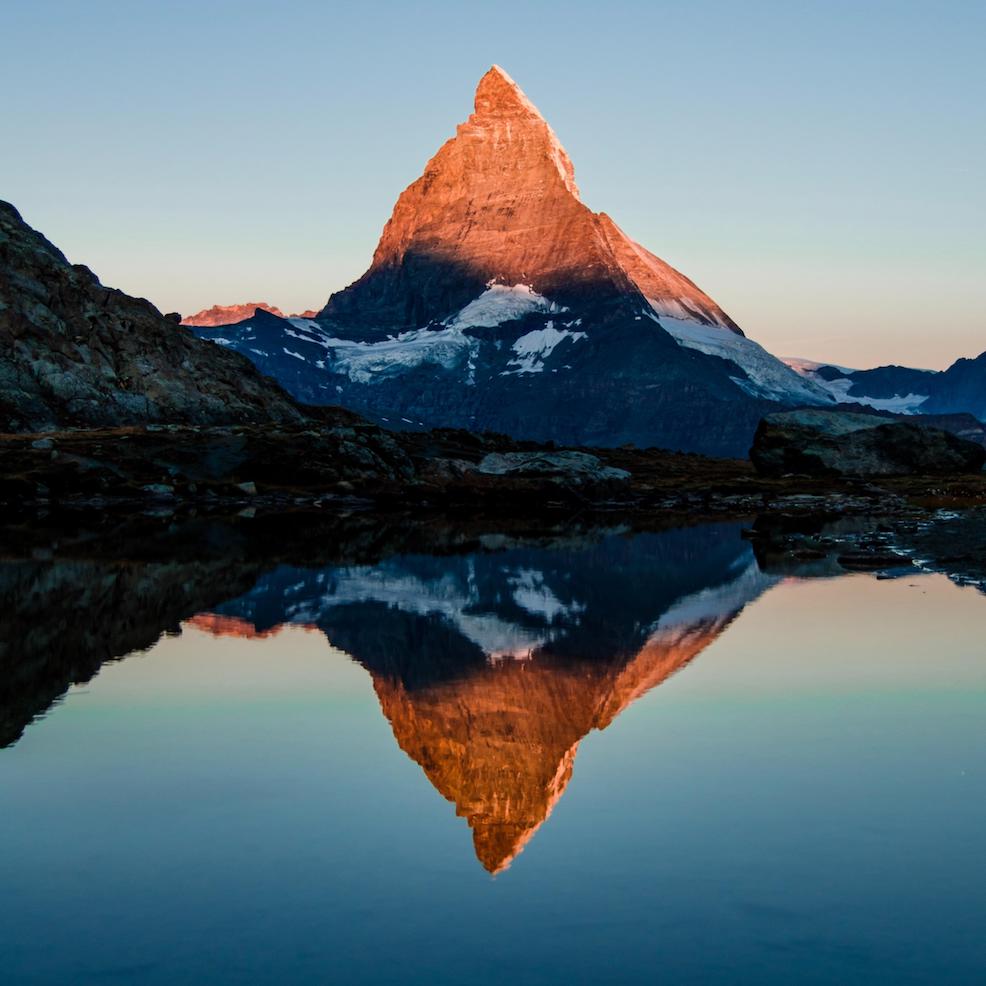} &
                \includegraphics[width = 0.17\linewidth]{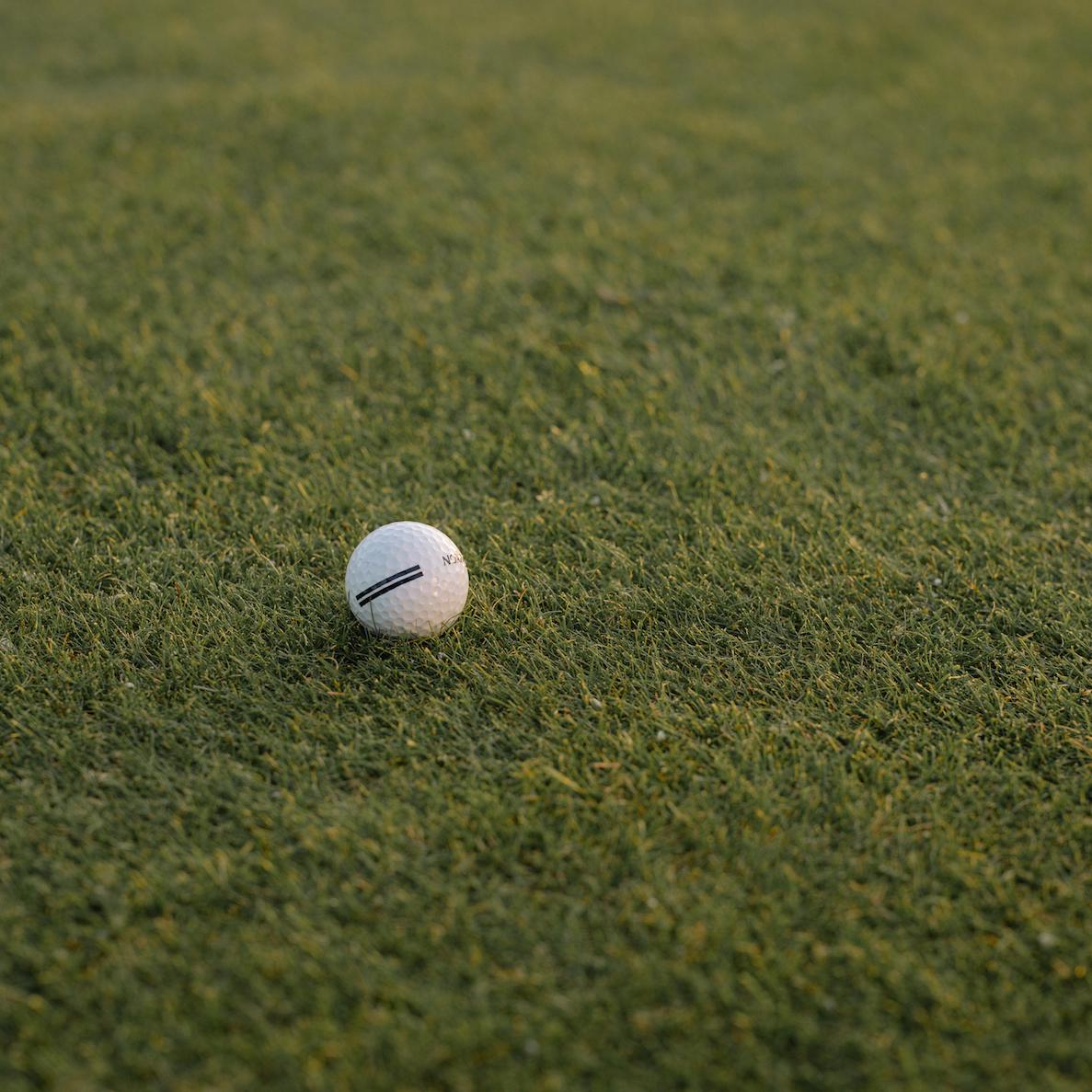} &
                \includegraphics[width = 0.17\linewidth]{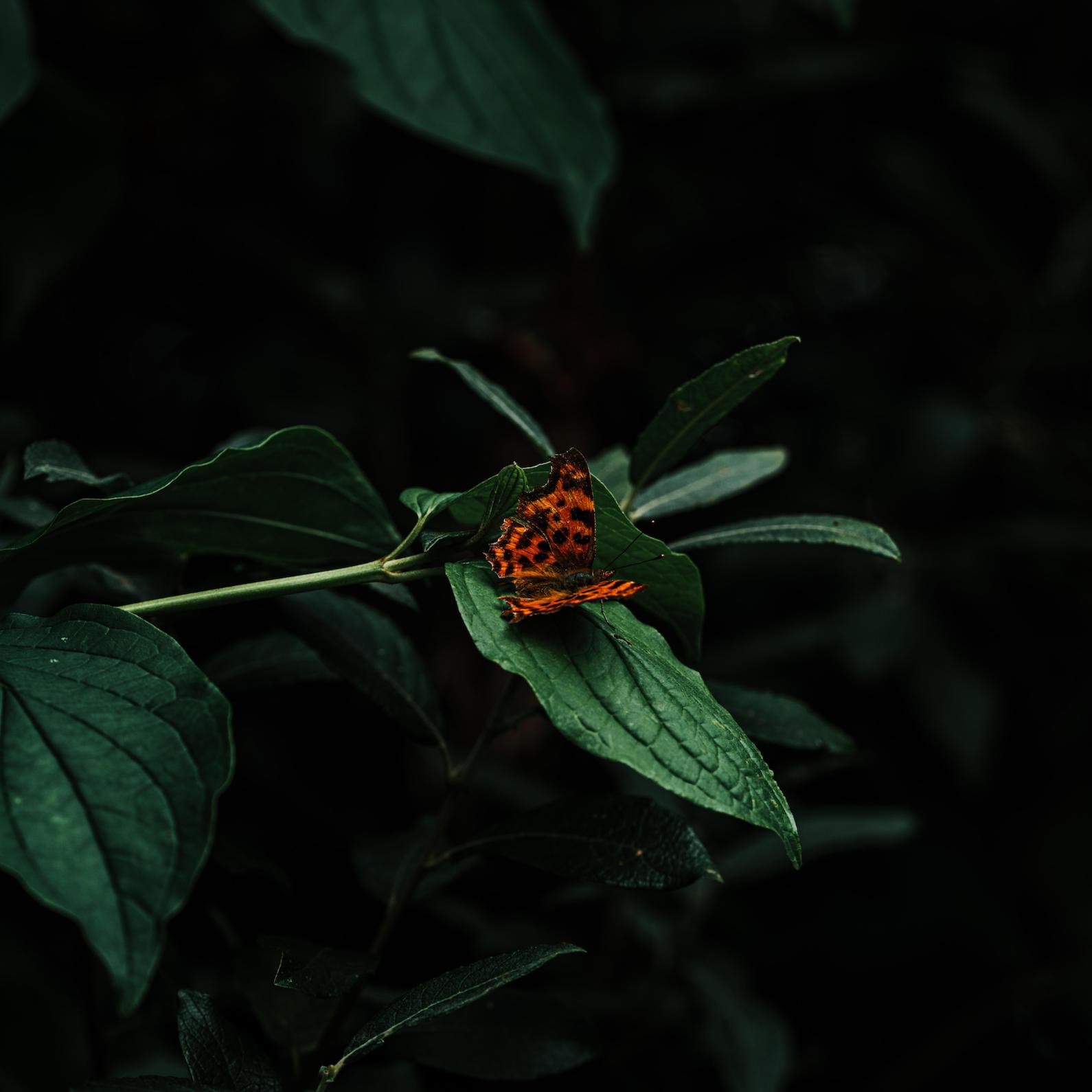}
                \\
        \rotatebox{90}{\ \ Grayscale} & 
                \includegraphics[width = 0.17\linewidth]{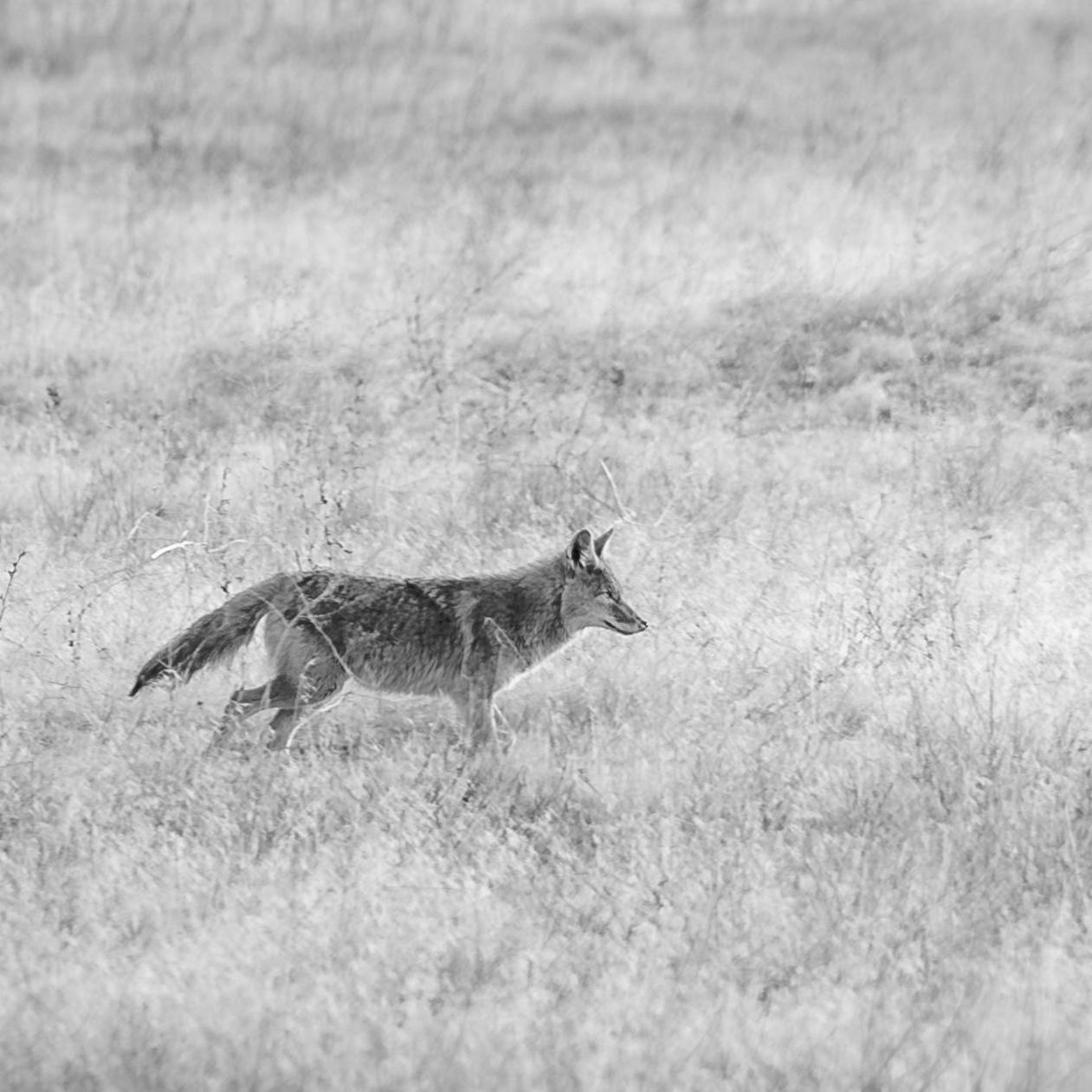}  & 
                \includegraphics[width = 0.17\linewidth]{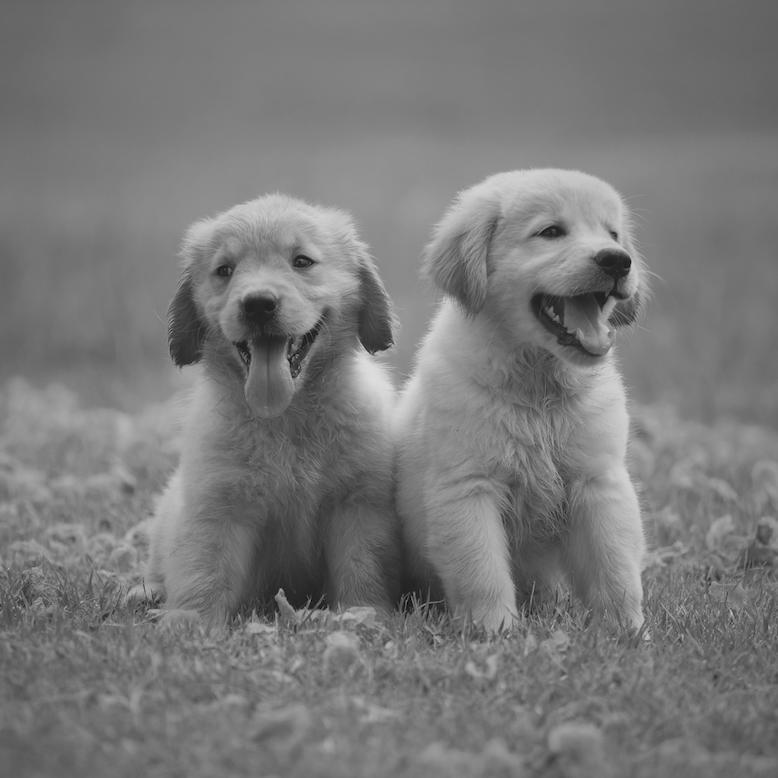} &
                \includegraphics[width = 0.17\linewidth]{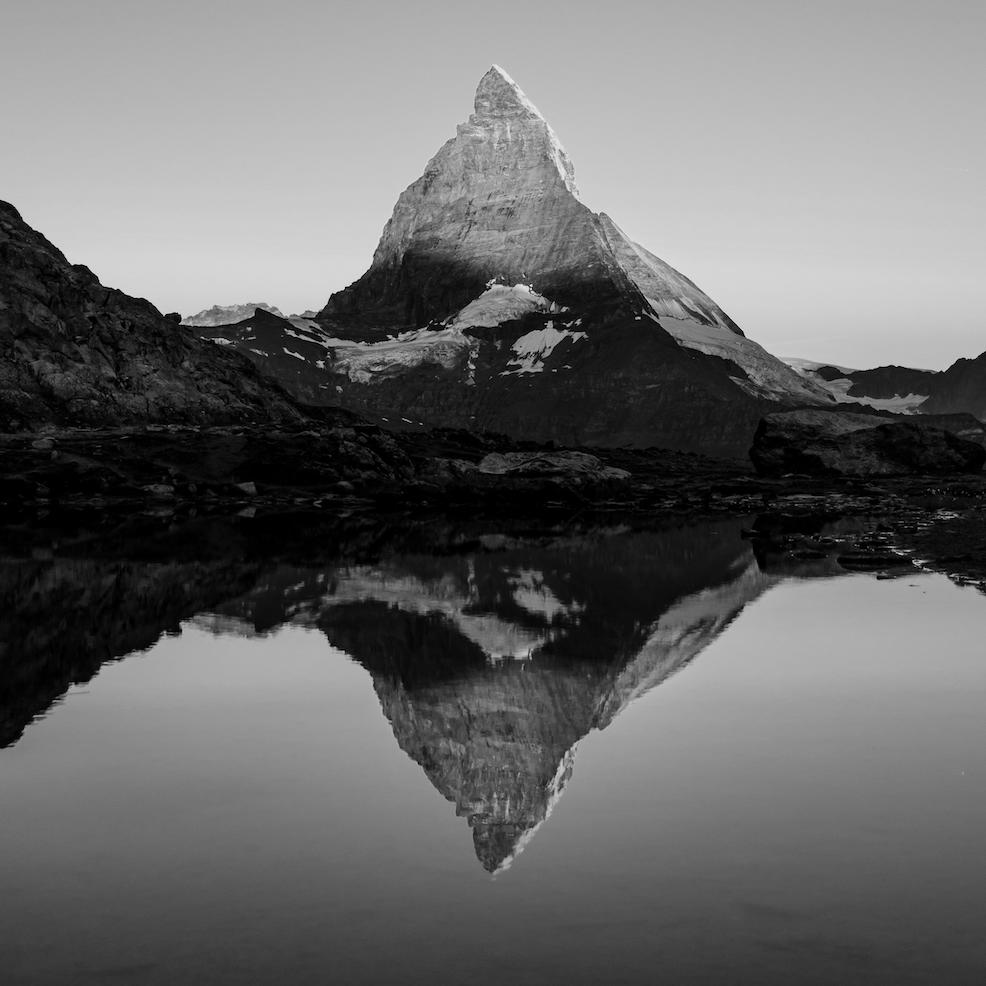} &
                \includegraphics[width = 0.17\linewidth]{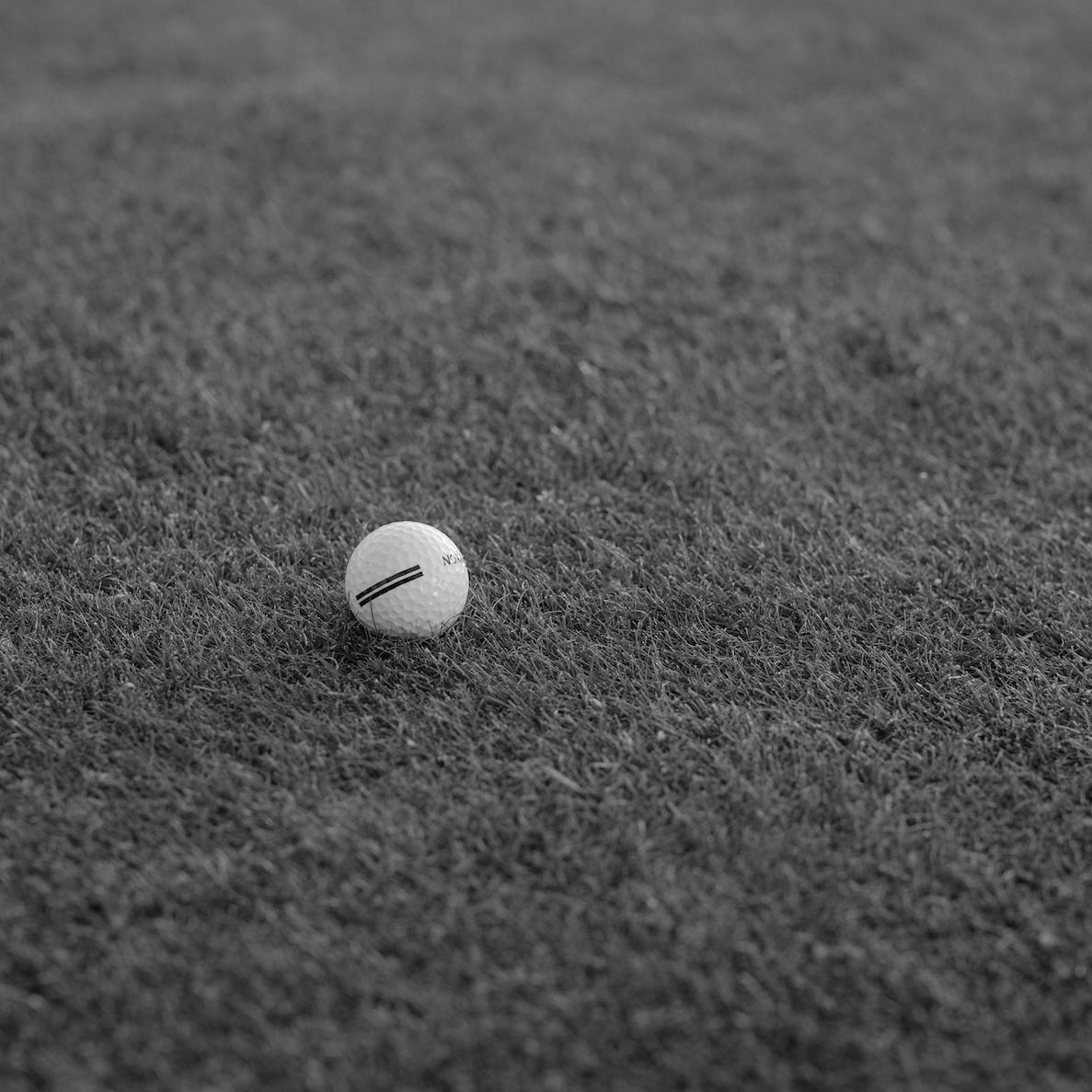} &
                \includegraphics[width = 0.17\linewidth]{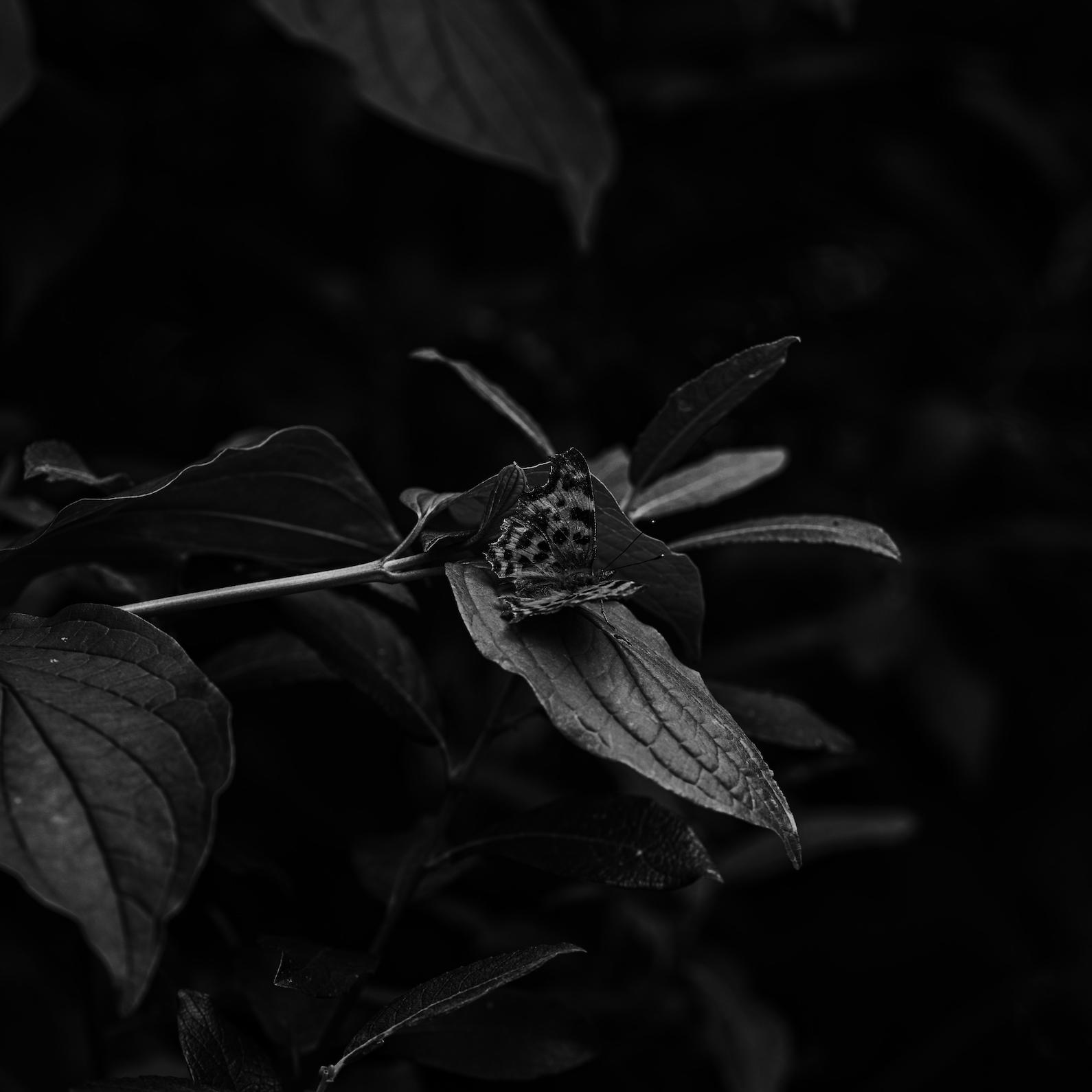}
                \\
        \rotatebox{90}{Solarization} & 
                \includegraphics[width = 0.17\linewidth]{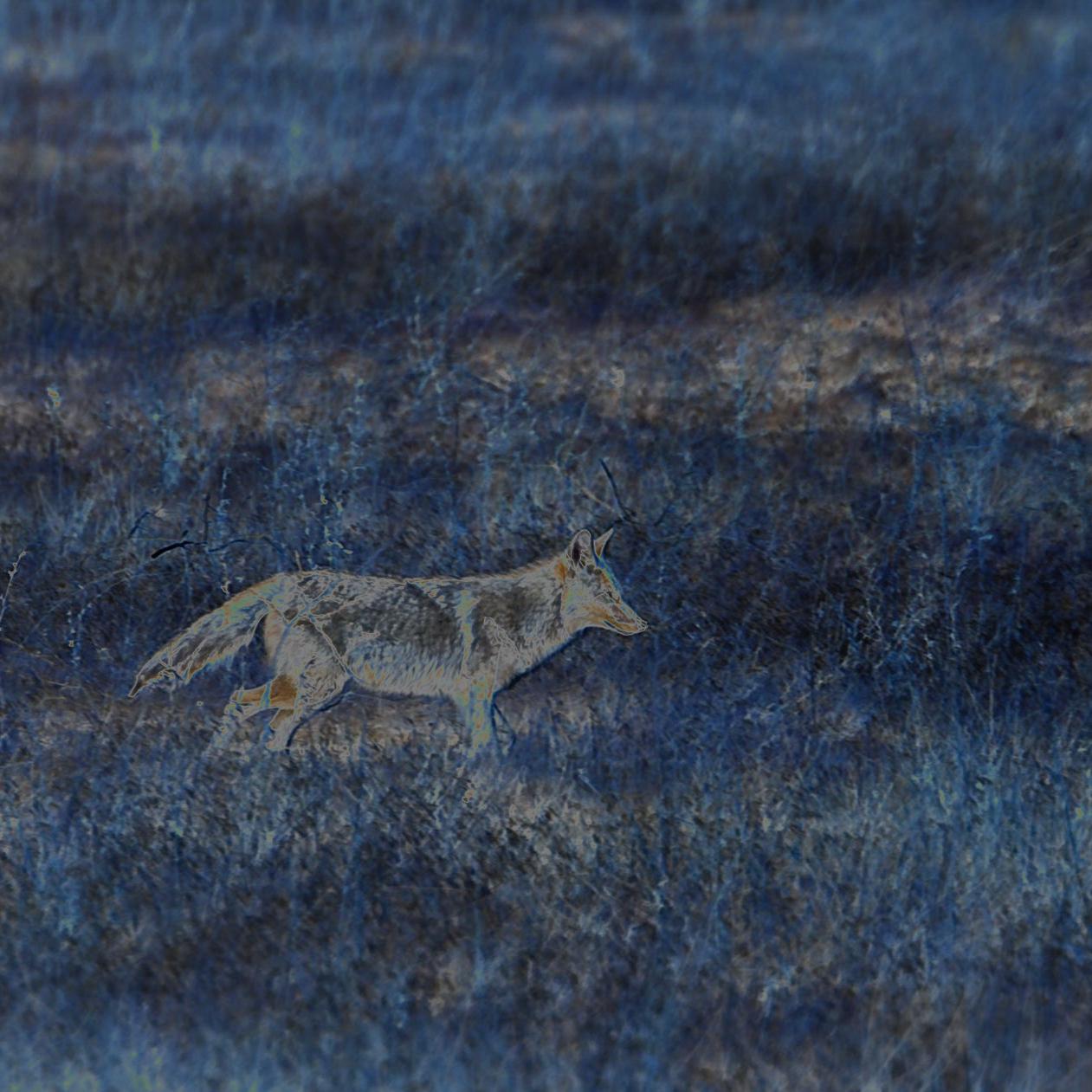}  & 
                \includegraphics[width = 0.17\linewidth]{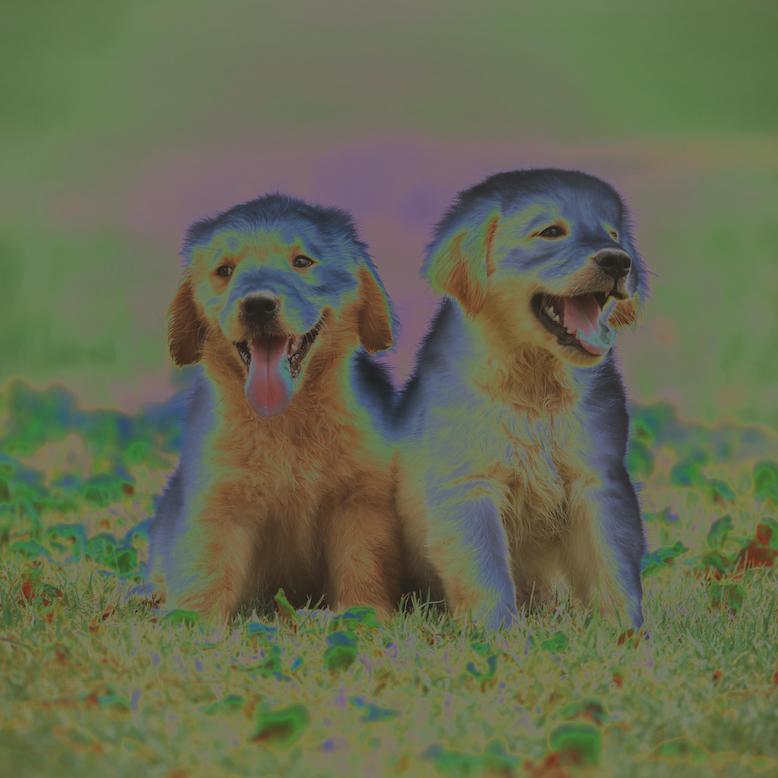} &
                \includegraphics[width = 0.17\linewidth]{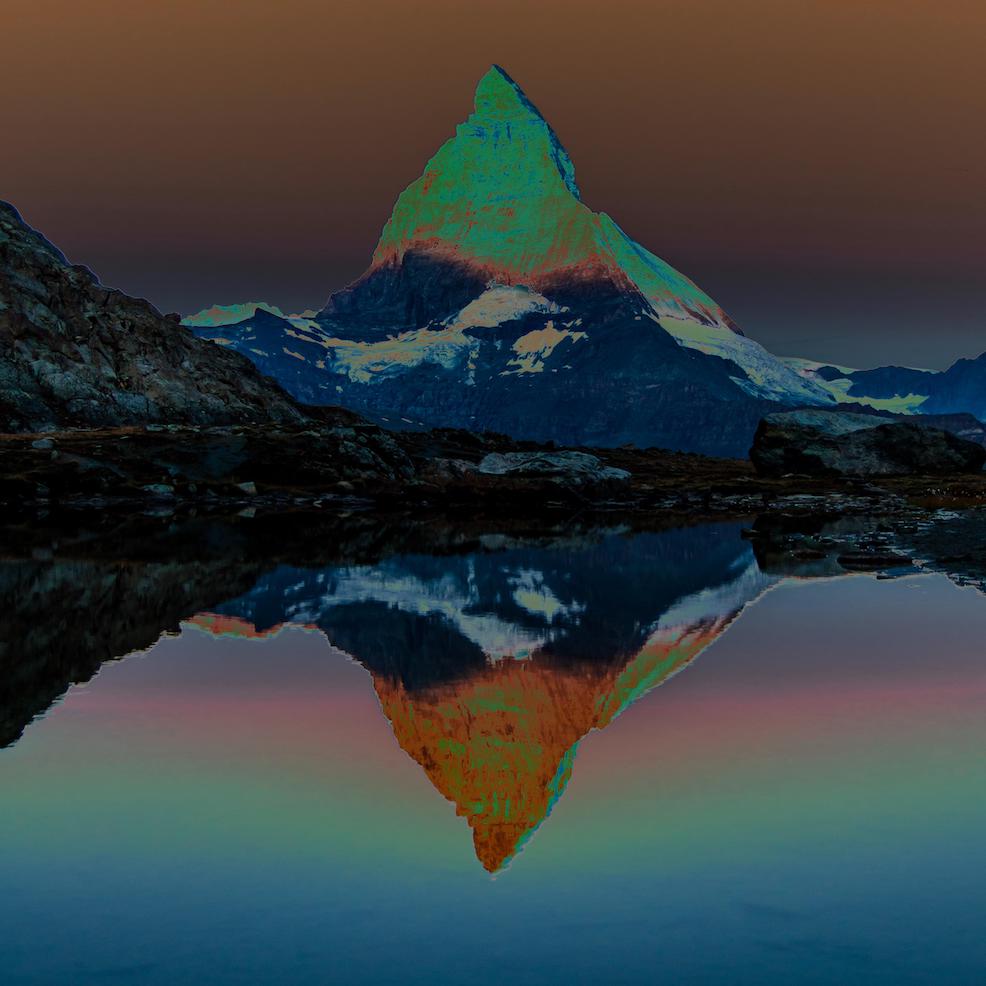} &
                \includegraphics[width = 0.17\linewidth]{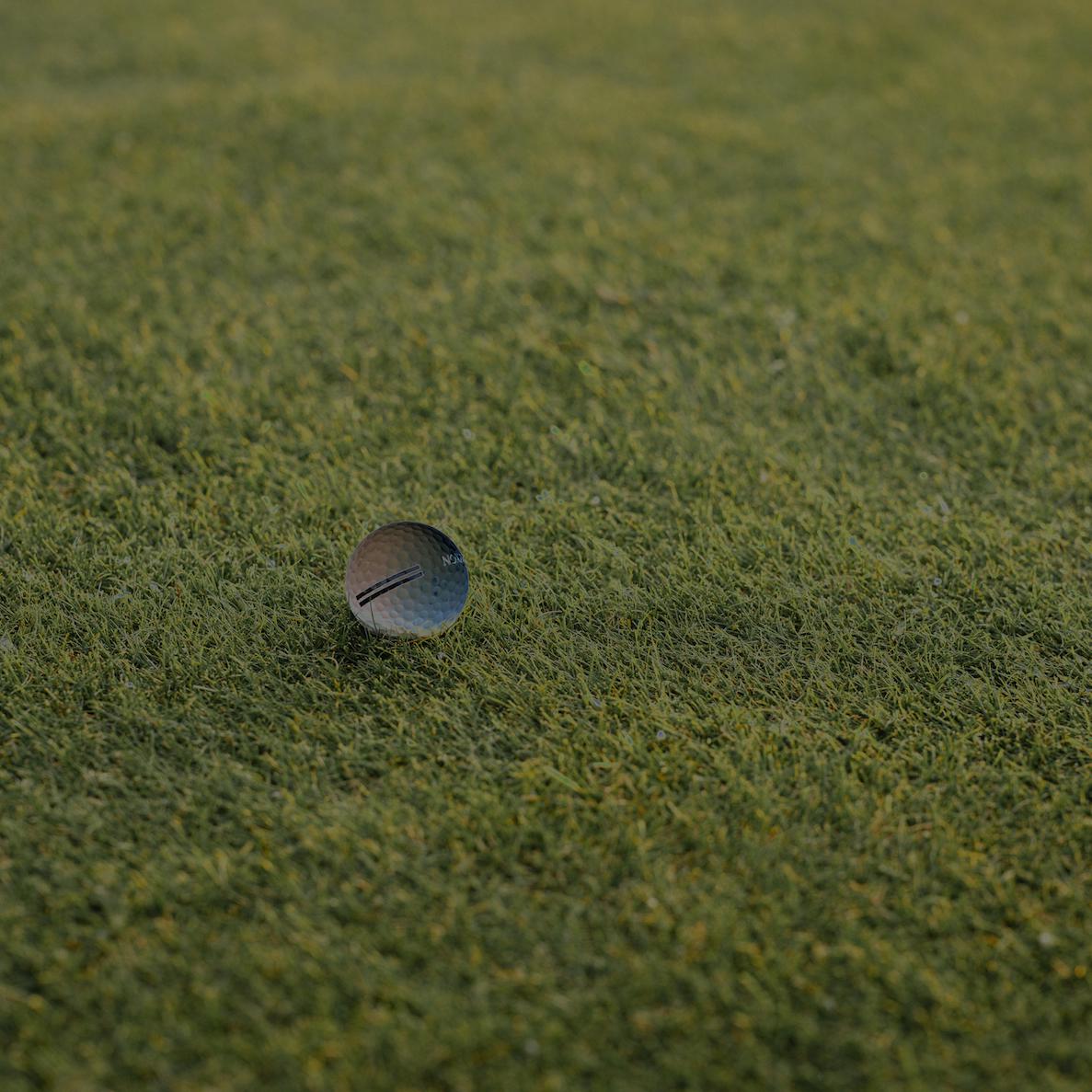} &
                \includegraphics[width = 0.17\linewidth]{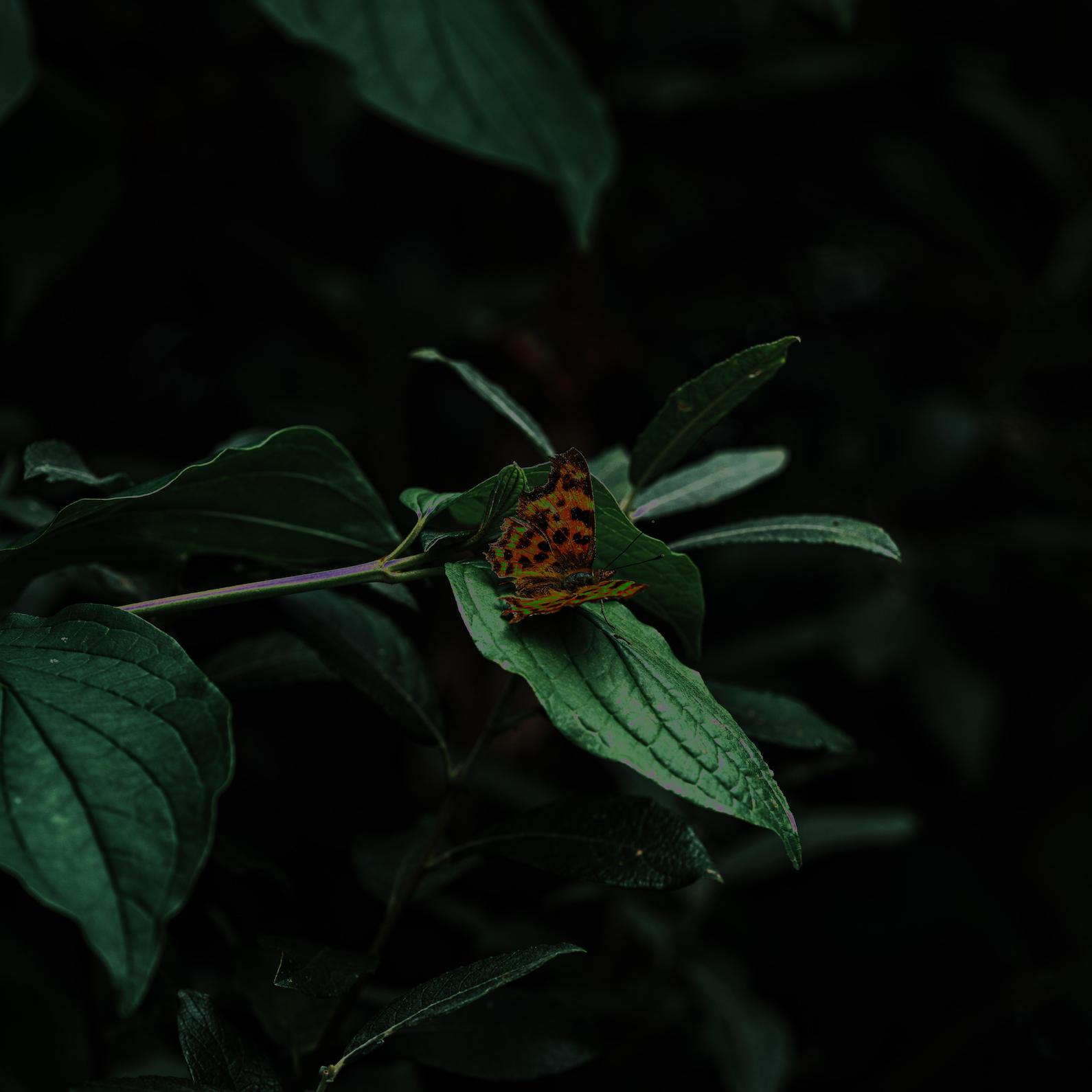}
    \end{tabular}
    \vspace{-6pt}
    \caption{Illustration of the 3 types of data-augmentations used in 3-Augment.
    \label{fig:illust_da}}
\end{figure}

We propose a simple data augmentation inspired by what is used in self-supervised learning (SSL).
We consider the following transformations:
\begin{itemize}
    \item Grayscale: \makebox{This favors color invariance and give more focus on shapes.}
    
    \item Solarization: This adds strong noise on the colour to be more robust to the variation of colour intensity and so focus more on shape. 
    
    \item Gaussian Blur: In order to slightly alter details in the image.

\end{itemize}
For each image, we select only one of this data-augmentation with a uniform probability over 3 different ones. In addition to these 3 aumgnentations choices, we include the common color-jitter and horizontal flip.
Figure~\ref{fig:illust_da} illustrates the different augmentations used in our 3-Augment approach.
In Table~\ref{tab:da_ablation} we provide an ablation on our different data-augmentation components.

\begin{table}
    \centering
    \scalebox{0.8}{
    \begin{tabular}{cccc|c c c}
        \toprule
         \multicolumn{4}{c}{Data-Augmentation} & \multicolumn{3}{c}{ImageNet-1k} \\
          ColorJitter & Grayscale & Gaussian Blur & Solarization   & Val\  \ & \ \ Real\ \  & \ \ V2 \\
          \midrule
          \textcolor{lightgray}{0.3} & \xmarkg &  \xmarkg &  \xmarkg  & 81.4 & 86.1 & 70.3\\
          \textcolor{lightgray}{0.3}  & \cmark &  \xmarkg &  \xmarkg  & 81.0 & 86.0 &  69.7\\
          \textcolor{lightgray}{0.3}  & \cmarkg &  \cmark &  \xmarkg  & 82.7 & 87.6 & \textbf{72.7} \\
          \rowcolor{Goldenrod}
          0.3 & \cmark &  \cmark &  \cmark  & \textbf{83.1} & \textbf{87.7} & 72.6 \\
          0.0 & \cmarkg &  \cmarkg &  \cmarkg  &  \textbf{83.1} &\textbf{87.7} & 72.0\\
        \bottomrule
    \end{tabular}}
    \caption{Ablation of the components of our data-augmentation strategy with ViT-B on ImageNet-1k. %
    }
    \label{tab:da_ablation}
\end{table}

\subsection{Cropping}
\label{sec:rrc}

\begin{figure}
    \centering
    \begin{tabular}{cc}
         RRC &  SRC \\
        \includegraphics[width =0.48\linewidth]{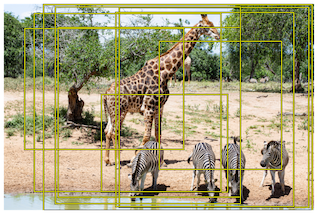}& \includegraphics[width =0.48\linewidth]{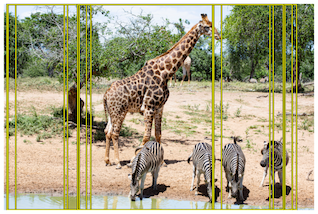}\\
    \end{tabular}
    \caption{Example of crops selected by two strategies: Resized Crop and Simple Random Crop.}
    \label{fig:rrc_bbox}
\end{figure}

\begin{figure}
    \centering
    \begin{tabular}{c@{\ \ }cc@{\ \ }cc@{\ \ }c}
         \multicolumn{2}{c}{\includegraphics[width = 0.3\linewidth]{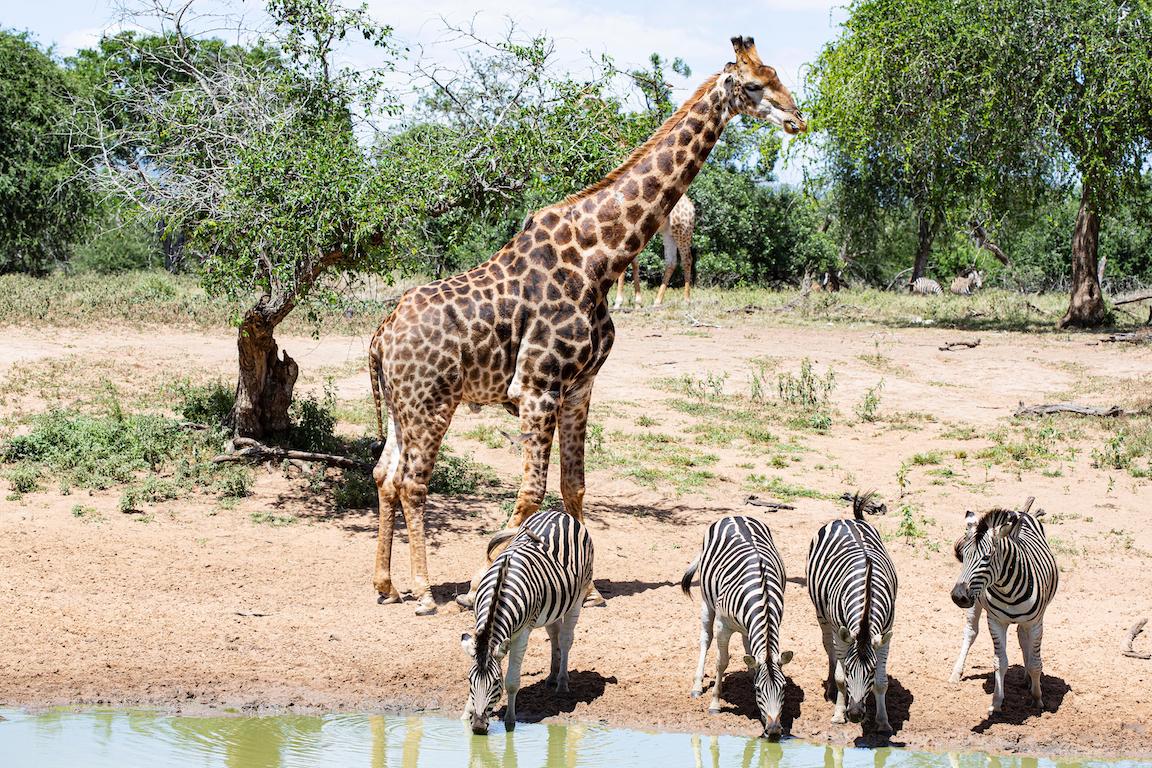}}& 
         \multicolumn{2}{c}{\includegraphics[width = 0.3\linewidth]{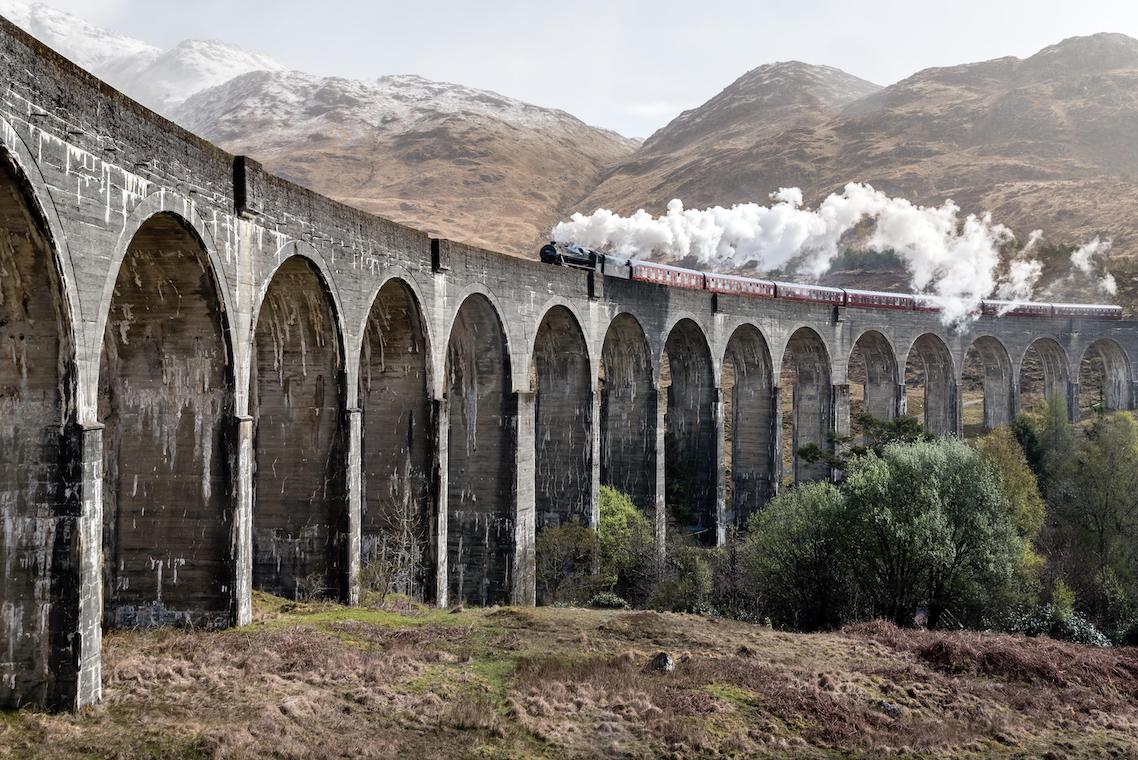}}&
         \multicolumn{2}{c}{\includegraphics[width = 0.25\linewidth]{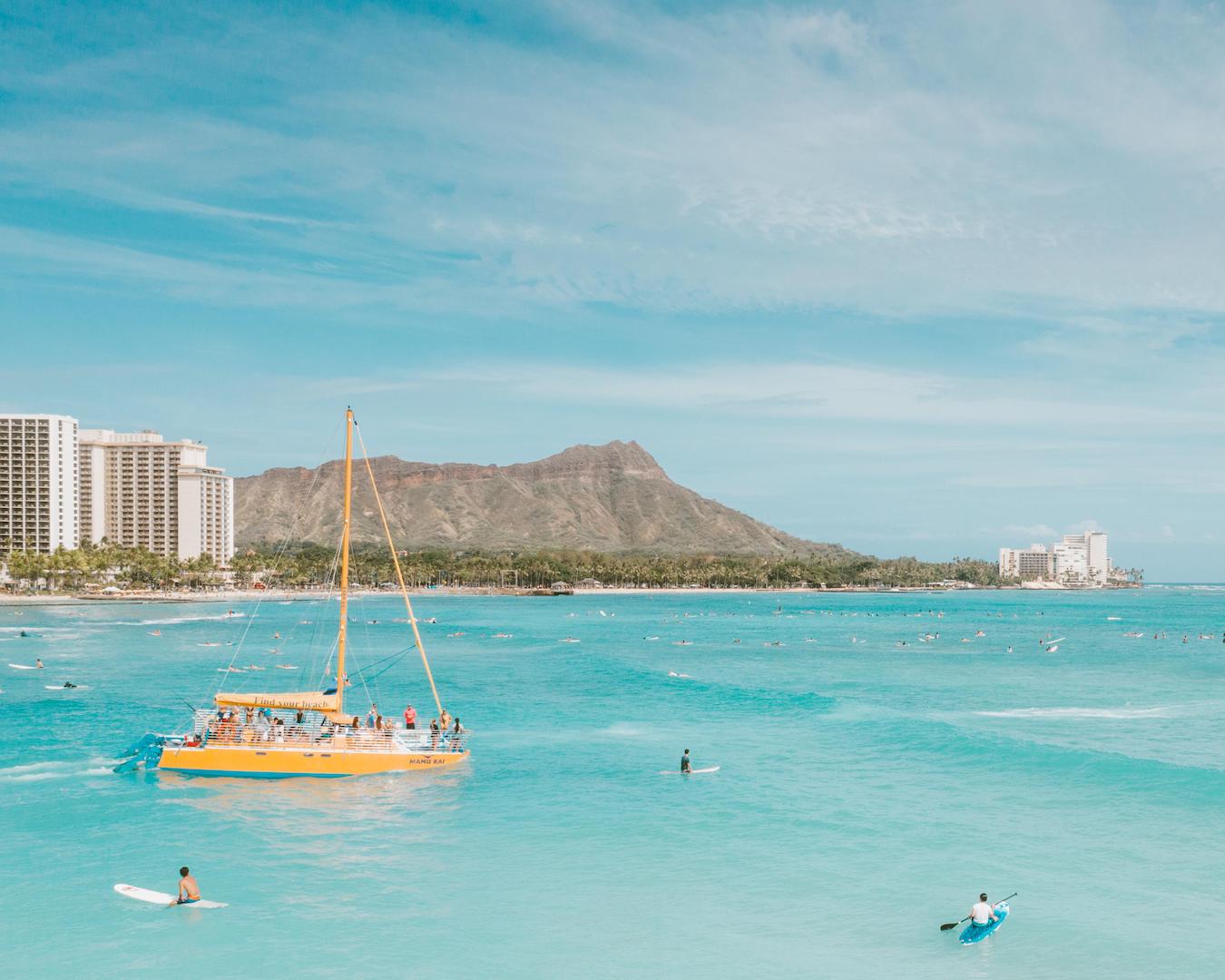}} \\
         SRC & RRC & SRC & RRC &  SRC & RRC\\
         \includegraphics[width = 0.14\linewidth]{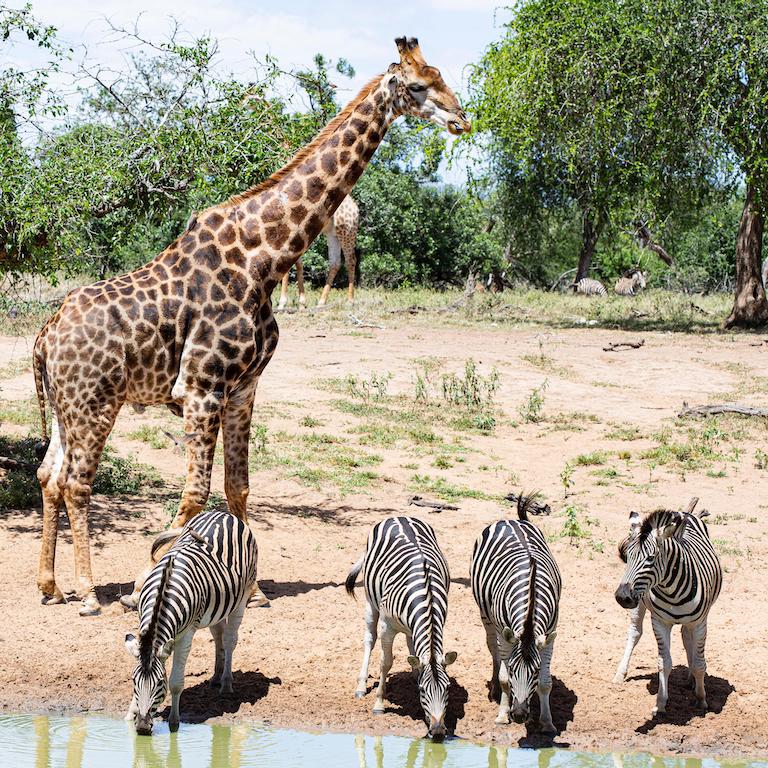} & %
         \includegraphics[width = 0.14\linewidth]{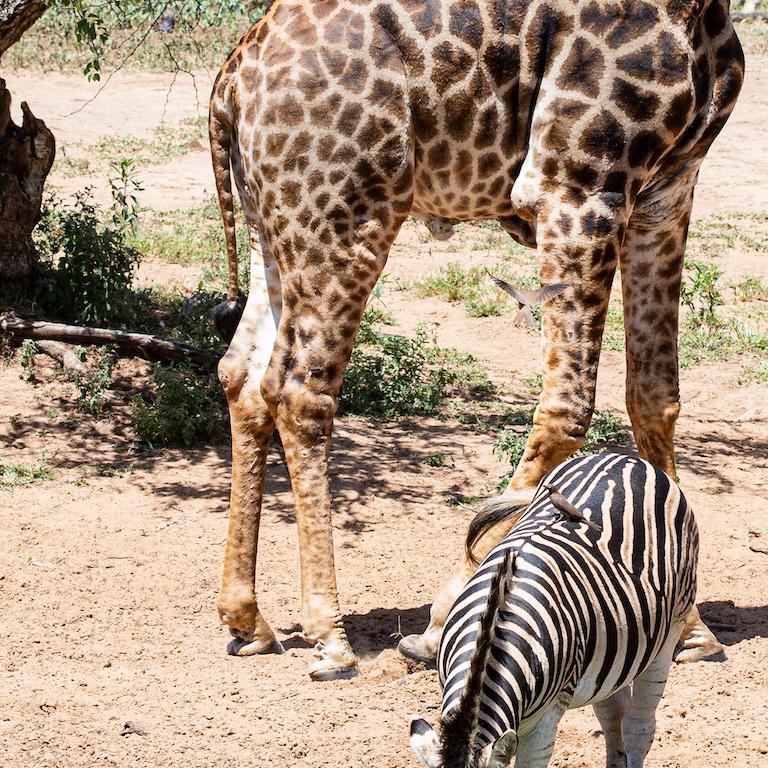} &
         \includegraphics[width = 0.14\linewidth]{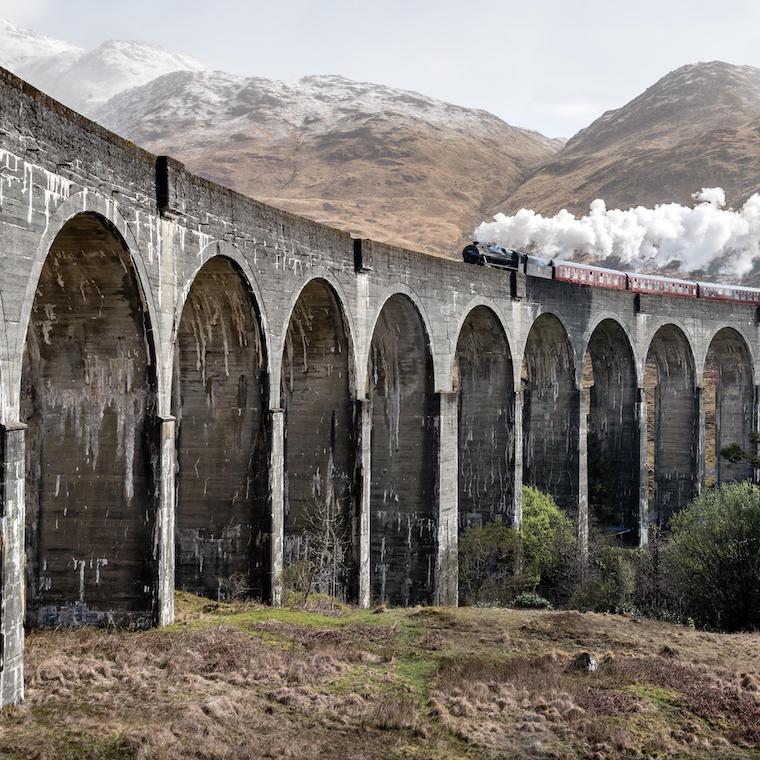} &
         \includegraphics[width = 0.14\linewidth]{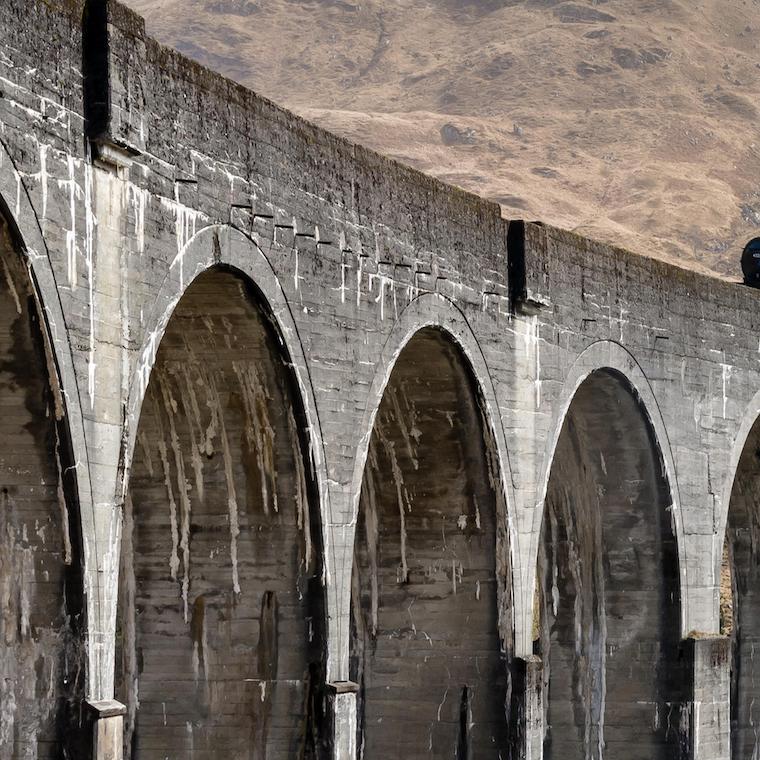} &
         \includegraphics[width = 0.14\linewidth]{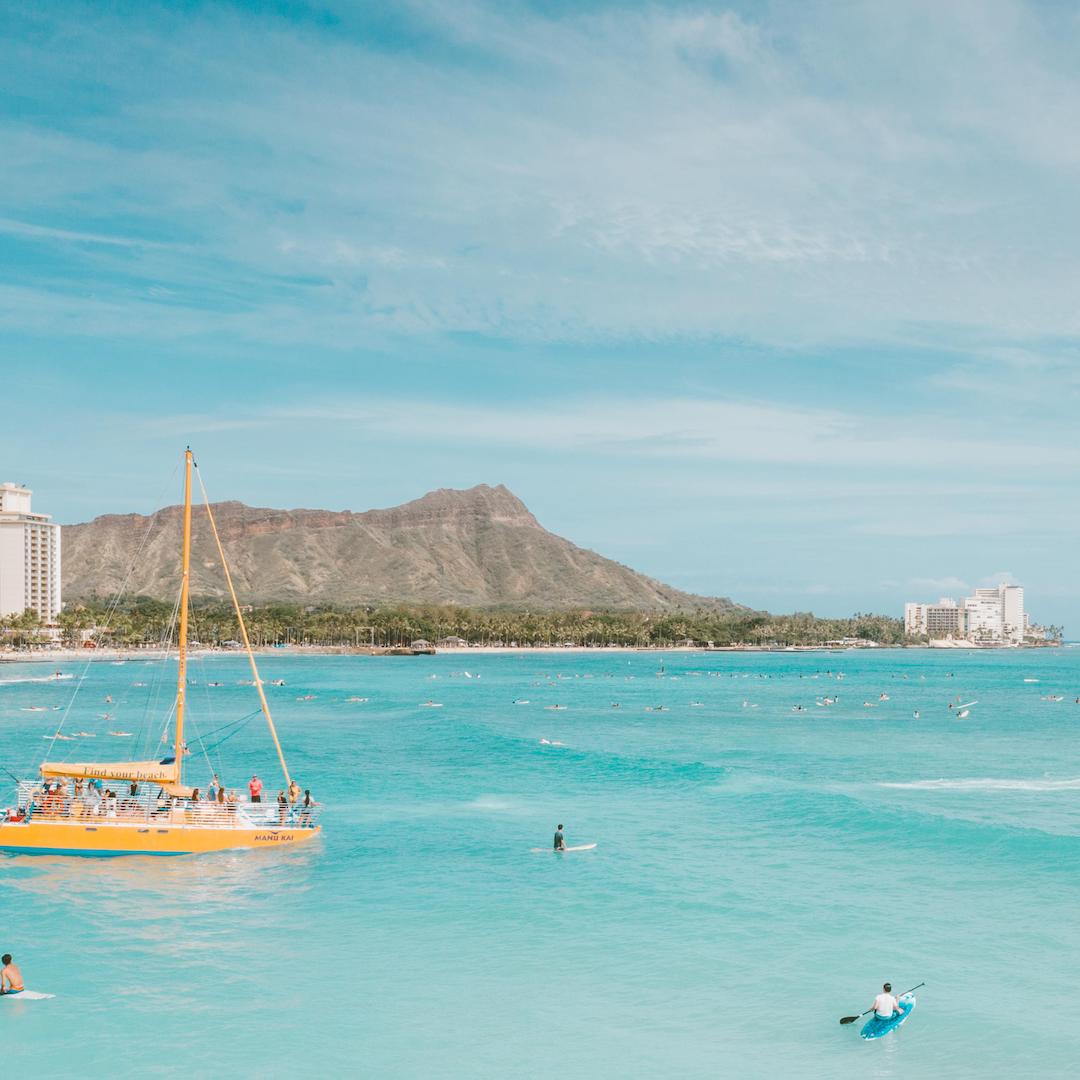} &
         \includegraphics[width = 0.14\linewidth]{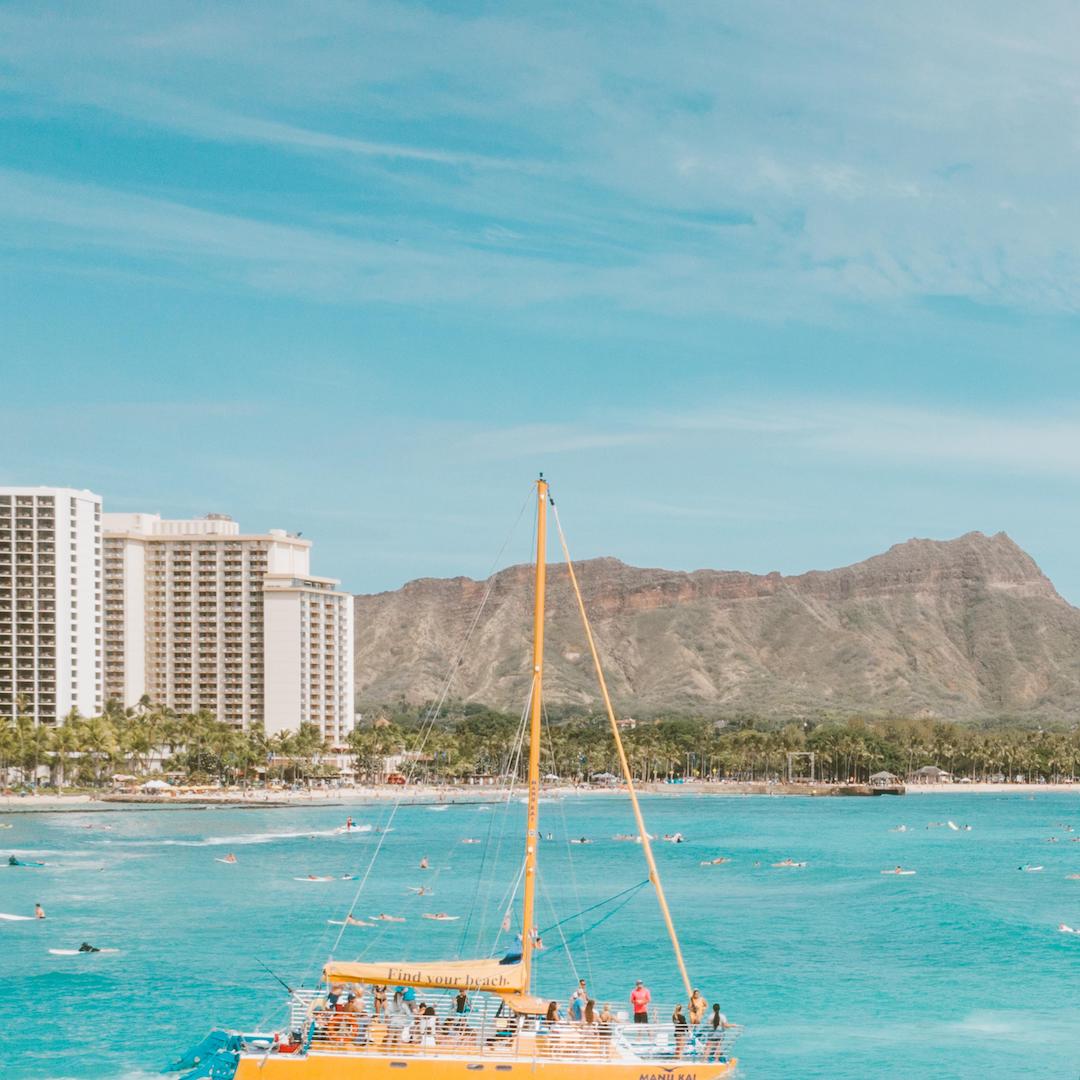} \\         
         
         \includegraphics[width = 0.14\linewidth]{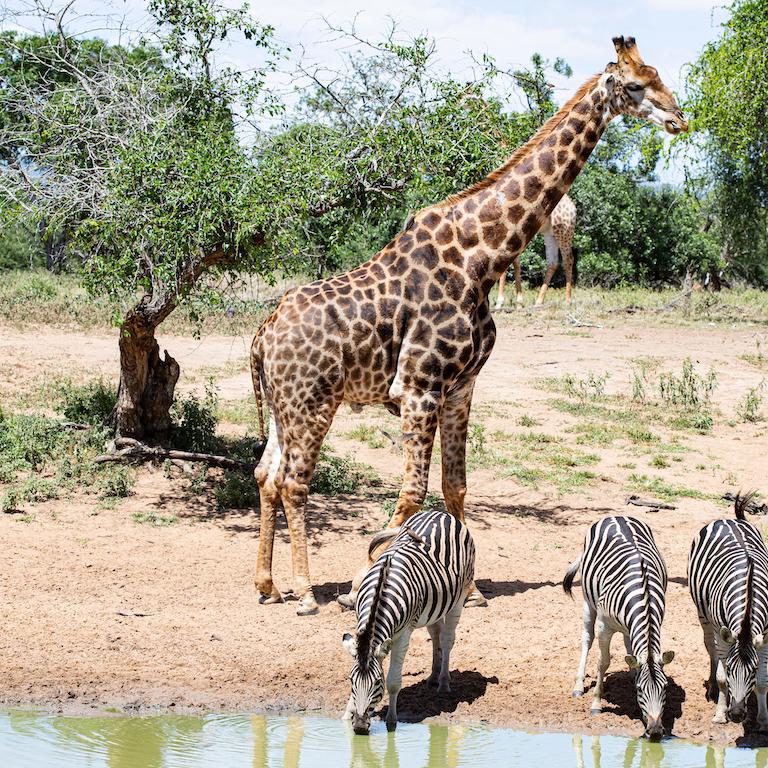} &
         \includegraphics[width = 0.14\linewidth]{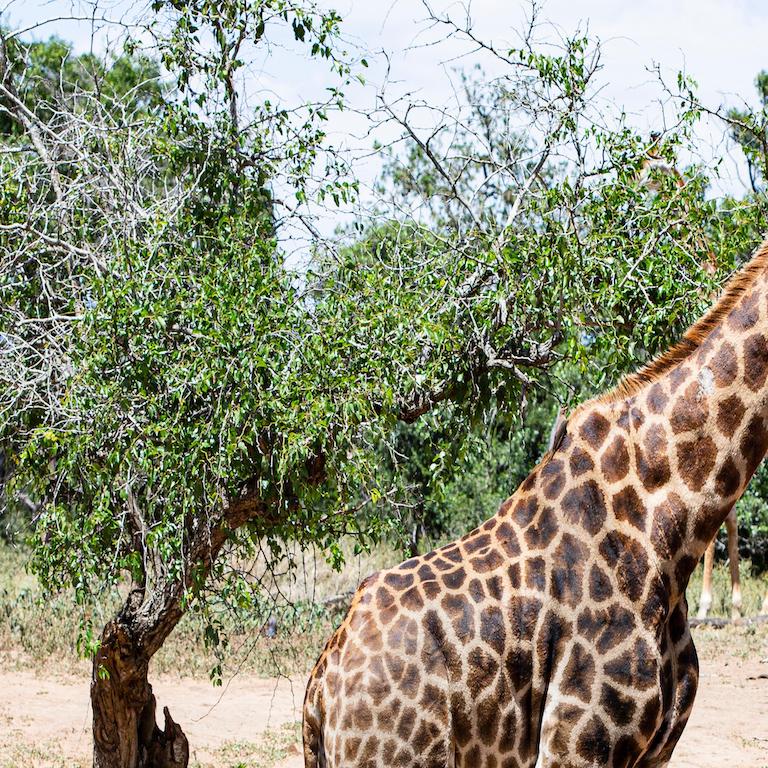} &
         \includegraphics[width = 0.14\linewidth]{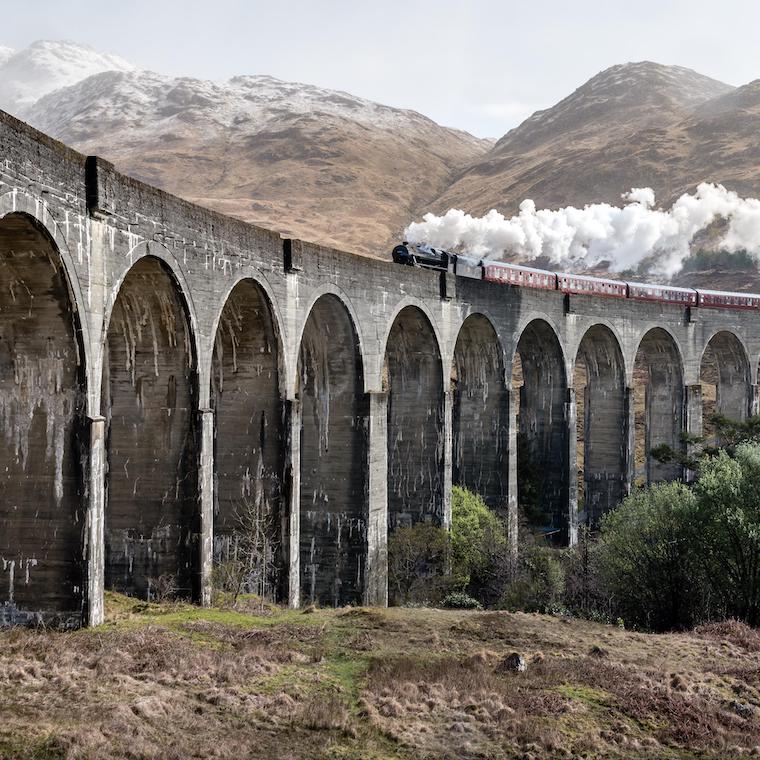} &
         \includegraphics[width = 0.14\linewidth]{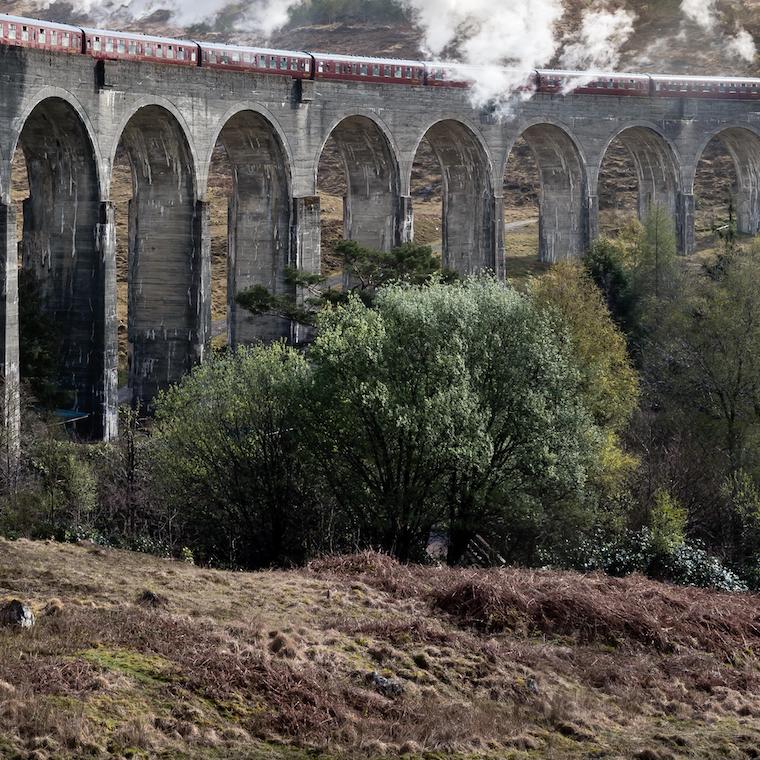} &
         \includegraphics[width = 0.14\linewidth]{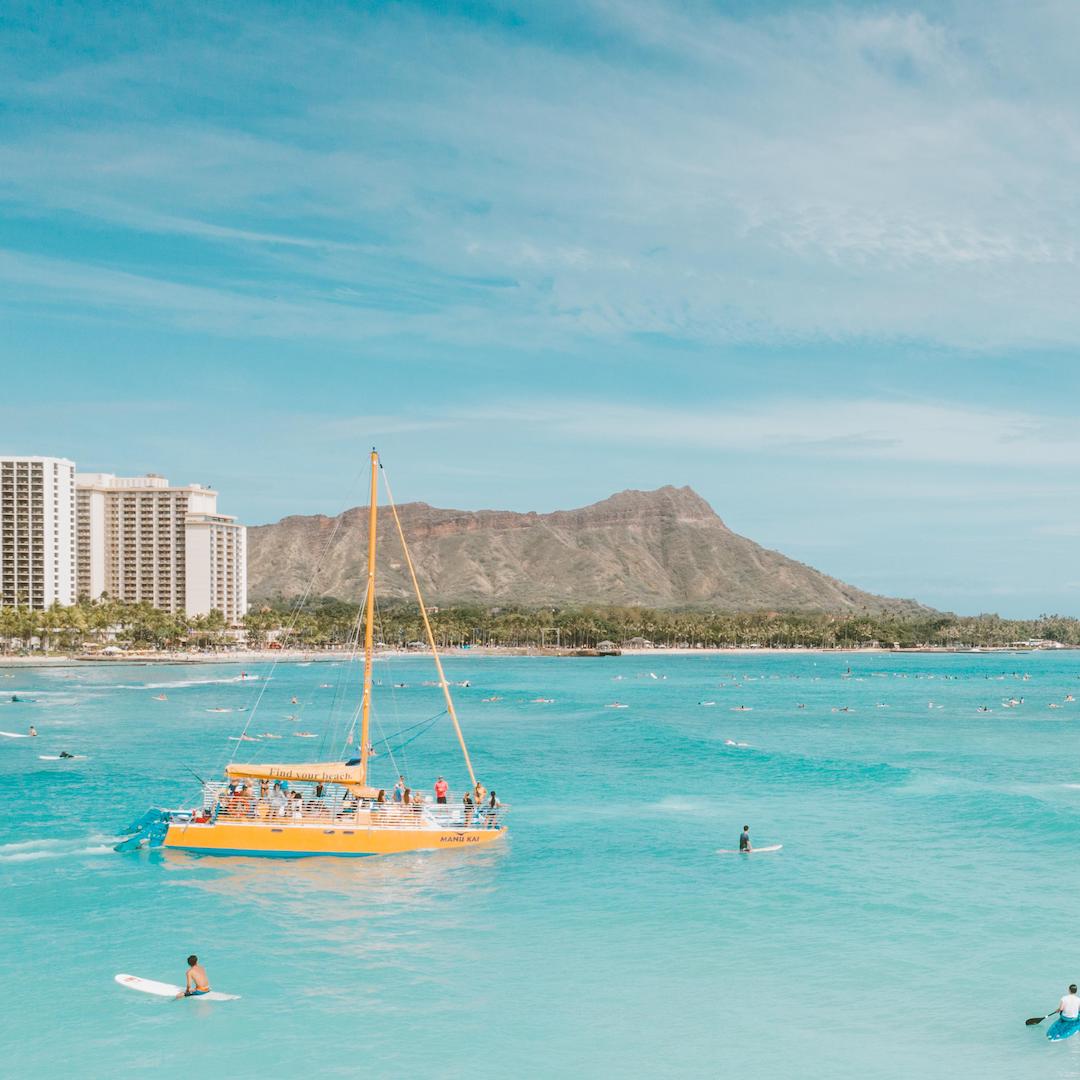} &
         \includegraphics[width = 0.14\linewidth]{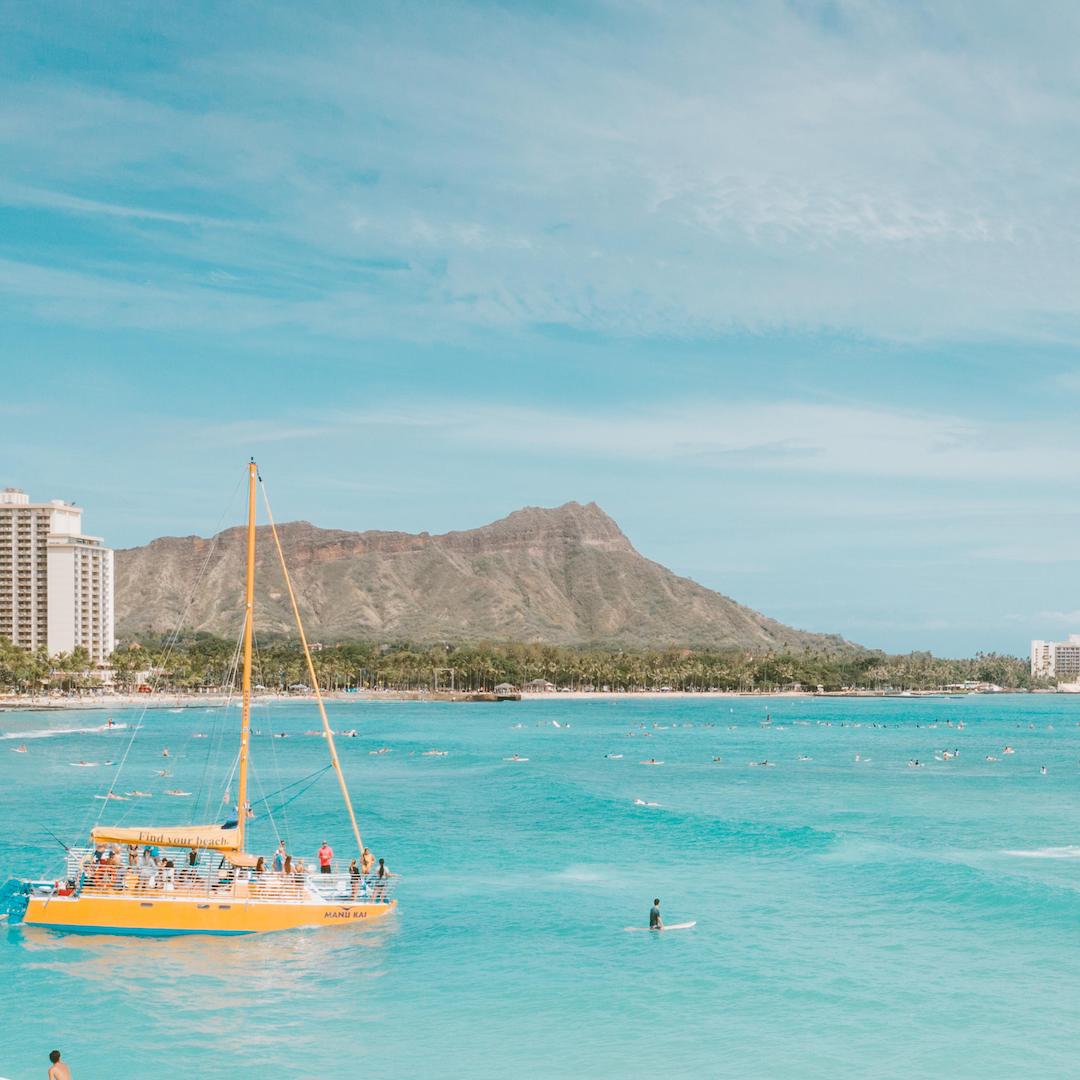} \\   
         
         \includegraphics[width = 0.14\linewidth]{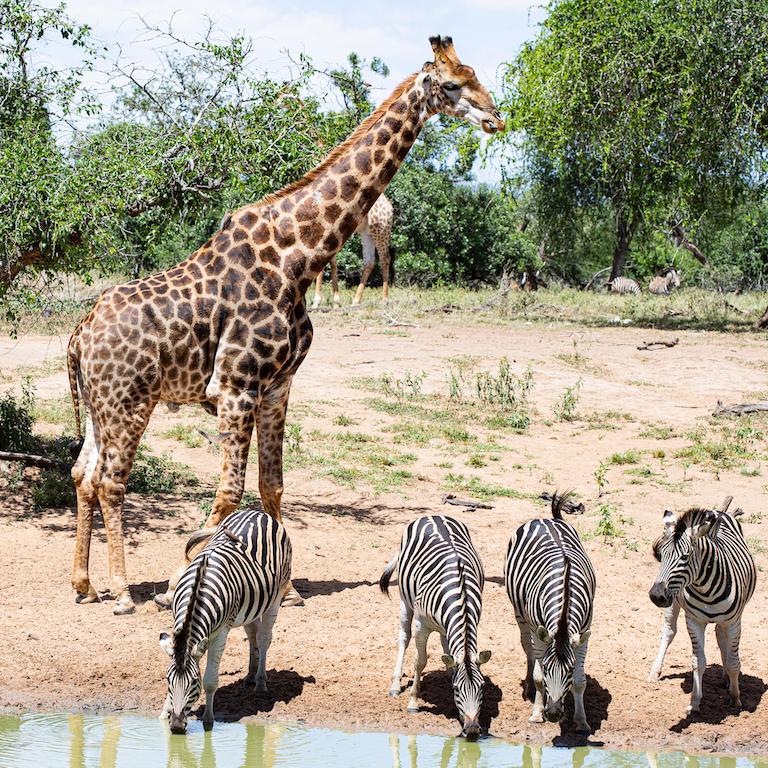} &
         \includegraphics[width = 0.14\linewidth]{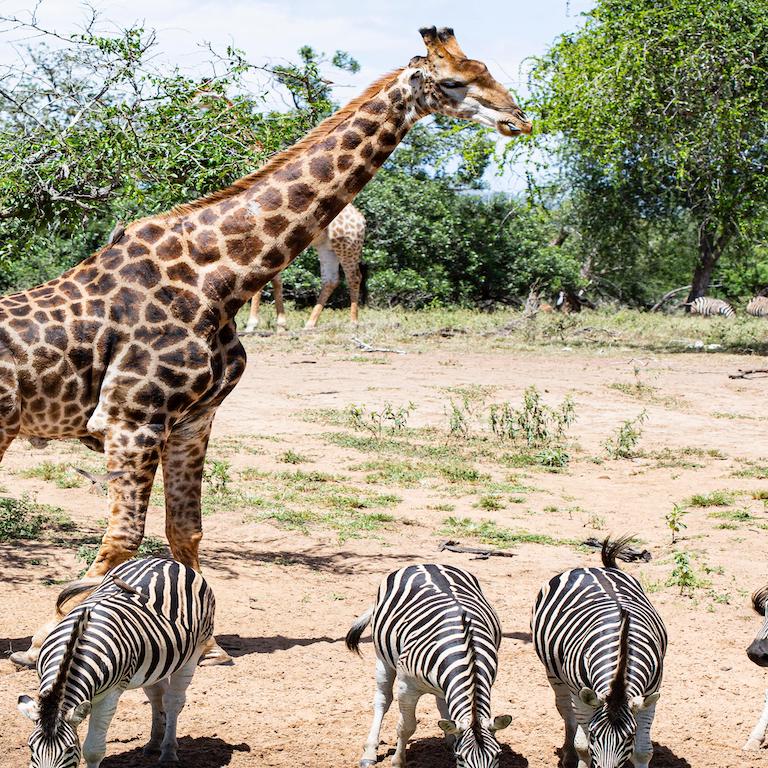} &
         \includegraphics[width = 0.14\linewidth]{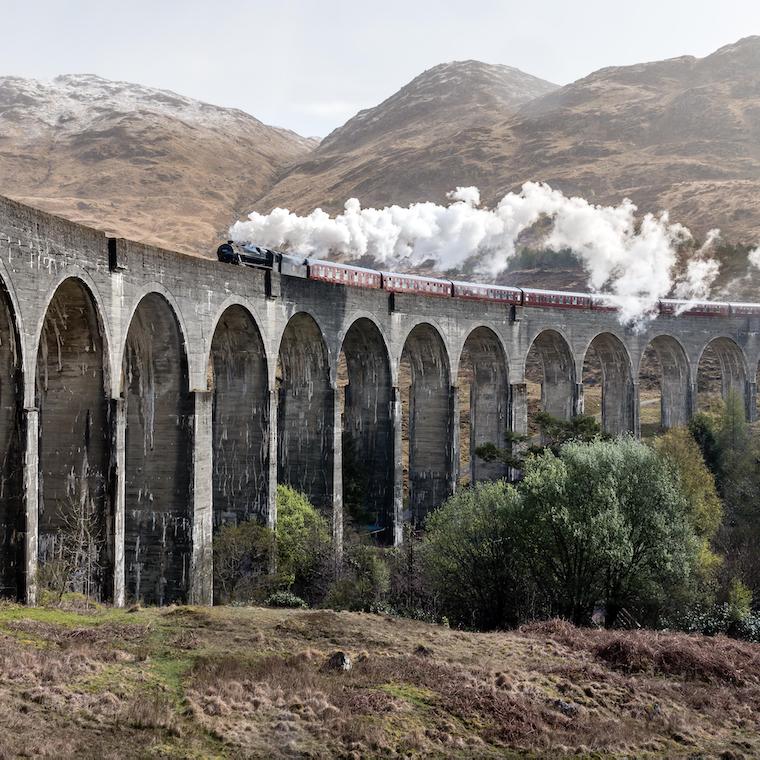} &
         \includegraphics[width = 0.14\linewidth]{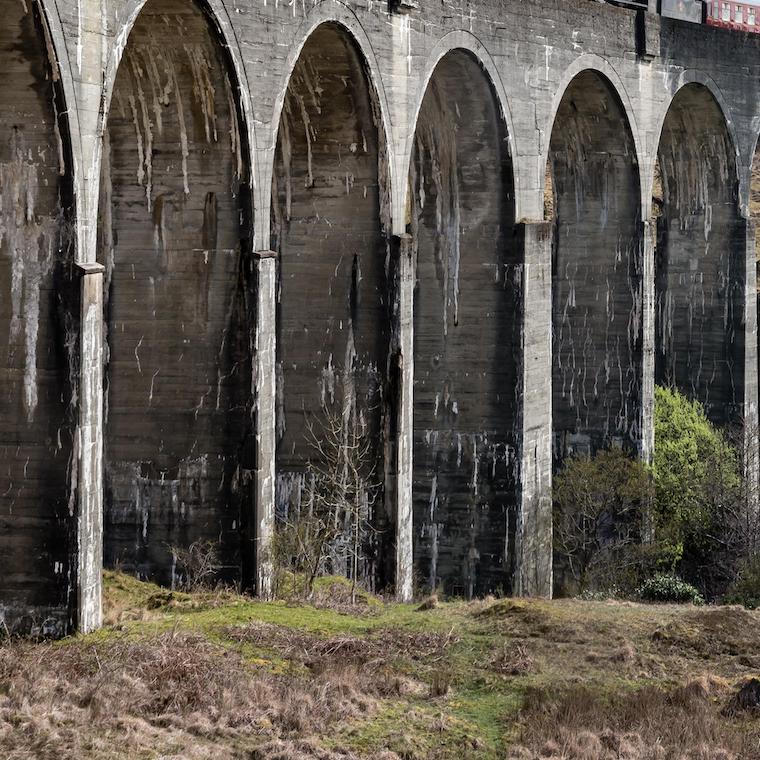} &
         \includegraphics[width = 0.14\linewidth]{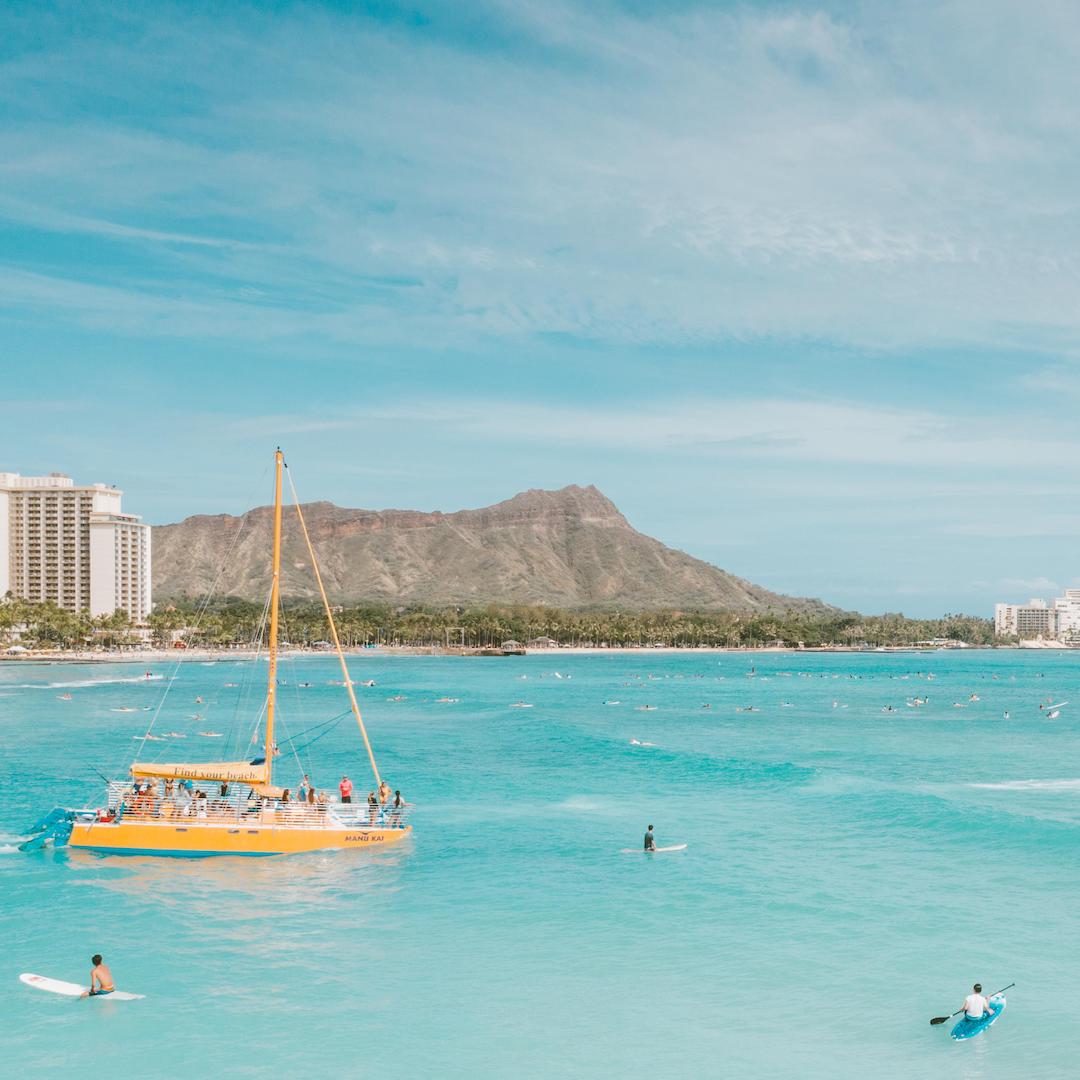} &
         \includegraphics[width = 0.14\linewidth]{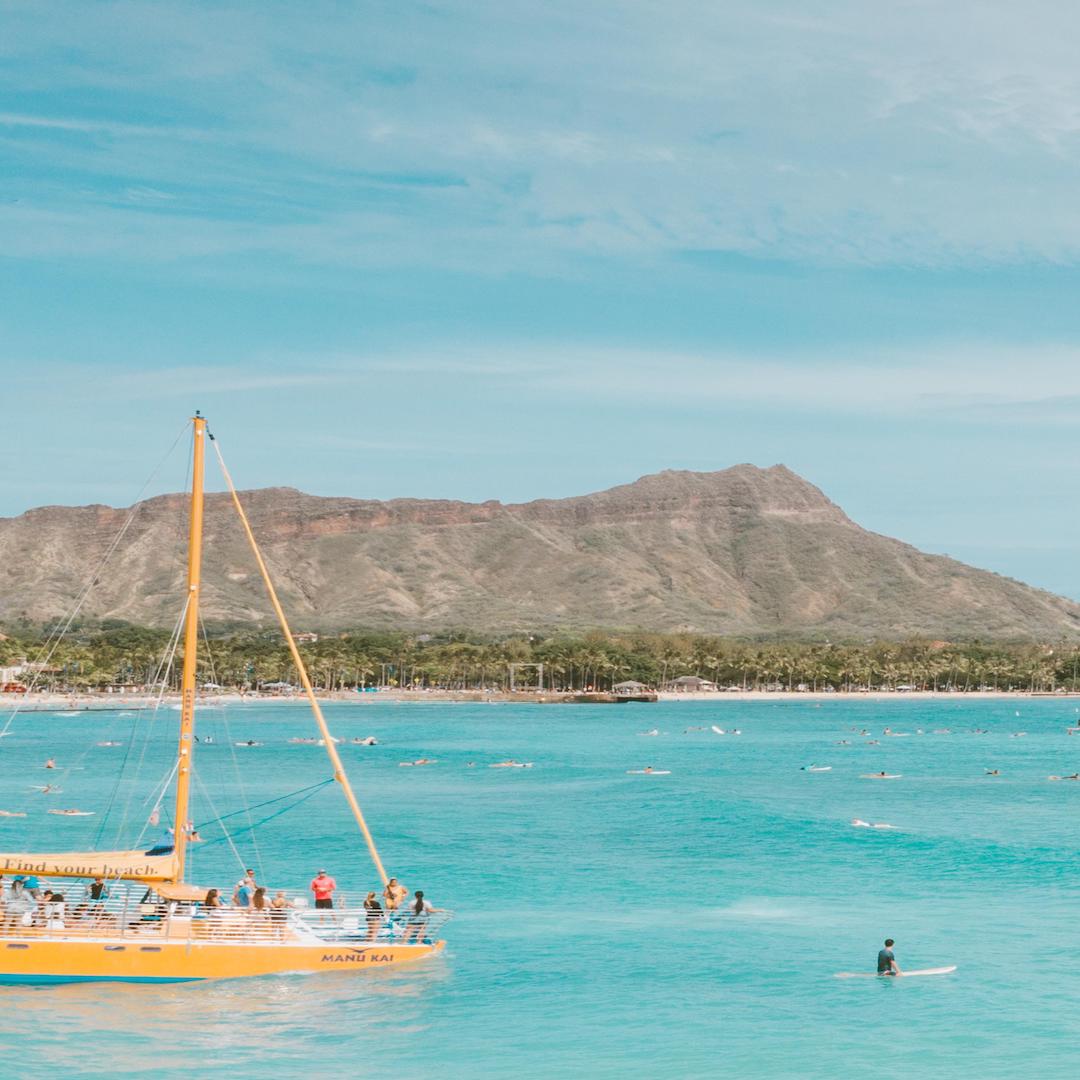} \\  
         
         \includegraphics[width = 0.14\linewidth]{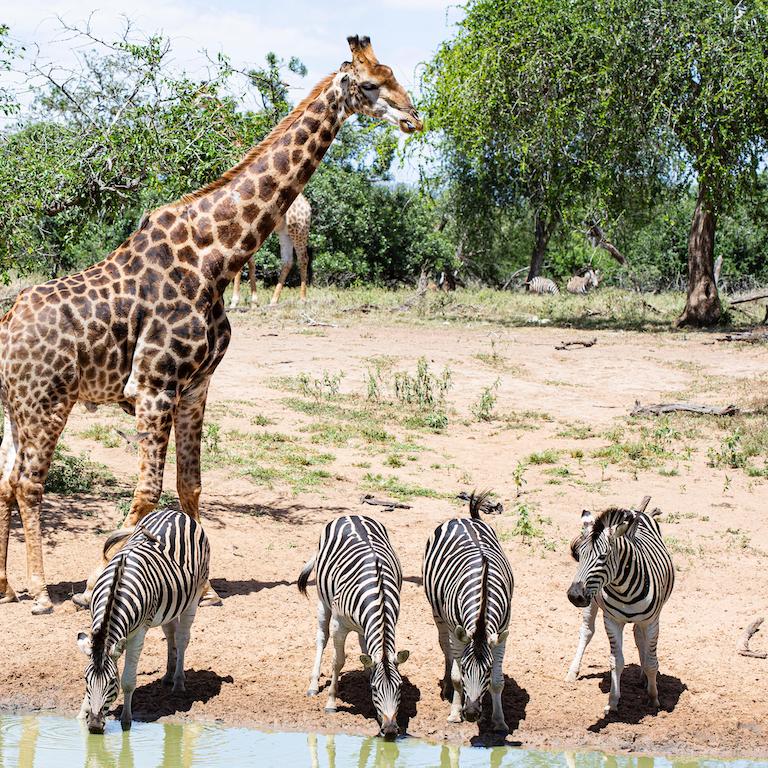} &
         \includegraphics[width = 0.14\linewidth]{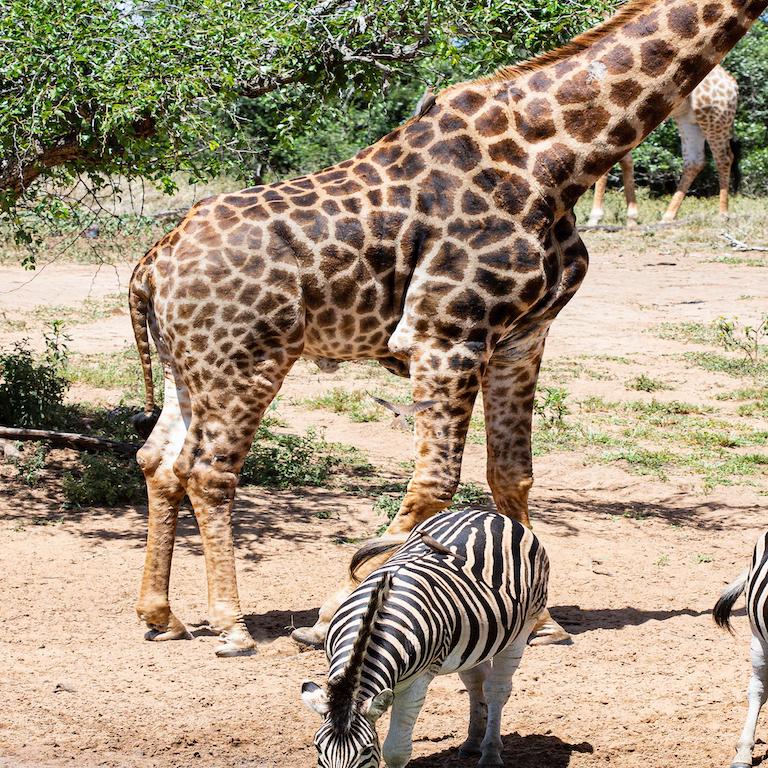} &
         \includegraphics[width = 0.14\linewidth]{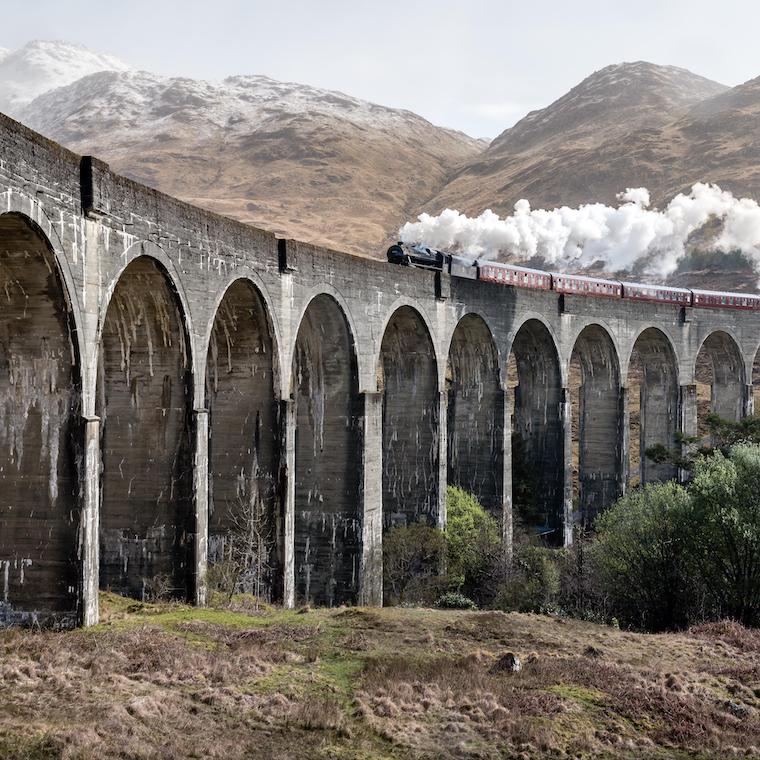} &
         \includegraphics[width = 0.14\linewidth]{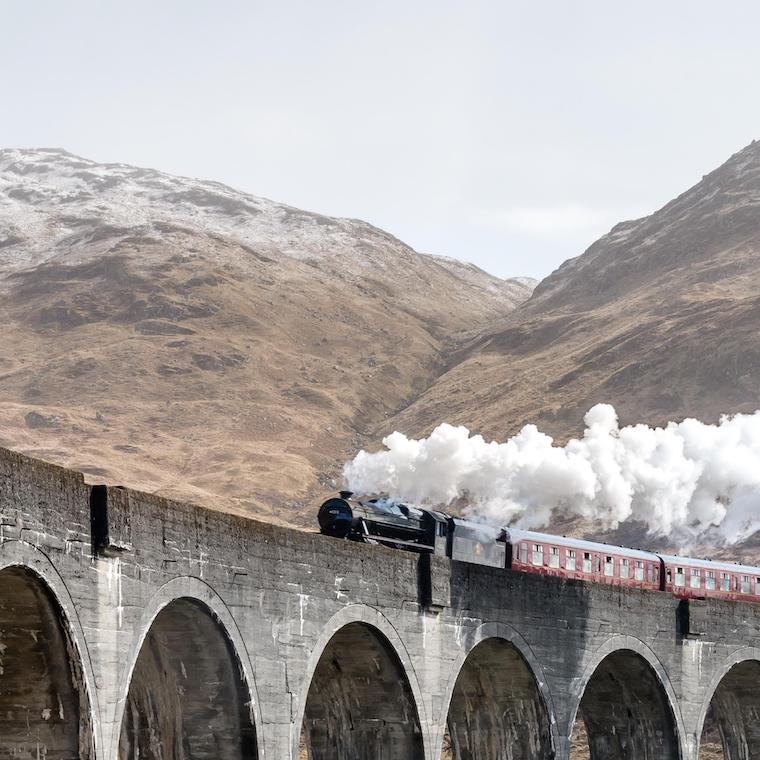} &
         \includegraphics[width = 0.14\linewidth]{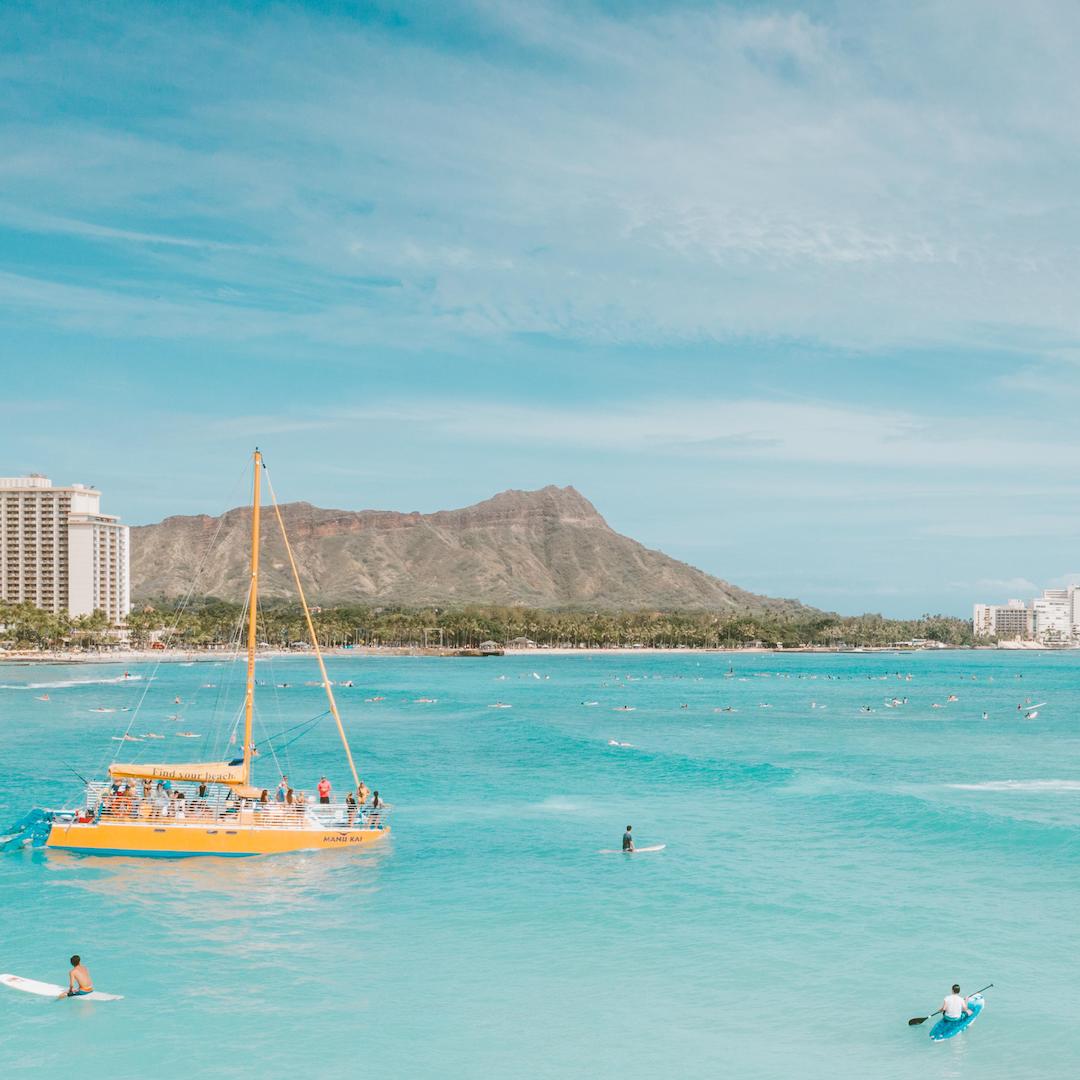} &
         \includegraphics[width = 0.14\linewidth]{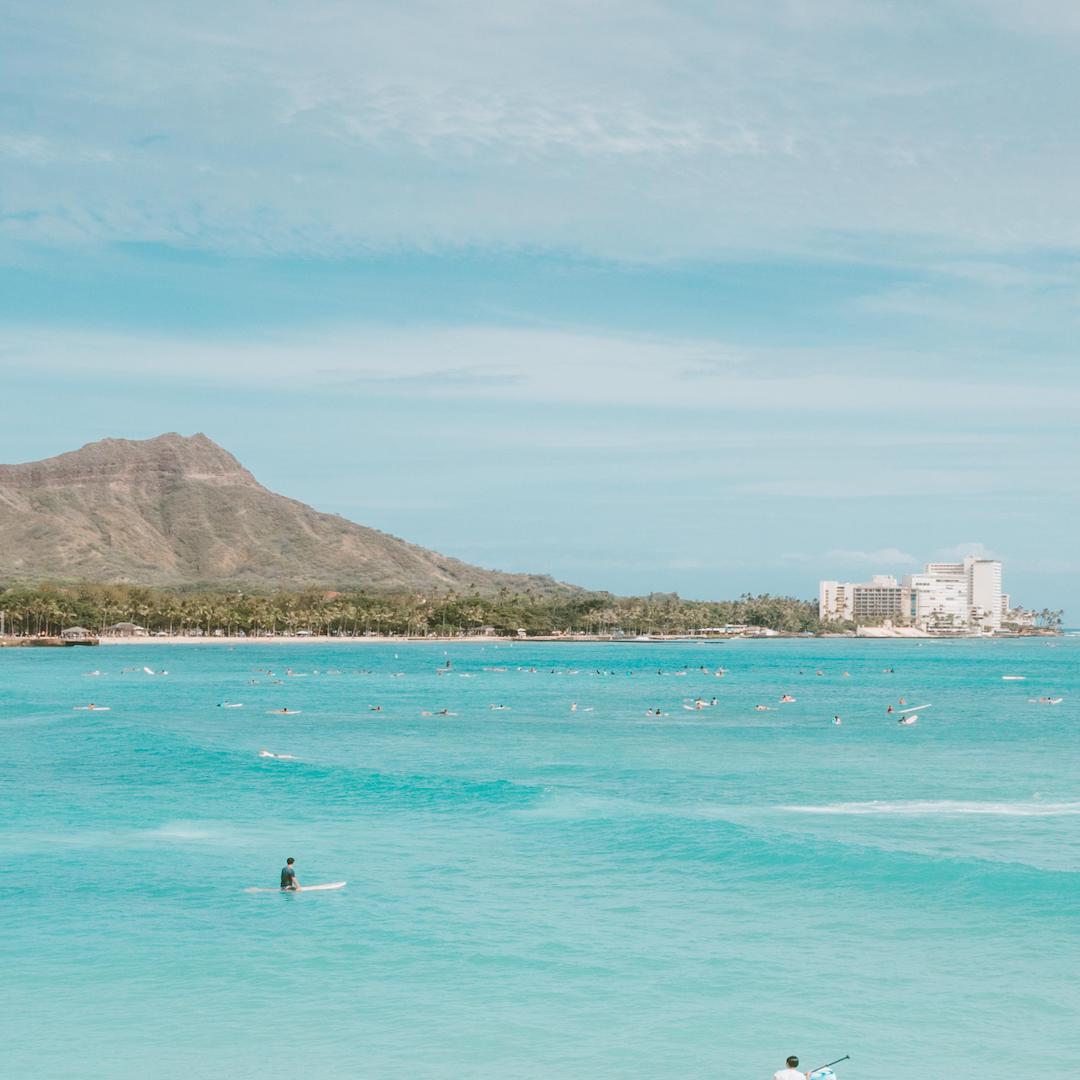} \\        
        
         \includegraphics[width = 0.14\linewidth]{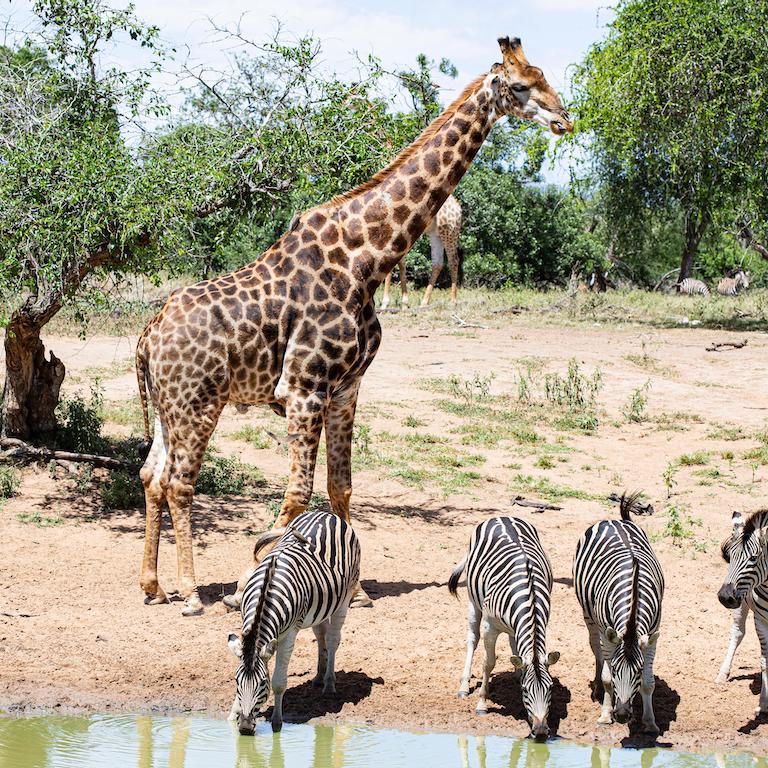} &
         \includegraphics[width = 0.14\linewidth]{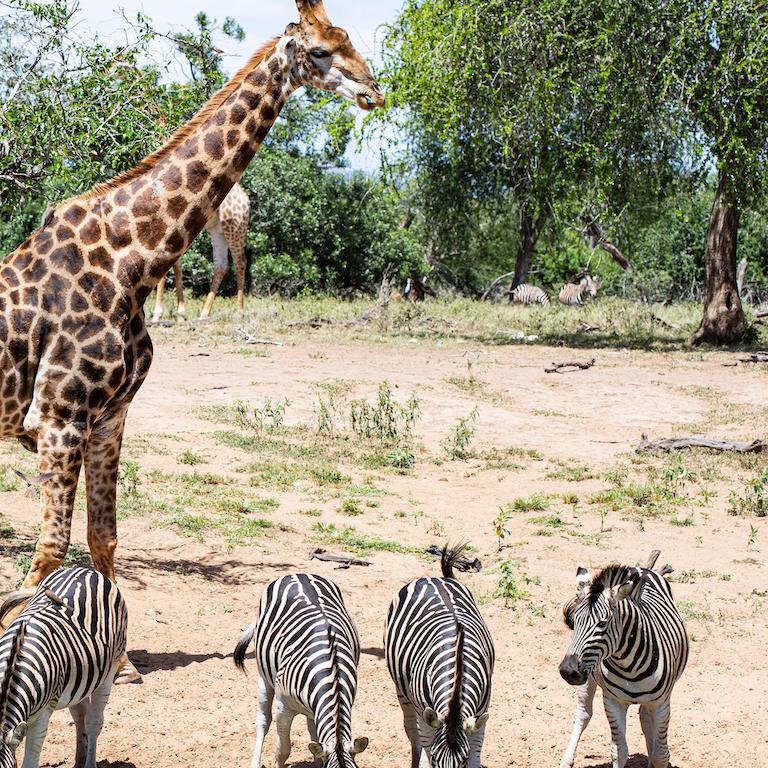} &
         \includegraphics[width = 0.14\linewidth]{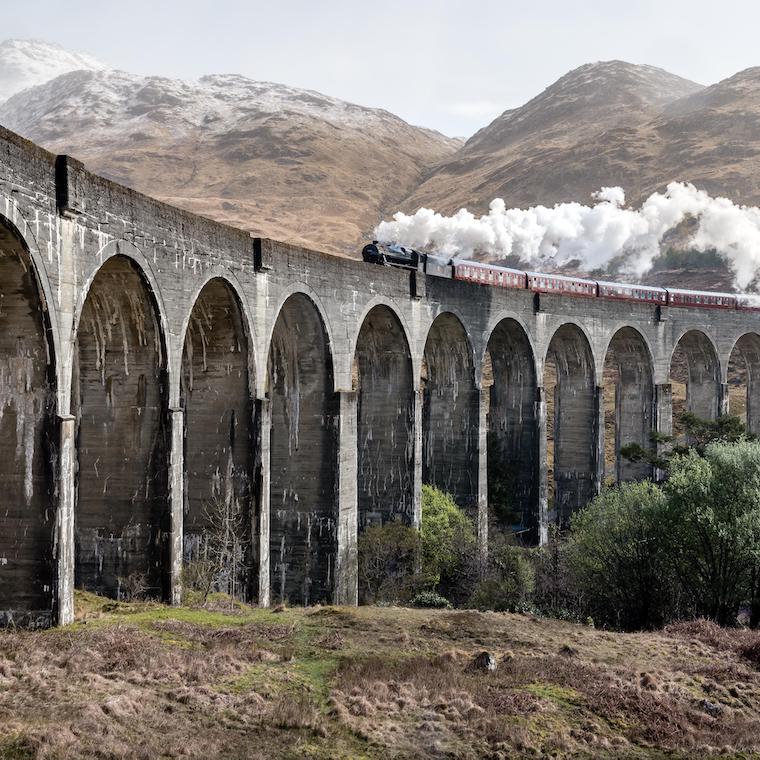} &
         \includegraphics[width = 0.14\linewidth]{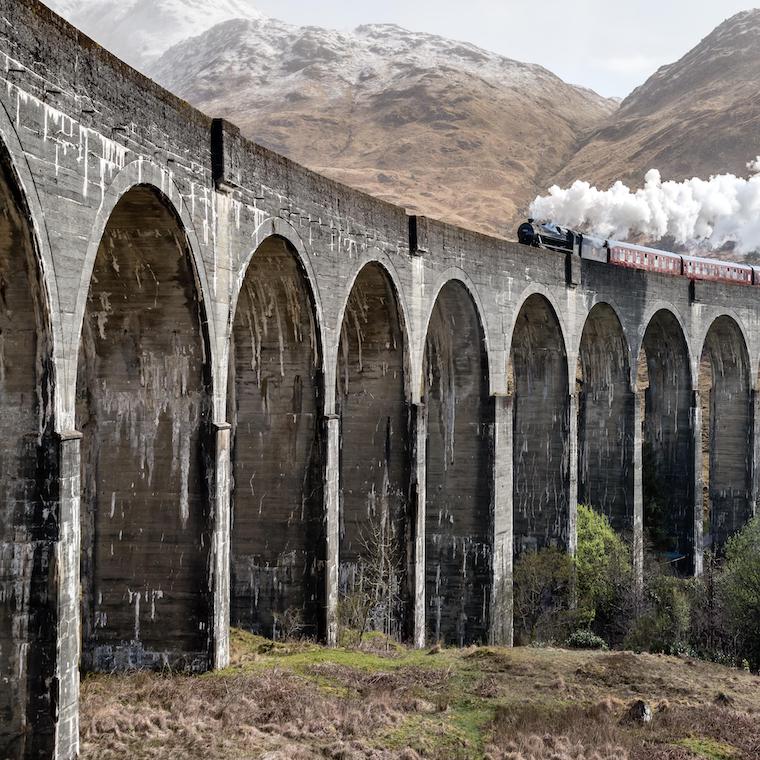} &
         \includegraphics[width = 0.14\linewidth]{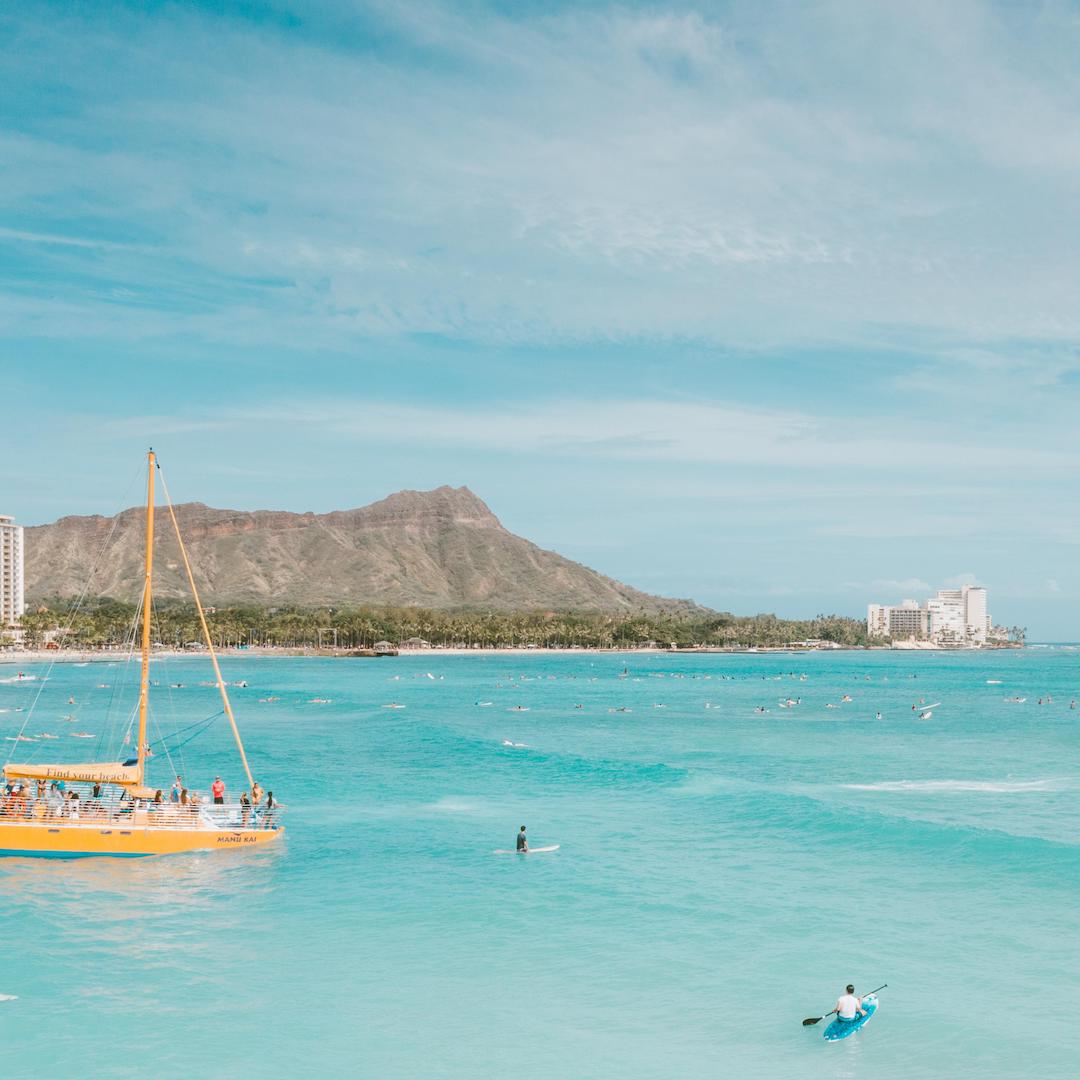} &
         \includegraphics[width = 0.14\linewidth]{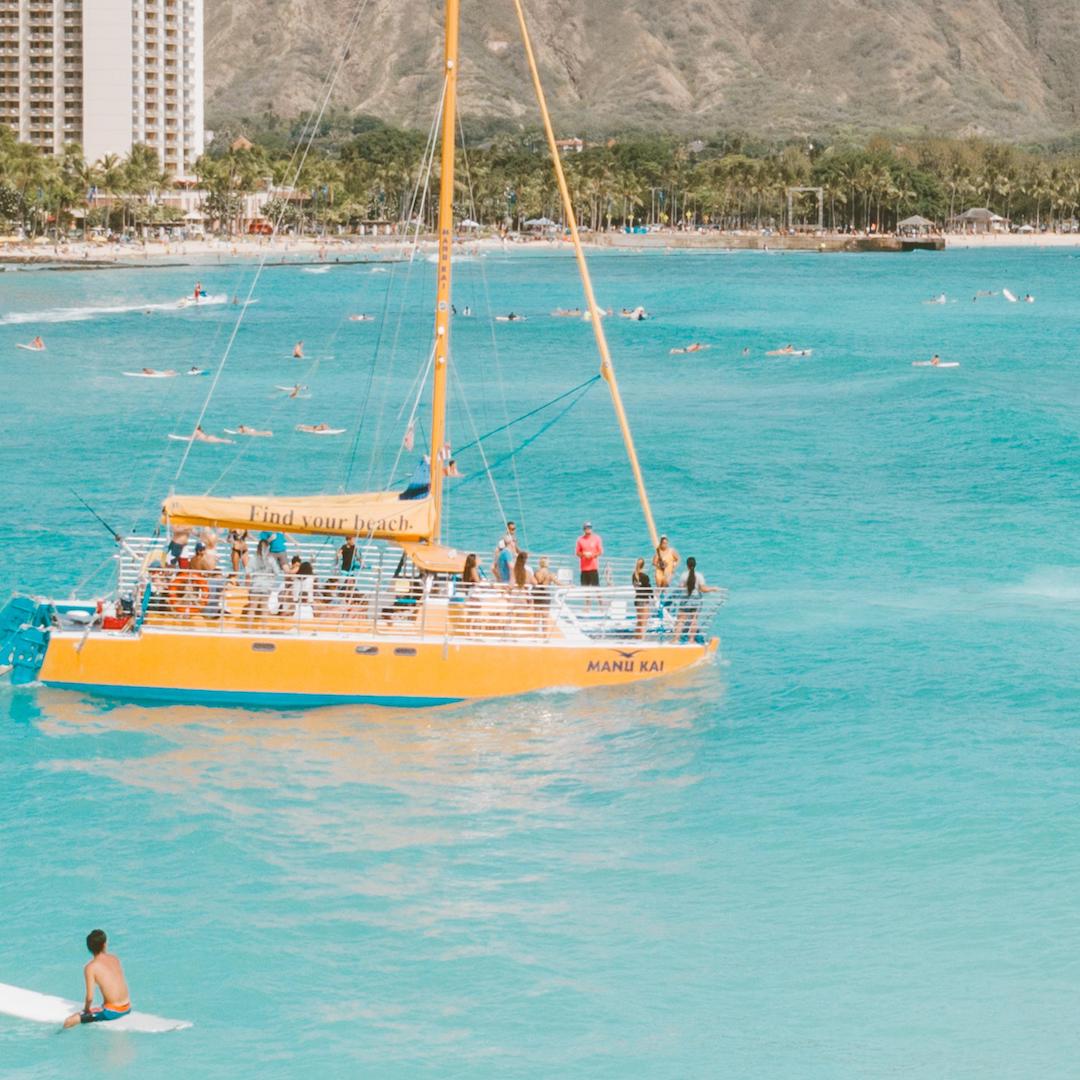} \\    
    \end{tabular}
    \caption{Illustration of Random Resized Crop (RRC) and Simple Random Crop (SRC). The usual RRC is a more aggressive data-augmentation than SRC: It has a more important regularizing effect and avoids overfitting by giving more variability to the images. At the same time it introduces a discrepancy of scale and aspect-ratio. It also leads to labeling errors, for instance when the object is not in the cropped region (e.g., train or boat). On Imagenet1k this regularization is overall regarded as beneficial. However our experiments show that it is detrimental on Imagenet21k, which is less prone to overfitting. 
    \label{fig:fig_rrc}}
\end{figure}

\paragraph{\textbf{Random Resized Crop (RRC)}} was introduced in the GoogleNet~\cite{Szegedy2015Goingdeeperwithconvolutions} paper. 
It serves as a regularisation to limit model overfitting,  while favoring that the decision done by the model is invariant to a certain class of transformations. 
This data augmentation was deemed important on Imagenet1k to prevent overfitting, which happens to occur rapidly with modern large models. 

This cropping strategy however introduces some discrepancy between train and test images in terms of the aspect ratio and the apparent size of objects~\cite{Touvron2019FixRes}.
Since ImageNet-21k includes significantly more images, it is less prone to overfitting.  Therefore we question whether the benefit of the strong RRC regularization compensates for its drawback when training on larger sets. 

\paragraph{\textbf{Simple Random Crop (SRC) }} is a much simpler way to extract crops. It is similar to the original cropping choice  proposed in AlexNet~\cite{Krizhevsky2012AlexNet}: We resize the image such that the smallest side matches the training resolution. Then we apply a reflect padding of 4 pixels on all sides, and finally we apply a square Crop of training size randomly selected along the x-axis of the image. 

Figure~\ref{fig:rrc_bbox} vizualizes  cropping boxes sampled for RRC and SRC. RRC provides a lot of diversity and very different sizes for crops. In contrast  SRC covers a much larger fraction of the image overall and preserve the aspect ratio, but offers less diversity: The crops overlaps significantly. 
As a result, when training on ImageNet-1k the performance is better with the commonly used RRC. For instance a ViT-S reduces its top-1 accuracy by $-0.9\%$ if we do not use RRC.

However, in the case of ImageNet-21k ($\times 10$ bigger than ImageNet-1k), there is less risk of overfitting and increasing the regularisation and diversity offered by RRC is less important. In this context, SRC offers the advantage of reducing the discrepancy in apparent size and aspect ratio.  More importantly, it gives a higher chance that the actual label of the image matches that of the crop: RRC is relatively aggressive in terms of cropping and in many cases the labelled object is not even present in the crop, as shown in Figure~\ref{fig:fig_rrc} where some of the crops do not contain the labelled object. For instance, with RRC there is a crop no zebra in the left example, or no train in three of the crops from the middle example. This is more unlikely to happen with SRC, which covers a much larger fraction of the image pixels. 
In Table~\ref{tab:rrc_ablation} we provide an ablation of random resized crop on ImageNet-21k, where we see that these observations translate as a significant gain in performance. %

\section{Experiments}
This section includes multiple experiments in image classification, with a special emphasis on Imagenet1k~\cite{deng2009imagenet,Recht2019ImageNetv2,Russakovsky2015ImageNet12}. We also report results for  downstream tasks in fine-grained classification and segmentation. 
We include a large number of ablations to better analyze different effects, such as the importance of the training resolution and longer training schedules. We provide additional results in the appendices. 

\subsection{Baselines and default settings}
 
The main task that we consider in this paper for the evaluation of our training procedure is image classification. 
We train on Imagenet1k-train and evaluate on Imagenet1k-val, with results on ImageNet-V2 to control  overfitting. 
We also consider the case where we can pretrain  on ImageNet-21k, 
Finally, we report transfer learning results on 6 different datasets/benchmarks. 

\paragraph{Default setting. }
When training on ImageNet-1k only,  by default we train during 400 epochs with a batch size 2048, following prior works~\cite{touvron2021going,xiao2021early}. 
Unless specified otherwise, both the training and evaluation are carried out at resolution $224 \times 224$ (even though we recommend to train at a lower resolution when targeting $224\times 224$ at inference time). 

When pre-training on ImageNet-21k, we pre-train by default during 90 epochs at resolution $224\times 224$, followed by a finetuning of 50 epochs on on ImageNet-1k. In this context we consider two fine-tuning resolutions: $224\times 224$ and $384\times 384$. 

\subsection{Ablations}

\subsubsection{Impact of training duration}

\begin{figure}
    \centering
    \begin{minipage}{0.48 \linewidth}
    \centering
    \includegraphics[width=0.85\linewidth]{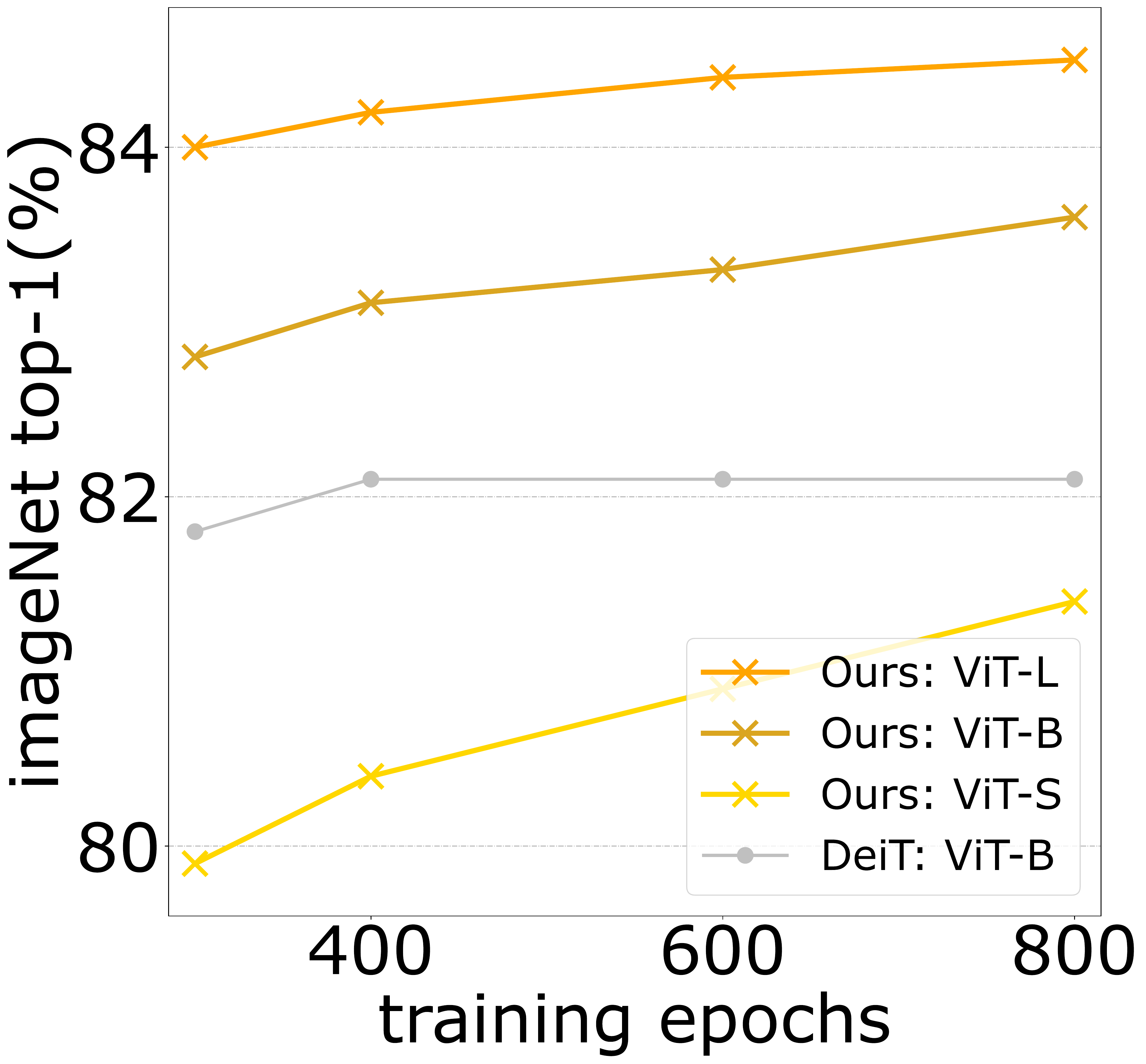}
    \caption{Top-1 accuracy on ImageNet-1k only at resolution $224\times 224$ with our training recipes and a different number of epochs %
    \label{fig:epochs}}
    \end{minipage}
    \hfill
    \begin{minipage}{0.49 \linewidth}
    \centering
    \includegraphics[width=0.85\linewidth]{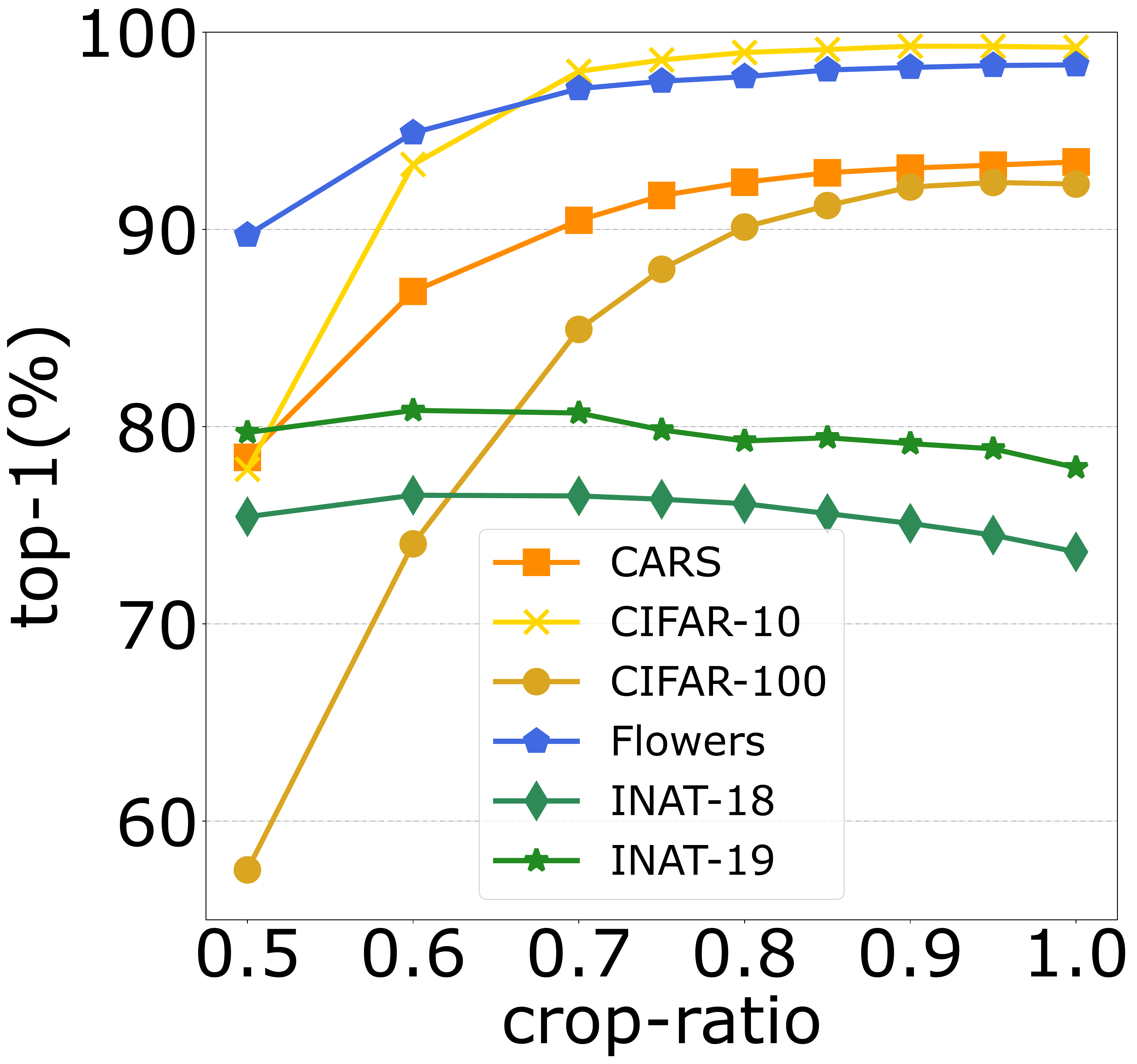}
    \caption{Transfer learning performance on 6 datasets with different test-time crop ratio. ViT-B pre-trained on ImageNet-1k at resolution $224$. 
    \label{fig:crop}}
    \end{minipage}
\end{figure}

In Figure~\ref{fig:epochs} we provide an ablation on the number of epochs, which show that ViT models do not saturate as rapidly as the DeiT training procedure~\cite{Touvron2020TrainingDI} when we increase the number of epochs beyond the 400 epochs adopted for our baseline.

For ImageNet-21k pre-training, we use 90 epochs for pre-training as in a few works~\cite{liu2021swin,Touvron2021AugmentingCN}. We finetune during 50 epochs on ImageNet-1k~\cite{Touvron2021AugmentingCN} and marginally adapt the stochastic depth parameter. We point out that this choice is mostly for the sake of consistency across models: we observe that training 30 epochs also provides similar results. 

\subsubsection{Data-Augmentation}
In Table~\ref{tab:da_comp} we compare our handcrafted data-augmentation 3-Augment with existing learned augmentation methods. 
With the ViT architecture, our data-augmentation is the most effective while being simpler than the other approaches. 
Since previous augmentations were introduced on convnets, we also provide results for a ResNet-50. In this case previous augmentation policies have similar (RandAugment, Trivial-Augment) or better results (Auto-Augment) on the validation set. This is no longer the case when evaluating on the independent set V2, for which the Auto-Augment better accuracy is not significant. 

\begin{table}
    \centering
    \scalebox{0.8}{
    \begin{tabular}{c|cc|c|ccc}
    \toprule
        \multirow{2}{*}{Method} & Learned augm. &  \# Nb of   &  \multirow{2}{*}{Model} & \multicolumn{3}{c}{ImageNet-1k} \\
        & methods & DA & & Val & Real & V2\\
        \midrule
         \multirow{3}{*}{Auto-Augment~\cite{Ekin2018AutoAugment}} & \multirow{3}{*}{\cmark} & \multirow{3}{*}{14}
         & \grtext{ResNet50}   & \grtext{\textbf{79.7}} & \grtext{\textbf{85.6}} & \grtext{\textbf{67.9}}\\
         & &  & ViT-B & 82.8 & 87.5 & 71.9\\
         & &  & ViT-L & 84.0 & \textbf{88.6} & 74.0 \\
         \midrule
        \multirow{3}{*}{RandAugment~\cite{Cubuk2019RandAugmentPA}}  & \multirow{3}{*}{\cmark} & \multirow{3}{*}{14}
        & \grtext{ResNet50}   & \grtext{79.5}  & \grtext{85.5} & \grtext{67.6} \\
        & &  & ViT-B &  82.7 & 87.4 & 72.2 \\
        &  & &  ViT-L & 84.0 & 88.3 & 73.8 \\
        \midrule
        \multirow{3}{*}{Trivial-Augment~\cite{Mller2021TrivialAugmentTY}} & \multirow{3}{*}{\cmark} & \multirow{3}{*}{14} 
        & \grtext{ResNet50}    & \grtext{79.5}  & \grtext{85.4} & \grtext{67.6} \\
        & &  & ViT-B  & 82.3 & 87.0 & 71.2 \\
        &  & & ViT-L & 83.6 & 88.1 & 73.7 \\
        \midrule
        \rowcolor{Goldenrod}
         &  & & \grtext{ResNet50} & \grtext{79.4}  & \grtext{85.5} & \grtext{67.8} \\
        \rowcolor{Goldenrod}
        &  & &ViT-B & \textbf{83.1} & \textbf{87.7} & \textbf{72.6} \\
        \rowcolor{Goldenrod}
        \multirow{-3}{*}{3-Augment (Ours)}  & \multirow{-3}{*}{\xmark} & \multirow{-3}{*}{3} &  ViT-L & \textbf{84.2} & \textbf{88.6} & \textbf{74.3} \\

    \bottomrule
    \end{tabular}}
    \caption{Comparison of some existing data-augmentation methods with our simple 3-Augment proposal inspired by data-augmentation used with self-supervised learning.
    \label{tab:da_comp}}
\end{table}

\begin{table}
    \centering
    \scalebox{0.8}{
    \begin{tabular}{c|cccc|ccc}
    \toprule
         \multirow{2}{*}{Model} & \multirow{2}{*}{Loss} & \multirow{2}{*}{LayerScale} &\multirow{2}{*}{Data Aug.} &\multirow{2}{*}{Epochs} & \multicolumn{3}{c}{ImageNet-1k} \\
         &  & &  &  & val & real & v2 \\
        \midrule
         \multirow{5}{*}{\rotatebox{90}{ViT-S}}
         & CE & \xmarkg & RandAugment & 300 & 79.8 & 85.3 & 68.1\\
         & BCE & \xmarkg & \textcolor{lightgray}{RandAugment} & \textcolor{lightgray}{300} & 79.8 & 85.9 & 68.2 \\
          & \textcolor{lightgray}{BCE} & \cmark & \textcolor{lightgray}{RandAugment} & \textcolor{lightgray}{300} &  80.1 & \textbf{86.1} & 69.1\\
          & \textcolor{lightgray}{BCE} & \cmarkg & \textcolor{lightgray}{RandAugment} & 400 & \textbf{80.7} & 86.0 & 69.3\\
          \rowcolor{Goldenrod}
          & \textcolor{lightgray}{BCE} & \cmarkg & 3-Augment & \textcolor{lightgray}{400} & 80.4 & \textbf{86.1} & \textbf{69.7}\\         
         
         \midrule
         \multirow{5}{*}{\rotatebox{90}{ViT-B}}
         & CE & \xmarkg & RandAugment & 300 & 80.9 & 85.5 & 68.5\\
         & BCE & \xmarkg & \textcolor{lightgray}{RandAugment} & \textcolor{lightgray}{300} & 82.2 & 87.2 & 71.4\\
          & \textcolor{lightgray}{BCE} & \cmark & \textcolor{lightgray}{RandAugment} & \textcolor{lightgray}{300} & 82.5 & 87.5 & 71.4 \\
          & \textcolor{lightgray}{BCE} & \cmarkg & \textcolor{lightgray}{RandAugment} & 400 & 82.7 & 87.4 & 72.2 \\
          \rowcolor{Goldenrod}
          & \textcolor{lightgray}{BCE} & \cmarkg & 3-Augment & \textcolor{lightgray}{400} & \textbf{83.1} & \textbf{87.7} & \textbf{72.6} \\
          \midrule
          \multirow{4}{*}{\rotatebox{90}{ViT-L}}
          & BCE & \xmarkg & RandAugment & 300 & 83.0 & 87.9 & 72.4\\
          & \textcolor{lightgray}{BCE} & \xmarkg & \textcolor{lightgray}{RandAugment} & 400 & 83.3 & 87.7 & 72.5\\
          & \textcolor{lightgray}{BCE} & \cmark & \textcolor{lightgray}{RandAugment} & \textcolor{lightgray}{400} & 84.0 & 88.3 & 73.8\\
          \rowcolor{Goldenrod}
           & \textcolor{lightgray}{BCE} & \cmarkg & 3-Augment  & \textcolor{lightgray}{400} & \textbf{84.2} & \textbf{88.6} & \textbf{74.3}\\
           \bottomrule
          
    \end{tabular}}
    \caption{Ablation on different training component with training at resolution $224 \times 224$ on ImageNet-1k. We perform avlations with ViT-S, ViT-B and ViT-L. We report top-1 accuracy (\%) on ImageNet validation set , ImageNet real and ImageNet v2. 
    \label{tab:ablation}}
\end{table}

\begin{table}
    \centering
    \scalebox{0.75}{
    \begin{tabular}{@{\ }ccccccccc|cccc@{\ }}
    \toprule
         \multirow{2}{*}{Crop.} & \multirow{2}{*}{LS} &  \multirow{2}{*}{Mixup} &  Aug. & \#Imnet21k & finetuning & \multicolumn{3}{c}{Imagenet-1k val top-1 }  & \multicolumn{3}{c}{Imagenet-1k v2 top-1 }\\
         & & & policy & epochs & resolution & ViT-S & ViT-B & ViT-L  &  ViT-S & ViT-B & ViT-L \\
         \midrule
         RRC                   & \xmark  & 0.8                      & RA  & \pzo 90                    & 224$^2$                   & 81.6 & 84.6 & 86.0 & 70.7 & 74.7 & 76.4  \\ 
         SRC                   & \xmarkg & \textcolor{gray}{0.8}    & RA  & \pzo \textcolor{gray}{90}  & \textcolor{gray}{224$^2$} & 82.1 & 84.8 & 86.3 & 71.8 & 75.0 &  76.7 \\
         \textcolor{gray}{SRC} & \cmark  & \textcolor{gray}{0.8}    & RA  & \pzo \textcolor{gray}{90}  & \textcolor{gray}{224$^2$} & 82.4 & 85.0 & 86.4 & 72.4 & 75.7 & 77.4  \\
         \textcolor{gray}{SRC} & \cmarkg & \xmark                   & RA  & \pzo \textcolor{gray}{90}  & \textcolor{gray}{224$^2$} & 82.3 & 85.1 & 86.5 & 72.4 & 75.6 &  77.2  \\
         \textcolor{gray}{SRC} & \cmarkg & \xmarkg                  & 3A  & \pzo \textcolor{gray}{90}  & \textcolor{gray}{224$^2$} & 82.6 & 85.2 & 86.8 &  72.6 & 76.1 & 78.3 \\
         \textcolor{gray}{SRC} & \cmarkg & \xmarkg & \textcolor{gray}{3A}  & 240                       & \textcolor{gray}{224$^2$} & \textbf{83.1} & \textbf{85.7} & \textbf{87.0} &  \textbf{73.8}  & \textbf{76.5} & \textbf{78.6}  \\
\midrule
         \textcolor{gray}{SRC} & \cmarkg & \xmarkg & \textcolor{gray}{3A}  & \textcolor{gray}{240}    & 384$^2$ & \emph{84.8} & \emph{86.7} & \emph{87.7} &  75.1  & 77.9  & 79.1  \\
    \bottomrule
    \end{tabular}}
    \caption{Ablation path: \textbf{augmentation and regularization} with ImageNet-21k pre-training (at resolution 224$\times$224) and ImageNet-1k fine-tuning. We measure the impact of changing Random Resize Crop (RRC) to Simple Random Crop (SRC), adding LayerScale (LS), removing Mixup, replacing RandAugment (RA) by 3-Augment (3A), and finally employing a longer number of epochs during the pre-training phase on ImageNet-21k. All experiments are done with Seed 0 with fixed hparams except the drop-path rate of stochastic depth, which depends on the model and is increased by 0.05 for the longer pre-training. We report 2 digits top-1 accuracy but note that the standard standard deviation is around $0.1$ on our ViT-B baseline. 
    Note that all these changes are neutral w.r.t. complexity except in the last row, where the fine-tuning at resolution 384$\times$384 significantly increases the complexity. }
    \label{tab:rrc_ablation}
\end{table}

\subsubsection{Impact of training resolution}

In Table~\ref{tab:fixRes} we report the evolution of the performance according to the training resolution. 
We observe that we benefit from the FixRes~\cite{Touvron2019FixRes} effect. 
By training at resolution 192$\times$192 (or 160$\times$160) we get a better performance at 224 after a slight fine-tuning than when training from scratch at 224$\times$224. 

We observe that the resolution has a regularization effect. While it is known that it is best to use a smaller resolution at training time~\cite{Touvron2019FixRes}, we also observe in the training curves that this show reduces the overfitting of the larger models. This is also illustrated by our results Table~\ref{tab:fixRes} with ViT-H and ViT-L. This is especially important with longer training, where models overfit without a stronger regularisation.
This smaller resolution implies that there are less patches to be processed, and therefore it reduces the training cost and increases the performance. In that respect it effect is comparable to that of MAE~\cite{He2021MaskedAA}.
We also report results with ViT-H 52 layers and ViT-H 26 layers parallel~\cite{Touvron2022ThreeTE} models with 1B parameters. Due to the lower resolution training it is easier to train these models.

\begin{table}[]
    \centering
    \scalebox{0.8}{
    \begin{tabular}{c|cc|cc|ccc}
    \toprule
        \multirow{2}{*}{Model} & \multicolumn{2}{c}{epochs} & \multicolumn{2}{c}{Resolution}  & \multicolumn{3}{c}{ImageNet top-1 acc} \\
        & Train. & FT & Train. & FT & val & real & v2 \\
        \midrule
        \multirow{8}{*}{ViT-B} & 
        \multirow{4}{*}{400} & \multirow{3}{*}{20} & 
        $128\times128$& \multirow{4}{*}{$224\times224$} & 83.2 & \textbf{88.1} & \underline{73.2} \\ %
        & & & $160\times160$&  & \underline{83.3} & \underline{88.0} & \textbf{73.4} \\
        & & & $192\times192$ &  & \textbf{83.5} & \underline{88.0} & 72.8 \\
        & & \_ &$224\times224$ & & 83.1 & 87.7 & 72.6 \\
        \cmidrule{2-8}
         &\multirow{4}{*}{800} & \multirow{3}{*}{20} & 
        $128\times128$& \multirow{4}{*}{$224\times224$} & 83.5 & \textbf{88.3} & 73.4  \\
        & & & $160\times160$&  & 83.6 & \underline{88.2} & \underline{73.5} \\
        & & & $192\times192$ &  & \textbf{83.8} & \underline{88.2} & \textbf{73.6}\\ %
        & & \_ &$224\times224$  & & \underline{83.7} & 88.1 & 73.1   \\
        \midrule
        \multirow{8}{*}{ViT-L} & 
        \multirow{4}{*}{400} & \multirow{3}{*}{20} &
        $128\times128$& \multirow{4}{*}{$224\times224$} & 83.9 & \textbf{88.8} &  \underline{74.3} \\
        & & & $160\times160$&  & \underline{84.4} & \textbf{88.8} & \underline{74.3} \\ %
        & & & $192\times192$  & & \textbf{84.5} & \textbf{88.8} & \textbf{75.1} \\
        & & \_  & $224\times224$ & & 84.2 & 88.6  & \underline{74.3} \\
         \cmidrule{2-8}
        &\multirow{4}{*}{800} & \multirow{3}{*}{20} &
        $128\times128$& \multirow{4}{*}{$224\times224$} & 84.5 &  \textbf{88.9} & 74.7 \\%
        & & & $160\times160$&  & \underline{84.7} & \textbf{88.9} &  \textbf{75.2}\\ %
        & & & $192\times192$  & & \textbf{84.9} & 88.7 & \underline{75.1} \\
        & & \_  & $224\times224$ & & 84.5 & \underline{88.8}  & 75.0 \\
        \midrule
        \multirow{8}{*}{ViT-H} & 
        \multirow{4}{*}{400} & \multirow{3}{*}{20} &
        $126\times126$& \multirow{4}{*}{$224\times224$} & 84.7  & \underline{89.2} & 75.2  \\
        & & & $154\times154$&  & \textbf{85.1} & \textbf{89.3} & \underline{75.3} \\ %
        & & & $182\times182$  &  & \textbf{85.1} & \underline{89.2} & \textbf{75.4}  \\
        & & \_  & $224\times224$ &  & 84.8 & 89.1 & \underline{75.3} \\
         \cmidrule{2-8}
        &\multirow{4}{*}{800} & \multirow{3}{*}{20} &
        $126\times126$& \multirow{4}{*}{$224\times224$} & \underline{85.1} & \textbf{89.2}  & 75.6 \\%
        & & & $154\times154$&  & \textbf{85.2} & \textbf{89.2} & \textbf{75.9} \\ %
        & & & $182\times182$  & & \underline{85.1}  & 88.9 & \textbf{75.9} \\ %
        & & \_  & $224\times224$ & & 84.6 & 88.5  & 74.9 \\    
        \midrule
        ViT-H-52 & 400 & 20 & $126\times126$ & $224\times224$ & 84.9 & 89.2 & 75.6\\
        \midrule
        ViT-H-26$\times$2 & 400 & 20 & $126\times126$ & $224\times224$ & 84.9 & 89.1 & 75.3 \\
    \bottomrule
    \end{tabular}}
    \caption{We compare ViT architectures pre-trained on ImageNet-1k only with different training resolution followed by a fine-tuning at resolution $224\times224$. We benefit from the FixRes effect~\cite{Touvron2019FixRes} and get better performance with a lower training resolution (e.g resolution $160\times160$ with patch size 16 represent 100 tokens vs 196 for $224\times224$. This represents a reduction of ~50\% of the number of tokens).}
    \label{tab:fixRes}
\end{table}

\subsubsection{Comparison with previous training recipes for ViT}
In Figure~\ref{fig:method_fig}, we compare training procedures used to pre-train the ViT architecture either on ImageNet-1k and ImageNet-21k.
Our procedure outperforms existing recipes with a large margin. For instance, with ImageNet-21k pre-training we have an improvement of $+3.0\%$ with ViT-L in comparison to the best approach.
Similarly, when training from scratch on ImageNet-1k we improve the accuracy by $+2.1\%$ for ViT-H  compared to the previous best approach, and by $+4.3\%$ with the best approach that does not use EMA. See also detailed results in our appendices. %

\subsection{Image Classification}

\paragraph{\textbf{ImageNet-1k.}}

In Table~\ref{tab:mainres} we compare ViT architectures trained with our training recipes on ImageNet-1k with other architectures. We include a comparison with the recent SwinTransformers~\cite{liu2021swin} and ConvNeXts~\cite{Liu2022convnext}.%

\paragraph{\textbf{Overfitting evaluation.}}

\begin{figure}
    \begin{minipage}{0.48 \linewidth}
    \centering
    ImageNet-1k
    \includegraphics[width=\linewidth]{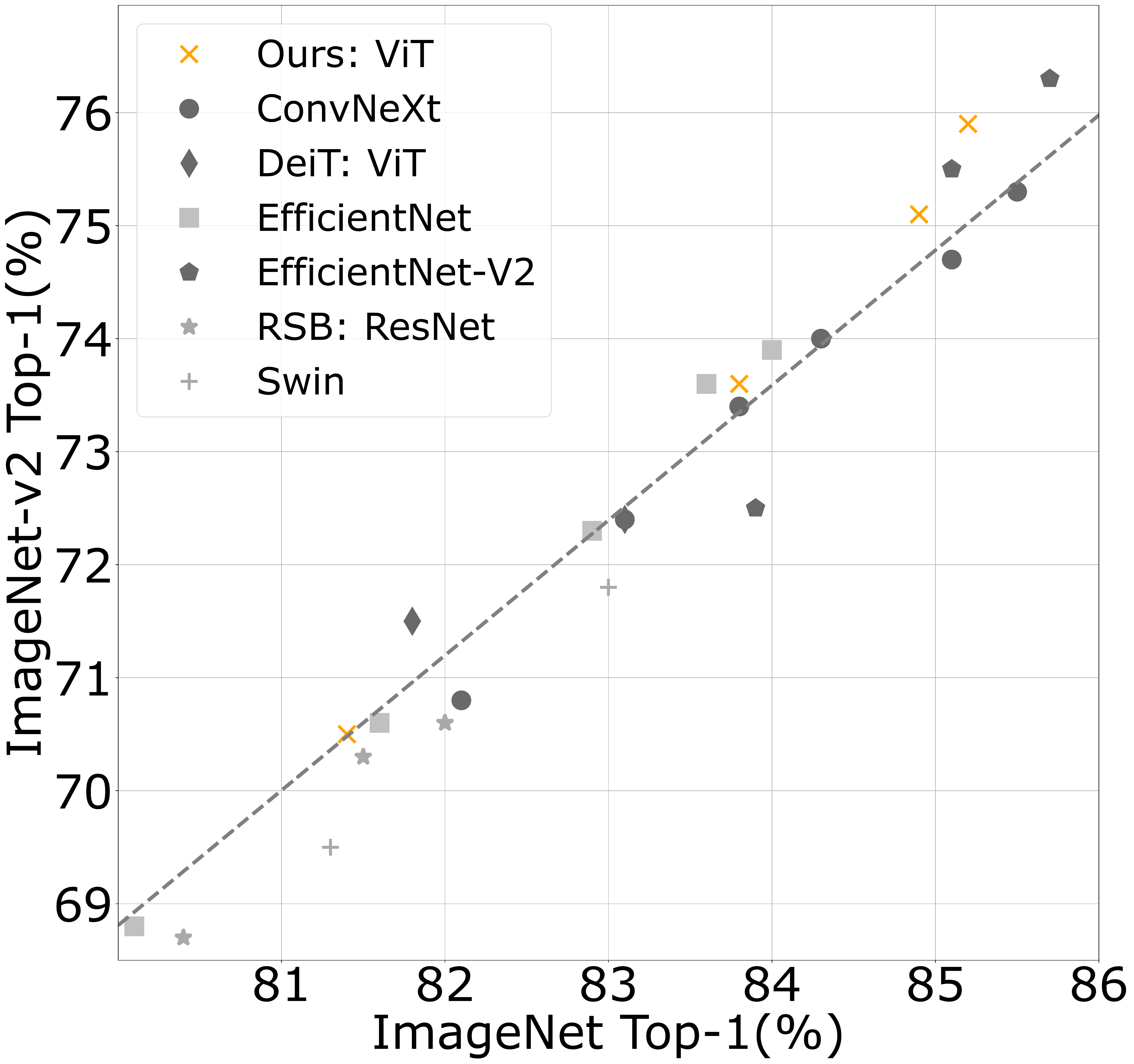}
    \end{minipage}
    \hfill
    \begin{minipage}{0.48 \linewidth}
    \centering
    ImageNet-21k
    \includegraphics[width=\linewidth]{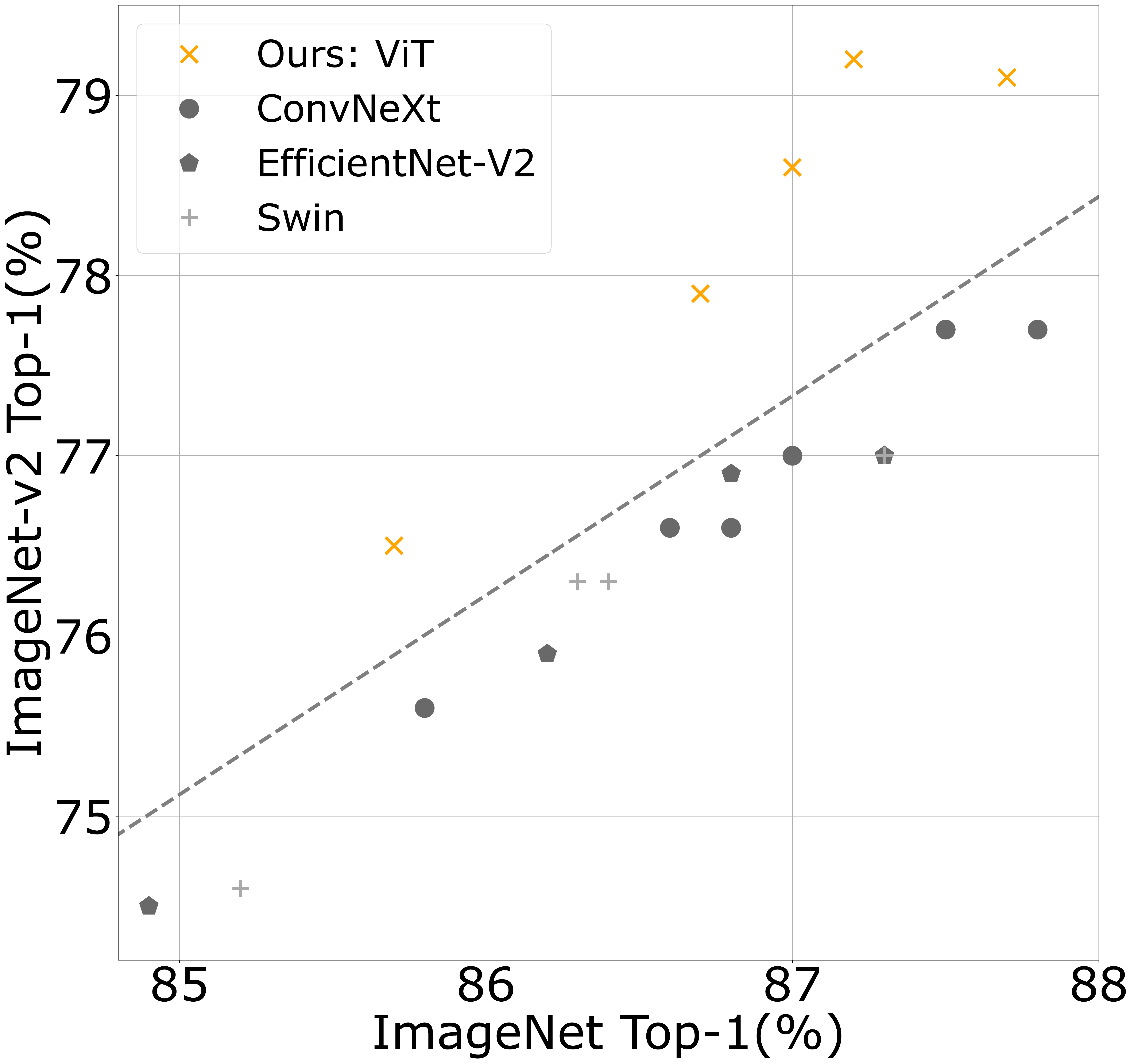}
    \end{minipage}
    \caption{Generalization experiment: top-1 accuracy on ImageNet1k-val versus ImageNet-v2 for models in Table~\ref{tab:mainres} and Table~\ref{tab:mainres_22k}. We display a linear interpolation of all points in order to compare the generalization capability (or level of overfitting) for the different models.
    \label{fig:imagenet_v2}}
\end{figure}

The comparison between Imagenet-val and -v2 is a way to quantify overfitting~\cite{Touvron2020FixingTT}, or at least the better capability to generalize in a nearby setting without any fine-tuning\footnote{Caveat: The measures are less robust with -V2 as the number of test images is 10000 instead of 50000 for Imagenet-val. This translates to a higher standard deviation ($0.2\%$).}.   
In Figure~\ref{fig:imagenet_v2} we plot ImageNet-val top-1 accuracy  vs ImageNet-v2 top-1 accuracy in order to evaluate how the models performed when evaluated on a test set never seen at validation time. 
Our models overfit significantly than all other models considered, especially on ImageNet-21k. This is a good behaviour that validates the fact that our restricted choice of hyper-parameters and variants in our recipe does not lead to (too much) overfitting.

\begin{table}
    \caption{
\textbf{Classification with Imagenet1k training.} 
We compare architectures with comparable FLOPs and number of parameters. All models are trained on ImageNet1k only without distillation nor self-supervised pre-training.
We report Top-1 accuracy on the validation set of ImageNet1k and ImageNet-V2 with different measure of complexity: throughput, FLOPs, number of parameters and peak memory usage. 
The throughput and peak memory are measured on a single V100-32GB GPU with batch size fixed to 256 and mixed precision. 
For ResNet~\cite{He2016ResNet} and RegNet~\cite{Radosavovic2020RegNet} we report the improved results from Wightman et al.~\cite{wightman2021resnet}. Note that different models may have received a different optimization effort. $\uparrow$R indicates that the model is fine-tuned at the resolution $R$ and -R indicates that the model is trained at resolution R.
\label{tab:mainres}}
    \centering
    \scalebox{0.78}{
    \begin{tabular}{@{\ }l@{}c@{\ \ }c@{\ \ \ }r@{\ \ }r|cc@{\ }}
        \toprule
        Architecture        & nb params & throughput & FLOPs & Peak Mem & Top-1  & V2 \\
                      & ($\times 10^6$) & (im/s) & ($\times 10^9$) & (MB)\ \ \ \  & Acc.  & Acc. \\[3pt]

\toprule
\multicolumn{7}{c}{\textbf{``Traditional'' ConvNets}} \\[3pt]
     ResNet-50~\cite{He2016ResNet,wightman2021resnet} &  25.6    & 2587  &  4.1      & 2182 & 80.4  & 68.7 \\
    ResNet-101~\cite{He2016ResNet,wightman2021resnet}&  44.5    & 1586  &  7.9      & 2269 & 81.5  & 70.3  \\
    ResNet-152~\cite{He2016ResNet,wightman2021resnet}&  60.2    &  1122 & 11.6      & 2359 & 82.0  & 70.6 \\
    \midrule
	 RegNetY-4GF~\cite{Radosavovic2020RegNet,wightman2021resnet}       & 20.6  & 1779  & \tzo4.0 & 3041 & 81.5  & 70.7 \\
	 RegNetY-8GF~\cite{Radosavovic2020RegNet,wightman2021resnet}       & 39.2  & 1158 & \tzo8.0 & 3939 & 82.2 & 71.1 \\
	 RegNetY-16GF~\cite{Radosavovic2020RegNet,Touvron2020TrainingDI}      & 83.6  & \pzo714 & \dzo16.0 & 5204   & 82.9  & 72.4 \\

    \midrule
	 EfficientNet-B4~\cite{tan2019efficientnet} & 19.0  & \pzo573 & \tzo4.2 &  10006  & 82.9  & 72.3\\
	 EfficientNet-B5~\cite{tan2019efficientnet} & 30.0  & \pzo268 & \tzo9.9 &  11046  & 83.6  & 73.6\\
	 
	 \midrule
	 EfficientNetV2-S~\cite{Tan2021EfficientNetV2SM}& 21.5 & 874 & 8.5 & 4515 & 83.9 & 74.0 \\
    EfficientNetV2-M~\cite{Tan2021EfficientNetV2SM}& 54.1 & 312 & 25.0 & 7127 & 85.1 & 75.5 \\
    EfficientNetV2-L~\cite{Tan2021EfficientNetV2SM}& 118.5 & 179 & 53.0 & 9540 & 85.7 & 76.3 \\
	
\toprule

\multicolumn{5}{c}{\textbf{Vision Transformers derivative}} \\ [5pt]
    PiT-S-224~\cite{Heo2021RethinkingSD} & 23.5 & 1809\pzo & 2.9 &  3293 & 80.9 & \_\\
    PiT-B-224~\cite{Heo2021RethinkingSD} & 73.8 & 615 & 12.5  &  7564 & 82.0 & \_\\
	Swin-T-224~\cite{liu2021swin} & 28.3 & 1109\pzo & 4.5 & 3345 & 81.3 &  69.5 \\
    Swin-S-224~\cite{liu2021swin} & 49.6 & 718 & 8.7 & 3470 &  83.0 &   71.8 \\
    Swin-B-224~\cite{liu2021swin} & 87.8  & 532 & 15.4 & 4695 & 83.5 &   \_ \\
    Swin-B-384~\cite{liu2021swin} & 87.9 &   160       &   47.2    & 19385 & 84.5 & \_ \\
    \toprule
 \multicolumn{7}{c}{\textbf{Vision MLP \& Patch-based ConvNets}} \\[3pt]

    Mixer-B/16~\cite{tolstikhin2021MLPMixer} &  59.9 &   993 & 12.6  & 1448  &   76.4 &  63.2 \\
    ResMLP-B24~\cite{Touvron2021ResMLPFN} &  116.0\pzo &      1120\pzo    & \dzo23.0 & 930  &   81.0 &  69.0 \\
    PatchConvNet-S60-224~\cite{Touvron2021AugmentingCN} & 25.2  & 1125\pzo & 4.0 & 1321 & 82.1 &  71.0\\
    PatchConvNet-B60-224~\cite{Touvron2021AugmentingCN} &  99.4 & 541 & 15.8 & 2790 & 83.5 &   72.6\\
    PatchConvNet-B120-224~\cite{Touvron2021AugmentingCN} & 188.6\pzo  & 280 & 29.9  & 3314 & 84.1 & 73.9\\
    
    ConvNeXt-B-224~\cite{Liu2022convnext} & 88.6 & 563 & 15.4 & 3029 & 83.8 & 73.4 \\
    ConvNeXt-B-384~\cite{Liu2022convnext} & 88.6 & 190 & 45.0 & 7851 &  85.1 & 74.7 \\
    
    ConvNeXt-L-224~\cite{Liu2022convnext} & 197.8 & 344 & 34.4 & 4865 &  84.3 & 74.0 \\
    ConvNeXt-L-384~\cite{Liu2022convnext} & 197.8 & 115 & 101.0 & 11938 &  85.5 & 75.3 \\
    
    \toprule
    \multicolumn{7}{c}{\textbf{Our Vanilla Vision Transformers}} \\ [5pt]
    \rowcolor{Goldenrod}
 
    \rowcolor{Goldenrod}
    ViT-S & 22.0  & 1891\pzo & \tzo4.6 & 987 & 81.4 &  70.5 \\
    \rowcolor{Goldenrod}
    ViT-S$\uparrow$384 &  22.0 & 424 & 15.5 & 4569 &  83.4 & 73.1 \\
    \rowcolor{Goldenrod}
    ViT-B & 86.6  & 831  & \dzo17.5 & 2078 & 83.8 &  73.6\\
    \rowcolor{Goldenrod}
    ViT-B$\uparrow$384 & 86.9  & 190 & \dzo55.5 & 8956 & 85.0 & 74.8 \\
    \rowcolor{Goldenrod}
    ViT-L & 304.4 & 277 & 61.6 & 3789 & 84.9 & 75.1    \\
    \rowcolor{Goldenrod}
    ViT-L$\uparrow$384 & 304.8 & \pzo67 & 191.2 & 12866 & 85.8 & 76.7 \\%
    \rowcolor{Goldenrod}
    ViT-H & 632.1 & 112 & 167.4 & 6984 & 85.2 & 75.9     \\

    \bottomrule
    \end{tabular}}
\end{table}

\paragraph{\textbf{ImageNet-21k.}}
In Table~\ref{tab:mainres_22k} we compare ViT architecture  pre-trained on ImageNet-21k with our training recipe then finetuned on ImageNet-1k. We can observe that the findings are similar to what we obtained on ImageNet-1k only.  %
 
\begin{table}
    \caption{
\textbf{Classification with Imagenet-21k training.} 
We compare architectures with comparable FLOPs and number of parameters. All models are trained on ImageNet-21k without distillation nor self-supervised pre-training.
We report Top-1 accuracy on the validation set of ImageNet-1k and ImageNet-V2 with different measure of complexity: throughput, FLOPs, number of parameters and peak memory usage. 
The throughput and peak memory are measured on a single V100-32GB GPU with batch size fixed to 256 and mixed precision. 
For Swin-L we decrease the batch size to 128 in order to avoid out of memory error and re-estimate the memory consumption.
$\uparrow$R indicates that the model is fine-tuned at the resolution $R$. 
 \label{tab:mainres_22k}}
    \centering
    \scalebox{0.8}{
    \begin{tabular}{@{\ }l@{}c@{\ \ }c@{\ \ \ }r@{\ \ }r|cc@{\ }}
        \toprule
        Architecture        & nb params & throughput & FLOPs & Peak Mem & Top-1  & V2 \\
                      & ($\times 10^6$) & (im/s) & ($\times 10^9$) & (MB)\ \ \ \  & Acc.  & Acc. \\[3pt]

\toprule
\multicolumn{7}{c}{\textbf{``Traditional'' ConvNets}} \\[3pt]
R-101x3$\uparrow$384~\cite{kolesnikov2019big} & 388 & \_ & 204.6 & \_ & 84.4 & \_ \\
R-152x4$\uparrow$480~\cite{kolesnikov2019big} & 937 & \_ & 840.5 & \_  & 85.4 & \_ \\
\midrule
EfficientNetV2-S$\uparrow$384~\cite{Tan2021EfficientNetV2SM}& 21.5 & 874 & 8.5 & 4515 & 84.9 & 74.5 \\
EfficientNetV2-M$\uparrow$480~\cite{Tan2021EfficientNetV2SM}& 54.1 & 312 & 25.0 & 7127 & 86.2 & 75.9 \\
EfficientNetV2-L$\uparrow$480~\cite{Tan2021EfficientNetV2SM}& 118.5 & 179 & 53.0 & 9540 & 86.8 & 76.9 \\
EfficientNetV2-XL$\uparrow$512~\cite{Tan2021EfficientNetV2SM} & 208.1 & \_ & 94.0 & \_& 87.3 & 77.0 \\

\toprule

    \multicolumn{7}{c}{\textbf{Patch-based ConvNets}} \\[3pt]
 
    ConvNeXt-B~\cite{Liu2022convnext} & 88.6 & 563 & 15.4 &3029 &  85.8 & 75.6 \\     
    ConvNeXt-B$\uparrow$384~\cite{Liu2022convnext} & 88.6 & 190 & 45.1 &7851 &  86.8 & 76.6\\     
    
    ConvNeXt-L~\cite{Liu2022convnext} & 197.8 & 344 & 34.4 & 4865 &  86.6 & 76.6\\     
    ConvNeXt-L$\uparrow$384~\cite{Liu2022convnext} & 197.8 & 115 & 101 &11938 &  87.5 & 77.7 \\  
    
    ConvNeXt-XL~\cite{Liu2022convnext} & 350.2 & 241 & 60.9 & 6951 &  87.0 & 77.0\\     
    ConvNeXt-XL$\uparrow$384~\cite{Liu2022convnext} & 350.2 & \pzo80 & 179.0 & 16260 &  87.8 & 77.7 \\

    \toprule

\multicolumn{7}{c}{\textbf{Vision Transformers derivative}} \\ [5pt]
     Swin-B~\cite{liu2021swin}& 87.8 & 532 & 15.4 & 4695 &  85.2 & 74.6 \\
     Swin-B$\uparrow$384~\cite{liu2021swin}& 87.9 & 160 & 47.0 & 19385 &  86.4 & 76.3 \\
     
     Swin-L~\cite{liu2021swin}& 196.5 & 337 & 34.5 & 7350 &  86.3 & 76.3 \\
     Swin-L$\uparrow$384~\cite{liu2021swin}& 196.7 & 100 & 103.9 & 33456 &  87.3 & 77.0 \\     

\toprule
\multicolumn{7}{c}{\textbf{Vanilla Vision Transformers}} \\ [5pt]
ViT-B/16~\cite{Steiner2021HowTT}  & 86.6 & 831 & 17.6 & 2078 & 84.0 &  \_\\
ViT-B/16$\uparrow$384~\cite{Steiner2021HowTT}  & 86.7 & 190 & 55.5 & 8956 & 85.5 & \_\\

ViT-L/16~\cite{Steiner2021HowTT}  & 304.4 & 277 & 61.6 & 3789 &  84.0 & \_\\
ViT-L/16$\uparrow$384~\cite{Steiner2021HowTT}  & 304.8 & \pzo67 & 191.1 & 12866 & 85.5 &\_ \\

    \toprule

    \rowcolor{Goldenrod!75}
    \multicolumn{7}{c}{\textbf{Our Vanilla Vision Transformers}} \\ [5pt]
    \rowcolor{Goldenrod}
    \rowcolor{Goldenrod}
    ViT-S & 22.0  & 1891\pzo & \tzo4.6 & 987 & 83.1 & 73.8 \\

    \rowcolor{Goldenrod}
    ViT-B & 86.6  & 831  & \dzo17.6 & 2078 & 85.7 & 76.5 \\
    \rowcolor{Goldenrod}
    ViT-B$\uparrow$384 & 86.9  & 190 & \dzo55.5 &  8956 & 86.7 & 77.9 \\
    \rowcolor{Goldenrod}
    ViT-L & 304.4 & 277 & 61.6 & 3789  & 87.0 &  78.6 \\
    \rowcolor{Goldenrod}
    ViT-L$\uparrow$384 & 304.8 & \pzo67 & 191.2 & 12866 & 87.7 & 79.1  \\
    \rowcolor{Goldenrod}
    ViT-H & 632.1 & 112 & 167.4 & 6984 & 87.2 & 79.2 \\

    \bottomrule
    \end{tabular}}
\end{table}

\paragraph{\textbf{Comparison with BerT-like pre-training.}}

In Table~\ref{tab:comp_ssl} we compare ViT models trained with our training recipes with ViT trained with different BerT-like approaches. We observe that for an equivalent number of epochs our approach gives comparable performance on ImageNet-1k and better on ImageNet-v2 as well as in segmentation on Ade. For BerT like pre-training we compare our method with MAE~\cite{He2021MaskedAA} and BeiT~\cite{bao2021beit} because they remain  relatively simple approaches with very good performance. As our approach does not use distillation or multi-crops we have not made a comparison with approaches such as PeCo~\cite{Dong2021PeCoPC} which use an auxiliary model as a psycho-visual loss and iBoT~\cite{Zhou2021iBOTIB}, which uses multi-crop and an exponential moving average of the model.
\begin{table}[t]
    \centering
    \scalebox{0.8}{
    \begin{tabular}{c|c|ccc|ccc}
    \toprule
        Pretrained & \multirow{2}{*}{Model} & 
        \multirow{2}{*}{Method}  &
        \# pre-training & \# finetuning & \multicolumn{3}{c}{ImageNet}   \\
        data & &  & epochs & epochs & val & Real & V2  \\
        \midrule
        \multirow{12}{*}{INET-1k} & \multirow{5}{*}{ViT-B}
         & \multirow{2}{*}{BeiT} & 300 & $100^{\small(1k)}$ & 82.9  & \_ & \_ \\
        & &  &    800 & $100^{\small(1k)}$ & 83.2  & \_ & \_\\
          \cmidrule{3-8}
        & & MAE$^\star$   & 1600 & $100^{\small(1k)}$ & \underline{83.6}  & \underline{88.1} &  \underline{73.2}  \\
         \cmidrule{3-8}
        & &  \cellcolor{Goldenrod}  & \cellcolor{Goldenrod} $400^{\small(1k)}$  & \cellcolor{Goldenrod} $20^{\small(1k)}$ & \cellcolor{Goldenrod} 83.5  & \cellcolor{Goldenrod} 88.0 & \cellcolor{Goldenrod} 72.8  \\
        
        & &  \multirow{-2}{*}{\cellcolor{Goldenrod} Ours} & \cellcolor{Goldenrod} $800^{\small(1k)}$ & \cellcolor{Goldenrod} $20^{\small(1k)}$ & \cellcolor{Goldenrod} \textbf{83.8}  & \cellcolor{Goldenrod} \textbf{88.2} & \cellcolor{Goldenrod} \textbf{73.6}  \\
        \cmidrule{2-8}
        
        & \multirow{7}{*}{ViT-L}
        & BeiT   & 800 & $30^{\small(1k)}$ & \underline{85.2}  & \_ & \_  \\
        \cmidrule{3-8}
      &  & \multirow{3}{*}{MAE} & 400 & $50^{\small(1k)}$ & 84.3  & \_ & \_ \\
       &  &  & 800 & $50^{\small(1k)}$ & 84.9  & \_ & \_ \\
      &  &  & 1600 & $50^{\small(1k)}$ & 85.1  & \_ & \_ \\
      \cmidrule{3-8}
      & & MAE$^\star$   & 1600 & $50^{\small(1k)}$ & \textbf{85.9}  & \textbf{89.4} & \textbf{76.5} \\
        \cmidrule{3-8}
       & & \cellcolor{Goldenrod}   &  \cellcolor{Goldenrod} $400^{\small(1k)}$ & \cellcolor{Goldenrod} $20^{\small(1k)}$ & \cellcolor{Goldenrod} 84.5  & \cellcolor{Goldenrod} \underline{88.8} &  \cellcolor{Goldenrod} \underline{75.1}  \\  
       & & \multirow{-2}{*}{\cellcolor{Goldenrod} Ours}   & \cellcolor{Goldenrod} $800^{\small(1k)}$ & \cellcolor{Goldenrod} $20^{\small(1k)}$ & \cellcolor{Goldenrod} 84.9   & \cellcolor{Goldenrod} 88.7  &  \cellcolor{Goldenrod} \underline{75.1}    \\  
     \midrule
     
     \multirow{8}{*}{INET-21k} & \multirow{4}{*}{ViT-B}
      &  \multirow{2}{*}{BeiT}  &  150 & $50^{\small(1k)}$  & 83.7  & 88.2 & 73.1  \\
         &  &  &  150 + $90^{\small(21k)}$ &   $50^{\small(1k)}$ & \underline{85.2}  &  \underline{89.4} & 75.4  \\
    \cmidrule{3-8}
    & &  \cellcolor{Goldenrod}   &  \cellcolor{Goldenrod} $90^{\small(21k)}$ &   \cellcolor{Goldenrod} $50^{\small(1k)}$ &  \cellcolor{Goldenrod} \underline{85.2}  &  \cellcolor{Goldenrod} \underline{89.4} &   \cellcolor{Goldenrod} \underline{76.1}  \\
     & & \multirow{-2}{*}{\cellcolor{Goldenrod} Ours}   &  \cellcolor{Goldenrod} $240^{\small(21k)}$ &   \cellcolor{Goldenrod} $50^{\small(1k)}$ &  \cellcolor{Goldenrod} \textbf{85.7}  &  \cellcolor{Goldenrod} \textbf{89.5} &   \cellcolor{Goldenrod} \textbf{76.5}  \\
    \cmidrule{2-8}
     & \multirow{4}{*}{ViT-L}
     & \multirow{2}{*}{BeiT} &  150 & $50^{\small(1k)}$  & 86.0  & 89.6 & 76.7  \\
    &  &  &  150 + $90^{\small(21k)}$ &   $50^{\small(1k)}$ & \textbf{87.5}  &  \textbf{90.1} & \textbf{78.8}  \\
    \cmidrule{3-8}
    &  &  \cellcolor{Goldenrod}   &  \cellcolor{Goldenrod} $90^{\small(21k)}$ &  \cellcolor{Goldenrod} $50^{\small(1k)}$ &   \cellcolor{Goldenrod} 86.8  &  \cellcolor{Goldenrod} 89.9 &  \cellcolor{Goldenrod} 78.3  \\
    &  & \multirow{-2}{*}{\cellcolor{Goldenrod} Ours}   &  \cellcolor{Goldenrod} $240^{\small(21k)}$ &  \cellcolor{Goldenrod} $50^{\small(1k)}$ &   \cellcolor{Goldenrod} \underline{87.0}  &  \cellcolor{Goldenrod} \underline{90.0} &  \cellcolor{Goldenrod} \underline{78.6} \\
     \bottomrule
    \end{tabular}}
    \caption{Comparison of self-supervised pre-training with our approach. As our approach is fully supervised, this table is given as an indication. All models are evaluated at resolution $224\times 224$. We report Image classification results on ImageNet val, real and v2 in order to evaluate overfitting. $^{\small(21k)}$ indicate a finetuning with labels on ImageNet-21k and $^{\small(1k)}$ indicate a finetuning with labels on ImageNet-1k. $^\star$ design the improved setting of MAE using pixel (w/ norm) loss.
    }
    \label{tab:comp_ssl}
\end{table}

\subsection{Downstream tasks and other architectures} 

\subsubsection{Transfer Learning}

In order to evaluate the quality of the ViT models learned through our training procedure we evaluated them  with transfer learning tasks. We focus on the performance of ViT models pre-trained on ImageNet-1k only at resolution $224 \times 224$ during 400 epochs on the 6 datasets shown in Table~\ref{tab:dataset}. Our results are presented in Table~\ref{tab:sota_tl}. In Figure~\ref{fig:crop} we measure the impact of the crop ratio at inference time on transfer learning results. We observe that on iNaturalist this parameter has a significant impact on the performance. As recommended in the paper Three Things~\cite{Touvron2022ThreeTE} we finetune only the attention layers for transfer learning experiments on Flowers, this improves performance by $0.2\%$.

\begin{table}[t]
    \caption{We compare Transformers based models on different transfer learning tasks with ImageNet-1k pre-training. We report results with our default training on ImageNet-1k (400 epochs at resolution $224 \times 224$). We also report results with convolutional architectures for reference. For consistency we keep our crop ratio equal to 1.0 on all datasets. Other works use 0.875, which is better for iNat-19 and iNat-18, see Figure~\ref{fig:crop}.%
    \label{tab:sota_tl}}
    \smallskip
    \centering
    \scalebox{0.8}{
    \begin{tabular}{l|cccccc}
    \toprule
    Model    & CIFAR-10 & CIFAR-100  & Flowers & Cars & iNat-18 & iNat-19  \\
    \midrule                                                                          
    Grafit ResNet-50~\cite{Touvron2020GrafitLF}  & \_   & \_ & 98.2 & 92.5 & 69.8 & 75.9 \\
    ResNet-152~\cite{Chu2020FeatureSA}  & \_ & \_ & \_ & \_ & 69.1 & \_ \\
    \midrule                                                                          
    ViT-B/16~\cite{dosovitskiy2020image}         & 98.1 & 87.1 & 89.5 & \_   & \_ & \_  \\
    ViT-L/16~\cite{dosovitskiy2020image}         & 97.9 & 86.4 & 89.7 & \_   & \_ & \_  \\
    \midrule
    ViT-B/16~\cite{Steiner2021HowTT}         & \_ & 87.8 & 96.0 & \_   & \_ & \_  \\
    ViT-L/16~\cite{Steiner2021HowTT}         & \_ & 86.2 & 91.4 & \_   & \_ & \_  \\
    \midrule
    DeiT-B & 99.1 & 90.8 & 98.4  & 92.1 & 73.2 &  77.7 \\
    \midrule 
    \rowcolor{Goldenrod}
    Ours ViT-S & 98.9 & 90.6 & 96.4 & 89.9 & 67.1 & 72.7 \\
    \rowcolor{Goldenrod}
    Ours ViT-B & 99.3 & 92.5 & 98.6 & 93.4 & 73.6 & 78.0\\ 
    \rowcolor{Goldenrod}
    Ours ViT-L & \textbf{99.3} & \textbf{93.4} & \textbf{98.9} & \textbf{94.5} & \textbf{75.6} & \textbf{79.3} \\ 
    \bottomrule
    \end{tabular}}
\end{table} %

\subsubsection{Semantic segmentation}
We evaluate our ViT baselines models (400 epochs schedules for ImageNet-1k models and 90 epochs for ImageNet-21k models) with semantic segmentation experiments on  ADE20k dataset~\cite{Zhou2017ScenePT}.
This dataset consists of 20k training and 5k validation images with labels over 150 categories. 
For the training, we adopt the same schedule as in Swin:~160k iterations with UperNet~\cite{xiao2018unified}. 
At test time we evaluate with a single scale and multi-scale.
Our UperNet implementation is based on the XCiT~\cite{el2021xcit} repository.
By default the UperNet head uses an embedding dimension of 512. In order to save compute, for  small and tiny models we set it to the size of their working dimension, i.e. 384 for small and 192 for tiny. We keep the 512 by default as it is done in XCiT for other models.
Our results are reported in Table~\ref{tab:sem_seg}.
We observe that vanilla ViTs trained with our training recipes have a better FLOPs-accuracy trade-off than recent architectures like XCiT or Swin.
\begin{table}[t]
       \caption{\textbf{ADE20k semantic segmentation} performance using UperNet \cite{xiao2018unified} (in comparable settings~\cite{Dong2021CSWinTA,el2021xcit,liu2021swin}). All models are pre-trained on ImageNet-1k  except models with $^\dagger$ symbol that are pre-trained on ImageNet-21k. We report the pre-training resolution used on ImageNet-1k and ImageNet-21k.
       \smallskip
       \label{tab:sem_seg}}
        \centering
        \scalebox{0.8}{
        \begin{tabular}{l|l|cc|cc}
        \toprule
             \multirow{2}{*}{Backbone} & Pre-training & \multicolumn{4}{c}{UperNet}  \\
             & \multirow{2}{*}{resolution} & \#params & FLOPs & Single scale & Multi-scale  \\
             & & ($\times 10^6$) & ($\times 10^9$) & mIoU  & mIoU \\
            \midrule 
             ResNet50 & $224\times224$ & 66.5 & \_ & 42.0 & \_ \\
             DeiT-S  & $224\times224$ & 52.0 & 1099 & \_ & 44.0 \\
             XciT-T12/16 & $224\times224$ & 34.2 & \pzo874  & 41.5  & \_\\
             XciT-T12/8  & $224\times224$ & 33.9 & 942 & 43.5 & \_ \\
             Swin-T   & $224\times224$ & 59.9 & \pzo945 & 44.5 & 46.1 \\

             \rowcolor{Goldenrod}
            Our ViT-T & $224\times224$ & 10.9 & 148 & 40.1 & 41.8 \\
            \rowcolor{Goldenrod}
            Our ViT-S & $224\times224$ & 41.7 & 588 & \textbf{45.6} & \textbf{46.8}\\
            \midrule
             XciT-M24/16  & $224\times224$ & 112.2 & 1213 & 47.6 & \_ \\
             XciT-M24/8  & $224\times224$ & 110.0 & 2161  & 48.4 & \_ \\

             PatchConvNet-B60 & $224\times224$ & 140.6 & 1258  & 48.1  & 48.6 \\

             PatchConvNet-B120 & $224\times224$ & 229.8 & 1550  & 49.4  & 50.3\\
             MAE ViT-B & $224\times224$ & 127.7 & 1283 & 48.1 &  \_ \\
             Swin-B  & $384\times384$ & 121.0 & 1188  & 48.1 & 49.7 \\
             \rowcolor{Goldenrod}
            Our ViT-B & $224\times224$ & 127.7 & 1283 & 49.3 &  50.2\\
            \rowcolor{Goldenrod}
            Our ViT-L & $224\times224$ & 353.6 & 2231 & \textbf{51.5} & \textbf{52.0}  \\            
            \midrule

            PatchConvNet--B60$^\dagger$ & $224\times224$ & 140.6 & 1258  & 50.5  & 51.1 \\

             PatchConvNet-L120$^\dagger$ & $224\times224$ & 383.7 & 2086  &  52.2 & 52.9   \\
            Swin-B$^\dagger$ ($640\times 640$) & $224\times224$ & 121.0 & 1841  & 50.0 & 51.6 \\
            Swin-L$^\dagger$ ($640\times 640$) & $224\times224$ & 234.0 & 3230  & \_ & 53.5 \\
            \rowcolor{Goldenrod}
            Our ViT-B$^\dagger$ & $224\times224$ & 127.7 & 1283 & 51.8 & 52.8\\
            \rowcolor{Goldenrod}
            Our ViT-B$^\dagger$ & $384\times384$ & 127.7 & 1283 & 53.4 & 54.1\\
            \rowcolor{Goldenrod}
            Our ViT-L$^\dagger$ & $224\times224$ & 353.6 & 2231 &  53.8 & 54.7 \\
            \rowcolor{Goldenrod}
            Our ViT-L$^\dagger$ & $320\times320$ & 353.6 & 2231 &  \textbf{54.6} & \textbf{55.6} \\
        \bottomrule     
        \end{tabular}
        } 
            \bigskip 
\end{table}

\subsubsection{Training with others architectures}

In Table~\ref{tab:comp_arch_train} we measure the top-1 accuracy on ImageNet-val, ImageNet-real and ImageNet-v2 with different architecture train with our training procedure at resolution $224 \times 224$ on ImageNet-1k only. 
We can observe that for some architectures like PiT or CaiT our training method will improve the performance. For some others like TNT our approach is neutral and for architectures like Swin it decreases the performance. This is consistent with the findings of Wightman et al.~\cite{wightman2021resnet} and illustrates the need to improve the training procedure in conjunction to the architecture to obtain robust conclusions.
Indeed, adjusting these architectures while keeping the training procedure fixed can probably have the same effect as keeping the architecture fixed and adjusting the training procedure. 
That means that with a fixed training procedure we can have an overfitting of an architecture for a given training procedure.
In order to take overfitting into account we perform our measurements on the ImageNet val and ImageNet-v2 to quantify the amount of overfitting.

\begin{table}
    \centering
    \scalebox{0.85}{
    \begin{tabular}{l|cc|c|ccc}
    \toprule
         \multirow{2}{*}{Model} & Params & Flops & \multicolumn{4}{c}{ImageNet-1k}  \\
         & ($\times 10^6$) &  ($\times 10^9$) & orig. &  val & real & v2 \\
        \midrule
        \textcolor{lightgray}{ViT-S~\cite{Touvron2020TrainingDI}}                      & \textcolor{lightgray}{22.0} & \textcolor{lightgray}{4.6} &\textcolor{lightgray}{79.8} & \textcolor{lightgray}{80.4} & \textcolor{lightgray}{86.1} & \textcolor{lightgray}{69.7} \\
        \textcolor{lightgray}{ViT-B~\cite{dosovitskiy2020image,Touvron2020TrainingDI} }& \textcolor{lightgray}{86.6} & \textcolor{lightgray}{17.6} &\textcolor{lightgray}{81.8} & \textcolor{lightgray}{83.1} & \textcolor{lightgray}{87.7} & \textcolor{lightgray}{72.6} \\
         \midrule
         PiT-S~\cite{Heo2021RethinkingSD}  & 23.5 & 2.9 & \textcolor{lightgray}{80.9} & 80.4 & 86.1 & 69.2 \\
         PiT-B~\cite{Heo2021RethinkingSD}  & 73.8 & 12.5 & \textcolor{lightgray}{82.0} & 82.4 & 86.8 & 72.0 \\
         \midrule
         TNT-S~\cite{Yuan2021TokenstoTokenVT} & 23.8 & 5.2 & \textcolor{lightgray}{81.5} & 81.4 & 87.2 & 70.6 \\
         TNT-B~\cite{Yuan2021TokenstoTokenVT} & 65.6 & 14.1 & \textcolor{lightgray}{82.9} & 82.9 & 87.6 & 72.2\\
         \midrule
         
         ConViT-S~\cite{dAscoli2021ConViTIV} & 27.8 & 5.8 & \textcolor{lightgray}{81.3} & 81.3 & 87.0 & 70.3 \\
         ConViT-B~\cite{dAscoli2021ConViTIV} & 86.5 & 17.5 & \textcolor{lightgray}{82.4} & 82.0 & 86.7 & 71.3 \\
         \midrule
         Swin-S~\cite{liu2021swin} & 49.6 & 8.7 &\textcolor{lightgray}{83.0} & 82.1 & 86.9  & 70.7 \\
         Swin-B~\cite{liu2021swin} & 87.8 & 15.4 &\textcolor{lightgray}{83.5} & 82.2 & 86.7 & 70.7  \\
          \midrule
         CaiT-B12~\cite{touvron2021going} & 100.0 & 18.2 &\textcolor{lightgray}{\_} & 83.3 & 87.7 & 73.3  \\

        \bottomrule
    \end{tabular}}
    \caption{We report the performance reached with our training recipe with 400 epochs at resolution $224\times 224$ for other transformers architectures. We have not performed an extensive grid search to adapt the hyper-parameters to each architecture. Our results are overall similar to the ones achieved in the papers where these architectures were originally published (reported in column 'orig.'), except for Swin Transformers, for which we observe a drop on ImageNet-val.
    \label{tab:comp_arch_train}}
\end{table}

\section{Conclusion}

This paper makes a simple contribution: it proposes improved baselines for vision transformers trained in a supervised fashion that can serve (1) either as a comparison basis for new architectures; (2) or for other training approaches such as those based on self-supervised learning. 
We hope that this stronger baseline will serve the community effort in making progress on learning foundation models that could serve many tasks. 
Our experiments have also gathered a few insights on how to train ViT for larger models with reduced resources without hurting accuracy, allowing us to train a one-billion parameter model with 4 nodes of 8 GPUs. 
\paragraph{Acknowledgement.} We thank Ishan Misra for his valuable feedback.

\clearpage

\bibliographystyle{splncs04}
\bibliography{egbib}
\clearpage

\appendix

\appendix

\begin{center}
{\Large ~ \\ Appendices \\ }
\end{center}

\section{Experimental details}
\label{appd:hparams}

\paragraph{\bf Fine-tuning at higher resolution}
\label{sec:finetuning_hres}
When pre-training on ImageNet-1k at resolution $224 \times 224$ we fix the train-test resolution discrepancy by finetuning at a higher resolution~\cite{Touvron2019FixRes}. 
Our finetuning procedure is inspired by DeiT, except that we adapt the stochastic depth rate according to the model size~\cite{touvron2021going}. We fix the learning reate to $lr=1\times10^{-5}$ with batch-size=$512$ during 20 epochs with a weight decay of $0.1$ without repeated augmentation. Other hyper-parameters are similar to those employed in DeiT fine-tuning.

\paragraph{\bf Stochastic depth}
We adapt the stochastic depth drop rate according to the model size.
We report stochastic depth drop rate values in Table~\ref{tab:std_depth_rate}.
\begin{table}[h!]
    \centering
    \scalebox{0.9}{
    \begin{tabular}{c|cc|cc}
    \toprule
         \multirow{2}{*}{Model} & \# Params & FLOPs &  \multicolumn{2}{c}{Stochastic depth drop-rate}\\[3pt] 
         & ($\times 10^6$)& ($\times 10^9$) & ImageNet-1k & ImageNet-21k \\
         \midrule
         
         ViT-T & \dzo5.7  & \dzo1.3  & 0.0 & 0.0 \\
         ViT-S & \pzo22.0 & \dzo4.6  & 0.0 & 0.0 \\
         ViT-B & \pzo86.6 & \pzo17.5 & 0.1 & 0.1 \\
         ViT-L & 304.4    & \pzo61.6 & 0.4 & 0.3 \\
         ViT-H & 632.1    & 167.4    & 0.5 & 0.5 \\
         \bottomrule
    \end{tabular}}
    \caption{Stochastic depth drop-rate according to the model size. For 400 epochs training on ImageNet-1k and 90 epochs training on ImageNet-21k. See section~\ref{appd:abaltion} for further adaption with longer training.}
    \label{tab:std_depth_rate}
\end{table}

\paragraph{\bf For transfer learning}  experiments we evaluate our models pre-trained at resolution $224 \times 224$ on ImageNet-1k only on 6 transfer learning datasets.
We give the details of these datasets in Table~\ref{tab:dataset} below.

\begin{table}[h!]
\centering
\scalebox{0.9}{
\begin{tabular}{l|rrr}
\toprule
Dataset & Train size & Test size & \#classes   \\
\midrule
iNaturalist 2018~\cite{Horn2018INaturalist}& 437,513   & 24,426 & 8,142 \\ 
iNaturalist 2019~\cite{Horn2019INaturalist}& 265,240   & 3,003  & 1,010  \\ 
Flowers-102~\cite{Nilsback08}& 2,040 & 6,149 & 102  \\ 
Stanford Cars~\cite{Cars2013}& 8,144 & 8,041 & 196  \\  
CIFAR-100~\cite{Krizhevsky2009LearningML}  & 50,000    & 10,000 & 100   \\ 
CIFAR-10~\cite{Krizhevsky2009LearningML}  & 50,000    & 10,000 & 10   \\ 
\bottomrule
\end{tabular}}%
\caption{Datasets used for our different transfer-learning tasks.  \label{tab:dataset}}
\end{table}

\section{Additional Ablations}
\label{appd:abaltion}

\paragraph{\textbf{Number of training epochs}}
In Table~\ref{tab:number_epochs} we provide an ablation on the number of training epochs on ImageNet-1k. 
We do not observe a  saturation when the increase of the number of training epochs, as observed with BerT like approaches~\cite{bao2021beit,He2021MaskedAA}.
For longer training we increase the weight decay from 0.02 to 0.05 and we increase the stochastic depth drop-rate by 0.05 every 200 epochs to prevent overfitting.

\begin{table}
    \centering
    \scalebox{0.9}{
    \begin{tabular}{cc|crc}
    \toprule
        \multirow{2}{*}{Model} & \multirow{2}{*}{epochs} & \multicolumn{3}{c}{ImageNet top1 acc.}  \\
         & & val & real & v2 \\
         \midrule
         \multirow{4}{*}{ViT-S} & 
         300 & 79.9  & 86.1 & 68.8 \\%
         & 400 & 80.4 & 86.1 & 69.7  \\
         & 600 & 80.8 & 86.7 & 69.9 \\%
         & 800 & 81.4 & 87.0 & 70.5 \\%
  \midrule
        \multirow{4}{*}{ViT-B} & 
        300 & 82.8 & 87.6 & 72.1 \\%
        & 400 & 83.1 & 87.7 & 72.6 \\
        & 600 & 83.2 & 87.8 & 73.3  \\%
        & 800 & 83.7 & 88.1 & 73.1  \\         
\midrule
        \multirow{4}{*}{ViT-L} & 
        300 &  84.1 & 88.5 & 74.1 \\ %
        & 400 & 84.2 & 88.6 & 74.3  \\
        & 600 & 84.4 & 88.6 & 74.6  \\%
        & 800 & 84.5 & 88.8 & 75.0 \\ %
\midrule
        \multirow{2}{*}{ViT-H} & 
        300 &  84.6 & 89.0 & 74.9 \\%
         & 400 & 84.8 & 89.1 & 75.3 \\
       \bottomrule         
    \end{tabular}}
    \caption{Impact on the performance of the number of training epochs on ImageNet-1k.\label{tab:number_epochs}}
\end{table}

\end{document}